\documentclass[10pt,journal,compsoc]{IEEEtran}

\usepackage{color}

\ifCLASSOPTIONcompsoc
\usepackage[nocompress]{cite}
\else
\usepackage{cite}
\fi

\usepackage[utf8]{inputenc}
\usepackage{longtable}
\usepackage{graphicx}
\usepackage{color}

\usepackage[cmex10]{amsmath}

\usepackage{array}

\usepackage{mdwmath}
\usepackage{mdwtab}
\usepackage{subfig}
\usepackage{url}

\usepackage{multirow}
\usepackage{booktabs}
\usepackage{amssymb}
\usepackage{lscape}

\usepackage{rotating}

\newcolumntype{?}{!{\vrule width 1pt}}

\newcommand{\image}{I}              		
\newcommand{\images}{\textbf{I}}     	
\newcommand{\nimages}{n}            		

\newcommand{\shape}{X}             		
\newcommand{\estimatedshape}{\hat{\shape}}
\newcommand{\vgrid}{V}              		

\newcommand{\point}{p}				

\newcommand{\vertex}{\textbf{v}}		
\newcommand{\vertices}{\mathcal{V}}	
\newcommand{\faces}{\mathcal{F}}		
\newcommand{\pointset}{S}			
\newcommand{\npoints}{N}

\newcommand{\viewpoint}{v}			

\newcommand{\recofunc}{f}               	
\newcommand{\encodingfunc}{h}      		
\newcommand{\decodingfunc}{g}      		
\newcommand{\discriminatingfunc}{D}      		
\newcommand{\confidencefunc}{C} 		
  
\newcommand{\params}{\theta}			

\newcommand{\featurevector}{\textbf{x}}   
\newcommand{\latentspace}{\mathcal{X}}	

\newcommand{\objectivefunc}{\mathcal{L}}
\newcommand{\loss}{\objectivefunc}			

\newcommand{\ie}{\emph{i.e., }}
\newcommand{\eg}{\emph{e.g., }}
\newcommand{\etal}{\emph{et al.}}

\newcommand{\noi}{\noindent}

\newcommand{\real}{\mathbb{R}}      	

\newcommand{\rthree}{\real^3}

\newcommand{\cameraparams}{\alpha}	

\newcommand{\mean}{\mu}					
\newcommand{\means}{\boldsymbol{\mean}}		
\newcommand{\sigmas}{\boldsymbol{\sigma}}		

\newcommand{\eigenvector}{\Lambda}			
\newcommand{\bases}{\textbf{B}}				
\newcommand{\base}{\textbf{b}}
\newcommand{\nbases}{n}

\newcommand{\domain}{\mathcal{D}}		
\newcommand{\stwo}{S^2}				
\newcommand{\shapefunc}{\zeta}			

\newcommand{\ltwo}{L_2}						
\newcommand{\lone}{L_1}						

\newcommand{\deformationfield}{\Delta}		
\newcommand{\deformation}{\delta}			
\newcommand{\templateshape}{\tilde{\shape}}	

\newcommand{\width}{W}						
\newcommand{\height}{H}						

\newcommand{\lossgan}{\loss_{3D-GAN}}

\graphicspath{{figures/}{figures/encoding/}{figures/volumetric/}{figures/surface/}}
 \DeclareGraphicsExtensions{.pdf,.eps,.png}

\begin{document}
\bstctlcite{IEEEexample:BSTcontrol}


\title{Image-based 3D Object Reconstruction: State-of-the-Art and Trends in the Deep Learning Era}


\author{Xian-Feng Han*, Hamid Laga*,  Mohammed Bennamoun~\IEEEmembership{Senior Member,~IEEE}
	
	\IEEEcompsocitemizethanks{\IEEEcompsocthanksitem (* Joint first author) Xian-Feng Han is with College of Computer and Information Science, Southwest University, Chongqing 400715, China,  with Tianjin University, Tianjin, 300350, China and with the University of Western Australia, Perth, WA 6009, Australia.
		\IEEEcompsocthanksitem (* Joint first author) Hamid Laga  is with the Information Technology, Mathematics and Statistics Discipline, Murdoch University (Australia), and with the Phenomics and Bioinformatics Research Centre, University of South Australia.
		E-mail: H.Laga@murdoch.edu.au
		\IEEEcompsocthanksitem Mohammed Bennamoun is with the University of Western Australia, Perth, WA 6009, Australia. Email: mohammed.bennamoun@uwa.edu.au
	}
	\thanks{Manuscript received April 19, 2005; revised December 27, 2012.}}

\markboth{Deep Learning-based 3D Object Reconstruction - A Survey}%
{Han, Laga, and Bennamoun \MakeLowercase{\etalnospace}: Deep Learning-based 3D Object Reconstruction - A Survey}

\IEEEtitleabstractindextext{
\begin{abstract}
3D  reconstruction  is a longstanding ill-posed problem, which has been explored for decades by the computer vision, computer graphics, and machine learning communities.  Since 2015, image-based 3D reconstruction using convolutional neural networks (CNN) has attracted increasing interest and demonstrated an impressive performance. Given this new era of rapid evolution,  this article  provides a comprehensive survey of the recent developments in this field. We  focus on the works which use deep learning techniques to estimate the 3D shape of generic objects either from a single or multiple RGB images.   We organize the literature based on the shape representations, the network architectures, and the training mechanisms they use.  While this survey is intended for methods which reconstruct generic objects, we also review some of the recent works which focus on specific object classes such as human body shapes and faces.  We provide an analysis and comparison of the performance of some key papers,  summarize some of the open problems in this field, and discuss promising directions for future research.

\end{abstract}

\begin{IEEEkeywords}
3D Reconstruction, Depth Estimation, SLAM, SfM, CNN, Deep Learning, LSTM, 3D face, 3D Human Body, 3D Video.
\end{IEEEkeywords}
}

\maketitle

\IEEEdisplaynotcompsoctitleabstractindextext

%
\IEEEpeerreviewmaketitle


\section{Introduction}
\label{sec:introducton}


The goal of image-based 3D reconstruction is to infer the 3D geometry and structure of  objects and scenes from one or multiple 2D images. This long standing ill-posed problem is fundamental to many applications such as robot navigation, object recognition and scene understanding, 3D modeling and animation,  industrial control, and medical diagnosis.

Recovering the lost dimension from just 2D images has been the goal of classic multiview stereo and  shape-from-X methods, which have  been extensively investigated for many decades. The first generation of methods approached the problem from the geometric perspective; they   focused on  understanding and formalizing, mathematically, the 3D to 2D projection process, with the aim to devise mathematical or algorithmic solutions to the ill-posed inverse problem. Effective solutions typically  require multiple images, captured using accurately calibrated cameras. Stereo-based techniques~\cite{hartley2003multiple}, for example, require matching features across images captured from slightly different viewing angles, and then use the triangulation principle to recover the 3D coordinates of the image pixels. Shape-from-silhouette, or shape-by-space-carving,  methods~\cite{laurentini1994visual} require accurately segmented 2D silhouettes.  These methods, which have led to reasonable quality 3D reconstructions,  require multiple images of the same object  captured by well-calibrated cameras. This, however, may not be practical or feasible in many situations.

Interestingly, humans are good at solving such ill-posed inverse problems by leveraging prior knowledge. They can  infer the approximate size and rough geometry of objects using only one eye. They can even guess what it would look like from another viewpoint. We can do this because all the previously seen objects and scenes have enabled us to build prior knowledge and develop mental models of what objects look like. The second generation of 3D reconstruction  methods tried to leverage this prior knowledge by formulating the 3D reconstruction problem as a recognition problem. The avenue of deep learning techniques, and more importantly, the increasing availability of large training data sets, have led to a new generation of methods that are able to recover the 3D geometry and structure of  objects from one or multiple RGB images without the complex camera calibration process.  Despite being recent, these methods have demonstrated  exciting and promising results on various tasks related to computer vision and graphics.


In this article,  we provide a comprehensive and structured review of the recent advances in 3D object reconstruction using deep learning  techniques. We first focus on generic shapes and then discuss specific cases, such as human body shapes faces reconstruction, and 3D scene parsing. We have gathered  $149$ papers, which appeared since $2015$  in leading computer vision, computer graphics, and machine learning conferences and journals\footnote{This  continuously and rapidly increasing number,  even at the time we are finalising this article, does not include many of the CVPR2019 and the upcoming ICCV2019 papers.}.  The goal is to help the reader navigate in this emerging field, which gained a significant momentum in the past few years. Compared to the existing literature, the main contributions of this article are as follows;
\begin{enumerate}
	\item To the best of our knowledge, this is the first survey 	paper in the literature which focuses on image-based 3D object reconstruction using deep learning.
	
	\item We  cover the contemporary literature with respect to this area. We present a comprehensive review of $149$ methods, which appeared since $2015$.
	
	\item We provide a comprehensive review and an insightful analysis on all aspects of 3D reconstruction using deep learning, including the training data, the choice of network architectures and their effect on the 3D reconstruction results, the training strategies, and the application scenarios. 
	
	\item We provide a comparative summary of the properties and performance of the reviewed methods for  generic 3D object reconstruction.  We  cover $88$  algorithms for generic 3D object reconstruction, $11$ methods related to 3D face reconstruction,  and $6$ methods for 3D human body shape reconstruction. 
	
	\item We provide a comparative summary of the methods in a tabular form. 
	
\end{enumerate} 

\noi The rest of this article is organized as follows; Section~\ref{sec:problemstatement} fomulates the problem and lays down the taxonomy.  Section~\ref{sec:encoding} reviews the latent spaces and the input encoding mechanisms. Section~\ref{sec:volumetric_3D_reconstruction} surveys the volumetric reconstruction techniques, while Section~\ref{sec:3D_surface_decoding} focuses on surface-based techniques. Section~\ref{sec:other_cues} shows how some of the state-of-the-art techniques use additional cues to boost the performance of 3D reconstruction. Section~\ref{sec:training_deeplearning} discusses the training procedures. Section~\ref{sec:applications} focuses on specific objects such as human body shapes and faces. Section~\ref{sec:datasets_evaluation} summarizes the most commonly used datasets to train, test, and evaluate the performance of various deep learning-based 3D reconstruction algorithms. Section~\ref{sec:performance_comparison} compares and discusses the performance of some key methods.   Finally, Section~\ref{sec:future_work} discusses potential future research directions while Section~\ref{sec:summary} concludes the paper  with some important remarks.


\section{Problem statement and taxonomy}
\label{sec:problemstatement}

Let $\images = \{\image_k, k=1, \dots, \nimages \}$ be a set of $\nimages \ge 1$ RGB images of one or multiple objects $\shape$.  3D reconstruction can be summarized as the process of learning a predictor $\recofunc_{\params}$ that can infer a shape $\hat{\shape}$ that is as close as possible to the unknown shape $\shape$. In other words, the  function $\recofunc_{\params}$ is the minimizer   of a reconstruction objective $ \objectivefunc(\images) = d\left(\recofunc_{\theta}(\images), \shape  \right)$. Here, $\theta$ is the set of parameters of $\recofunc$  and  $d(\cdot, \cdot)$ is a certain measure of distance between the target shape $\shape$ and the reconstructed shape $\recofunc(\images)$.  The reconstruction objective  $\objectivefunc$ is also known as the  \emph{loss function} in the deep learning literature. 

This survey discusses and categorizes the state-of-the-art based on the nature of the input $\images$, the representation of the output, the deep neural network architectures used during training and testing to approximate the predictor $\recofunc$,  the training procedures they use, and their degree of supervision, see Table~\ref{tab:overall_taxonomy} for a visual summary. In particular, \emph{the input $\images$ } can be \textbf{(1)} a single image, \textbf{(2)} multiple  images  captured  using RGB cameras whose intrinsic and extrinsic parameters can be  \emph{known} or \emph{unknown}, or \textbf{(3)} a video stream, \ie a sequence of  images with temporal correlation. The first case is very challenging because of the ambiguities in the 3D reconstruction.  When the input is a video stream, one can exploit the temporal correlation to facilitate the 3D reconstruction while ensuring that the reconstruction is smooth and consistent across all the frames of the video stream.  Also, the input can be depicting  one or multiple 3D objects belonging to known or unknown shape categories.  It can also include additional information such as silhouettes, segmentation masks,  and semantic labels as priors to guide the reconstruction. 

\emph{The representation of the output}  is crucial to the choice of the network architecture. It also impacts the computational efficiency and quality of the reconstruction. In particular, 
\begin{itemize}
	\item \emph{Volumetric representations}, which have been extensively adopted in early deep leaning-based 3D reconstruction techniques, allow the parametrization of 3D shapes using regular voxel grids. As such, 2D convolutions used in image analysis can be easily extended to 3D.  They are, however, very expensive in terms of  memory requirements, and only a few techniques can achieve sub-voxel accuracy. 
	
	
	\item \emph{Surface-based representations}:  Other  papers  explored surface-based representations such as meshes and  point clouds. While being memory-efficient, such  representations are not regular structures and thus, they do not easily fit into deep learning architectures.
	
	\item \emph{Intermediation: }  While some 3D reconstruction algorithms predict the 3D geometry of an object from RGB images directly, others decompose the problem into sequential steps, each step predicts an intermediate representation.
\end{itemize}

\begin{table}[t]
\centering{
	\caption{\label{tab:overall_taxonomy}Taxonomy of the state-of-the-art  image-based 3D object reconstruction using deep learning. }
	\resizebox{\linewidth}{!}{%
	
	\begin{tabular}{?@{ }c|c|c|c @{ }?}  
		\noalign{\hrule height 1pt}
		\multirow{5}{*}{\textbf{Input}} & \multirow{3}{*}{Training} & 1 vs. muli RGB, &    \\
							    & 					     & 3D ground truth, &\multirow{3}{3.5cm}{One vs. multiple objects, Uniform vs. cluttered background.} \\
							    &					    & Segmentation.     &  \\
				\cline{2-3}
							    & \multirow{2}{*}{Testing} &  1 vs. muli RGB, &  \\
							    &					   &  Segmentation     & \\
		\noalign{\hrule height 1pt}
		\multirow{3}{*}{\textbf{Output}} & Volumetric &  \multicolumn{2}{c?}{High vs. low resolution}	\\
				\cline{2-4}
								& \multirow{2}{*}{Surface} 	   &  \multicolumn{2}{c?}{Parameterization, template deformation,}	\\
								&		   &  \multicolumn{2}{c?}{Point cloud.}\\
				\cline{2-4}
								& \multicolumn{3}{c?}{Direct vs. intermediating} \\
		\noalign{\hrule height 1pt}
		
		\multicolumn{1}{?c|}{\multirow{5}{1.5cm}{\textbf{Network architecture}}}  & \multicolumn{2}{c?}{\textbf{Architecture at training}} & \textbf{Architecture at testing}  \\
								\cline{2-4} 
								& \multicolumn{2}{c?}{Encoder - Decoder} & \multirow{3}{*}{Encoder - Decoder} 	 \\
								&  \multicolumn{2}{c?}{TL-Net} & \\
								&  \multicolumn{2}{c?}{(Conditional) GAN} & \\
								\cline{2-4} 
								& \multicolumn{2}{c?}{3D-VAE-GAN} &  3D-VAE\\

		\noalign{\hrule height 1pt}
		
		\multicolumn{1}{?c|}{\multirow{5}{1.5cm}{\textbf{Training}}}  & \multirow{2}{2cm}{Degree of supervision}   &  \multicolumn{2}{c?}{2D vs. 3D supervision. Weak supervision.} \\
				\cline{3-4}
													  &			& \multicolumn{2}{c?}{Loss functions.} 	\\
				\cline{2-4}
													  & \multirow{2}{2cm}{Training procedure}  &   \multicolumn{2}{c?}{Adversarial training.  Joint 2D-3D embedding.} \\
													    &								 &  \multicolumn{2}{c?}{Joint training with other tasks.} \\
		\noalign{\hrule height 1pt}
	\end{tabular}
	}
}
\end{table}

\noi A variety of \emph{network architectures}  have been utilized to  implement the predictor $\recofunc$. The backbone architecture, which can be different during training and testing,  is composed of an encoder $\encodingfunc$ followed by a decoder $\decodingfunc$, \ie $\recofunc = \decodingfunc \circ \encodingfunc$. The encoder maps the input into a latent variable $\featurevector$, referred to as a feature vector or a code,  using a sequence of convolutions and pooling operations, followed by fully connected layers of neurons. The decoder, also called the generator, decodes the feature vector into the desired output by using  either fully connected layers or a deconvolution network (a sequence of convolution and upsampling operations, also referred to as upconvolutions). The former is suitable for unstructured output, \eg 3D point clouds, while the latter is used to reconstruct volumetric grids or parametrized surfaces. Since the introduction of this vanilla architecture, several extensions have been proposed by varying the architecture (\eg ConvNet vs. ResNet, Convolutional Neural Networks (CNN) vs. Generative Adversarial Networks (GAN), CNN vs. Variational Auto-Encoders, and 2D vs. 3D convolutions), and by cascading multiple blocks each one achieving a specific task.

While the architecture of the network and its building blocks are important, the performance depends highly on the way it is trained.  In this survey, we will  look at:
\begin{itemize}
	\item \emph{Datasets: } There are  various datasets that are currently available for training and evaluating deep learning-based 3D reconstruction. Some of them use real data, other are CG-generated. 
	
	\item \emph{Loss functions: }   The choice of the loss function   can significantly impact on the reconstruction quality.  It also defines the degree of supervision. 
	
	\item \emph{Training procedure sand  degree of  supervision: }  Some methods require real images annotated with their corresponding 3D models, which are very expensive to obtain. Other methods rely on a combination of real and synthetic data. Others avoid completely 3D supervision by using  loss functions that exploit  supervisory signals that are easy to obtain. 
\end{itemize}


\noi  The following sections review in detail these aspects.

\section{The encoding stage}
\label{sec:encoding}

Deep learning-based 3D reconstruction algorithms encode the input  $\images$ into a  feature vector $\featurevector = \encodingfunc(\images) \in \latentspace$ where $\latentspace$ is  the latent space.  A good mapping function $\encodingfunc$ should satisfy the following  properties:
\begin{itemize}
	\item Two inputs $\images_1$ and $\images_2$ that represent similar 3D objects should be mapped into $\featurevector_1$ and $\featurevector_2\in \latentspace$ that are close to each other in the latent space. 
	
	\item A small perturbation $ \partial\featurevector$ of   $\featurevector$ should correspond to a small perturbation of the shape of the input. 
	
	\item The latent representation induced by  $\encodingfunc$ should be invariant to extrinsic factors such as the  camera pose. 
	
	\item A 3D model and its corresponding 2D images should be  mapped onto the same point in the latent space. This will ensure that the representation is not ambiguous and thus facilitate the reconstruction. 
\end{itemize}
	
\noi The first two conditions have been addressed by using encoders that map the input onto discrete (Section~\ref{sec:encoding_discrete}) or continuous (Section~\ref{sec:encoding_continuous}) latent spaces. These  can be flat or hierarchical (Section~\ref{sec:encoding_hierarchical}).  The third one has been addressed by using disentangled representations (Section~\ref{sec:disentangled}). The latter has been addressed by using TL-architectures during the  training phase. This is covered in Section~\ref{sec:joint_image_3D_embedding} as one of the many training mechanisms that have been used in the literature. Table~\ref{tab:encoder_taxonomy} summarizes this taxonomy.

\begin{table}
\centering{
\caption{\label{tab:encoder_taxonomy}Taxonomy of the encoding stage. FC: fully-connected layers. VAE: Variational Auto-Encoder.  }

{%
	\begin{tabular}{@{}cc@{}}
	\toprule
	\textbf{Latent spaces} & \textbf{Architectures} \\
	\midrule
	Discrete (\ref{sec:encoding_discrete}) vs. continuous (\ref{sec:encoding_continuous})&ConvNet, ResNet,\\
	Flat vs. hierarchical 	(\ref{sec:encoding_hierarchical}) & FC, 3D-VAE \\ 
	Disentangled representation (\ref{sec:disentangled}) & \\
	\bottomrule

	\end{tabular}
	}
}
\end{table}

\subsection{Discrete latent spaces}  
\label{sec:encoding_discrete}

Wu \etal   \ in their seminal work~\cite{wu20153d} introduced  3D ShapeNet, an encoding network which maps a 3D shape, represented as  a discretized volumetric grid of size  $30^3$, into a latent representation of size $4000\times1$.  Its core network is composed of $n_{conv} = 3$ convolutional layers (each one using  3D convolution filters),  followed by  $n_{fc}=3$ fully connected layers. This standard vanilla architecture has been used for 3D shape classification and retrieval~\cite{wu20153d},  and for 3D reconstruction from depth maps represented as voxel grids~\cite{wu20153d}. It has also been  used in the 3D encoding branch of the TL architectures during the  training of 3D reconstruction networks, see Section~\ref{sec:joint_image_3D_embedding}.

2D encoding networks that map input images into a latent space follow the same architecture as 3D ShapeNet~\cite{wu20153d}  but use  2D convolutions~\cite{yan2016perspective,grant2016deep,wu2017marrnet,choy20163d,tulsiani2017multi,sun2018pix3d,zisserman2017silnet,Tulsiani_2018_CVPR}. Early works differ in the type and number of layers they use. For instance,  Yan \etal~\cite{yan2016perspective} use $n_{conv} = 3$ convolutional layers with $64, 128$, ad $256$ channels, respectively,  and $n_{fc} = 3 $ fully-connected layers  with  $1024$, $1024$, and $512$ neurons, respectively.  Wiles and Zisserman~\cite{zisserman2017silnet}   use $n_{conv} = 6$ convolutional layers of $3, 64, 128, 256, 128$, and $160$ channels, respectively.   Other works add pooling layers~\cite{choy20163d,johnston2017scaling}, and leaky Rectified Linear Units (ReLU)~\cite{choy20163d,johnston2017scaling,Yang_2018_ECCVlearning}.  For example, Wiles and Zisserman~\cite{zisserman2017silnet}    use max pooling layers between each pair of convolutional layers, except after the first layer and before the last layer.  ReLU layers  improve learning since the gradient during the back propagation  is never zero.

Both 3D shape and 2D image encoding networks can  be implemented using  deep residual networks (ResNet)~\cite{he2016deep}, which add  residual connections between the  convolutional layers, see for example~\cite{choy20163d,wu2017marrnet,sun2018pix3d}. Compared to conventional networks such as VGGNet~\cite{simonyan2014very}, ResNets improve and speed up the learning process for very deep networks. 

\subsection{Continuous latent spaces}
\label{sec:encoding_continuous}

Using the encoders presented in the previous section, the latent space $\latentspace$ may not be continuous and thus it does not allow easy interpolation. In other words, if $\featurevector_1  =\encodingfunc(\images_1)$ and $\featurevector_2 = \encodingfunc(\images_2)$, then there is no guarantee that $\frac{1}{2}(\featurevector_1 + \featurevector_2)$ can be decoded into a valid 3D shape. Also, small perturbations of $\featurevector_1$ do not necessarily correspond to small perturbations of the input.  Variational Autoencoders (VAE)~\cite{kingma2014auto} and their 3D extension (3D-VAE)~\cite{wu2016learning}  have one fundamentally unique property  that makes them suitable for generative modeling: their latent spaces are, by design, continuous, allowing easy  sampling and interpolation. The key idea  is that instead of mapping the input into a feature vector, it is mapped into a  mean vector $\means$ and a vector of standard deviations $\sigmas$ of a multivariate Gaussian distribution. A sampling layer then takes these two vectors,  and generates, by random sampling from the  Gaussian distribution,  a feature vector $\featurevector$, which will serve as input to the subsequent decoding stages. 


This architecture has been used to learn  continuous latent spaces  for volumetric~\cite{wu2016learning,liu2017learning}, depth-based~\cite{soltani2017synthesizing}, surface-based~\cite{henderson2018learning}, and  point-based~\cite{mandikal20183d,gadelha2018multiresolution} 3D reconstruction.  In Wu \etal~\cite{wu2016learning}, for example, the image encoder takes  a $256\times256$ RGB image and outputs two $200$-dimensional  vectors representing, respectively, the mean and the standard deviation of  a Gaussian distribution in the $200$-dimensional space. 
Compared to standard encoders, 3D-VAE can be used to randomly sample from the latent space, to generate variations of an input, and to reconstruct multiple plausible 3D shapes from an input image~\cite{mandikal20183d,gadelha2018multiresolution}. It  generalizes well to images that have not been  seen during the training.


\subsection{Hierarchical latent spaces}
\label{sec:encoding_hierarchical}

Liu \etal~\cite{liu2017learning} showed that  encoders that map the input into a single latent representation cannot  extract rich structures   and thus may lead to blurry reconstructions.  To improve the quality of the reconstruction, Liu \etal~\cite{liu2017learning}  introduced a more complex internal variable structure, with the specific goal of encouraging the learning of a hierarchical arrangement of latent feature detectors.  The approach starts with  a global latent variable layer  that is hardwired to a set of local latent variable layers, each tasked with representing one level of feature abstraction. The skip-connections tie together the latent codes in a top-down directed fashion: local codes closer to the input will tend to represent lower-level features while local codes farther away from the input will tend towards representing higher-level features. Finally, the local latent codes are concatenated to a flattened structure when fed into the task-specific models such as 3D reconstruction.

\subsection{Disentangled representation}
\label{sec:disentangled}

The appearance of an object in an image is affected by multiple factors such as the object's shape, the camera pose, and the lighting conditions. Standard encoders represent all these variabilities in the learned code $\featurevector$. This is not desirable in applications such as recognition and classification, which should be invariant to extrinsic factors such as pose and lighting~\cite{laga20193d}. 3D reconstruction can also benefit from disentangled representations where shape, pose, and lighting are represented with different codes. To this end, Grant \etal~\cite{grant2016deep} proposed an encoder, which maps an RGB image into a shape code and a transformation code. The former is  decoded into a 3D shape. The latter, which  encodes lighting conditions and pose, is decoded into \textbf{(1)}  another $80\times80$ RGB image with correct lighting, using upconvolutional layers,  and \textbf{(2)} camera pose using fully-connected layers (FC). To enable a disentangled representation, the network is trained in such a way that in the forward pass, the image decoder receives input from the shape code and the transformation code. In the backward pass,  the signal from the image decoder to the shape code is suppressed to force it to only represent shape.

Zhu \etal~\cite{zhu2017rethinking} followed the same idea by decoupling the 6DOF pose parameters and shape. The network  reconstructs from the 2D input the 3D shape but in a canonical pose. At the same time, a pose regressor estimates the 6DOF pose parameters, which are then applied to the reconstructed canonical shape. Decoupling pose and shape reduces the number of free parameters in the network, which results in improved efficiency. 
   
\section{Volumetric decoding} 
\label{sec:volumetric_3D_reconstruction}



Volumetric representations  discritize the  space around a 3D object into a 3D voxel  grid $\vgrid$. The finer the discretization   is, the more accurate the representation will be.  The goal is  then to recover a grid $\hat{\vgrid} = \recofunc_{\params}(\images) $ such that the 3D shape $\hat{\shape}$ it represents is as close as possible to the unknown real 3D shape $\shape$.  The main advantage of using volumetric grids  is that many of the existing deep learning architectures that have been designed for 2D image analysis can be easily  extended to 3D data by replacing the 2D pixel array with its 3D analogue and then processing the grid using 3D convolution and pooling operations.  
This section looks at the different volumetric representations   (Section~\ref{sec:volumetric_reps}) and reviews the  decoder architectures for low-resolution (Section~\ref{sec:lowres_reconstruction_decoding}) and   high-resolution (Section~\ref{sec:highres_volumes}) 3D reconstruction.  


\begin{table*}
\label{tab:volumetric_decoder_taxonomy}
\centering{
\caption{Taxonomy of the various volumetric decoders used in the literature. Number in parentheses are the corresponding section numbers. MDN: Mixture Density Network. BBX: Bounding Box primitives. Part.: partitioning.  }

\resizebox{1\linewidth}{!}
{%
	\begin{tabular}{?@{ }c|c|c| c|c|c|c|c|c| c|c@{ }?}
	\noalign{\hrule height 1pt}
	\multirow{5}{*}{\textbf{}} &
	\multicolumn{2}{c|}{\textbf{Representation (\ref{sec:volumetric_reps}) }} &
	\multicolumn{6}{c|}{\textbf{Resolution}}  	 &
	\multicolumn{2}{c?}{\textbf{Architecture}}  \\
	\cline{2-11}
	&
	\multirow{4}{*}{Sampling} & \multirow{4}{*}{Content} &  \multirow{4}{.9cm}{Low res: $32^3$, $64^3$ (\ref{sec:lowres_reconstruction_decoding})} & \multicolumn{5}{c|}{High resolution (\ref{sec:highres_volumes})}  &   \multirow{4}{*}{Network} &  \multirow{4}{*}{Intermediation (\ref{sec:intermediating})}  \\
			 \cline{5-9}
	
	 &		     &    &   &\multicolumn{2}{c|}{{Space part. (\ref{sec:space_partitioning})} }        &  \multirow{3}{1cm}{Shape part. (\ref{sec:shape_partitionning})}          &  \multirow{3}{1cm}{Subspace  param. (\ref{sec:subspace_parameteirzation})}&  \multirow{2}{1.5cm}{Refinement (\ref{sec:coarse_to_fine_refinement})} & & \\
	\cline{5-6}
	&		     &  		   &   &  Fixed   & Learned  & 		 &  &  &  & \\
	&		     & 			   &   &  Octree & Octree	 &      	 &  &   & &  \\
	\cline{2-3}\cline{5-11}
	& Regular,	     & Occupancy, &  & Normal,	&  HSP, OGN, 	  & Parts,	   &  PCA,  &  Upsampling,  & FC, & (1) image $\to$ voxels, \\
	& Fruxel,       &SDF,		   &  & O-CNN,	& Patch-guide & Patches & DCT  & Volume slicing, & UpConv.&  (2) image $\to$  (2.5D,   \\
	& Adaptive     & TSDF	   &  & OctNet	& 			  &		  & 	       &  Patch synthesis, & &   silh.) $\to$ voxels  \\
	&		     &			   &   & 	 	& 	   		  &	 	& 	     &    Patch refinement & &  \\

	\noalign{\hrule height 1.3pt}
	 \cite{choy20163d}  & Regular & Occupancy &\checkmark & $-$& $-$& $-$& $-$& $-$& LSTM + UpConv &  image $\to$ voxels \\ 
	 \hline

	 \cite{girdhar2016learning}   &Regular & Occupancy &\checkmark & $-$&$-$ &$-$ & $-$& $-$& UpConv&  image $\to$ voxels\\  
	\hline

	\cite{wu2016learning} & Regular & Occupancy& \checkmark & $-$ & $-$  & $-$  & $-$  & $-$  &  UpConv&  image $\to$ voxels  \\ 
	\hline

	\cite{yan2016perspective}& Regular &  Occupancy & \checkmark & $-$& $-$& $-$& $-$& $-$& UpConv & image $\to$ voxels  \\
	\hline

	
	\cite{wu2017marrnet} & Regular &  Occupancy &\checkmark &$128^3$ & $-$ &$-$ &$-$ & $-$ & UpConv& (2)  \\ 
	\hline

	
	\cite{dai2017shape} & Regular & SDF & \checkmark& $-$& $-$ & $-$ &$-$ &patch synthesis  & UpConv &  scans $\to$  voxels\\ 
	\hline
	
	
	\cite{gadelha20173d} &  Regular& Occupancy & \checkmark & $-$&$-$ & $-$& $-$& $-$& UpConv & image $\to$ voxels  \\ 
	\hline
	
	\cite{johnston2017scaling}& Regular & Occupancy & \checkmark  & $-$& $-$ & $-$ &  DCT & $-$ & IDCT  &   image $\to$ voxels\\
	\hline
	
	
	\cite{liu2017learning} &Regular & Occupancy & \checkmark &  $-$& $-$ & $-$ & $-$ & $-$   & UpConv & image $\to$ voxels \\ 
	\hline

	
	
	\cite{wang2017shape} &Regular  & Occupancy & $-$ &  $128^3$ & $-$ & $-$ &$-$  &Volume slicing  &  \small{CNN $\to$ LSTM $\to$ CNN}&    image $\to$ voxels \\ 
	\hline
	\cite{zhu2017rethinking} & Regular  & Occupancy & \checkmark & $-$ & $-$ & $-$ & $-$ & $-$ & UpConv&  image $\to$ voxels\\ 
	
	\hline	
	\cite{zou20173d} &  Regular & TSDF & $-$  & $-$& $-$& Parts & $-$ & $-$ & LSTM + MDN  & depth $\to$ BBX \\ 

	\hline
	\cite{knyaz2018image} & Fruxel & Occupancy  & $-$ & $128^3$  & $-$ & $-$  & $-$ & $-$  & UpConv & image $\to$ voxels  \\ 
	
	\hline
	\cite{kundu20183d} &Regular &  TSDF & $-$  & $-$ & $-$  &  &PCA  & $-$ &  FC &  image $\to$ voxels  \ \\  
	
	\hline
	\cite{wang2017cnn} & Adaptive & $-$ & $-$ & O-CNN& $-$& $-$ & $-$ & $-$ & $-$  &  $-$ \\  

	\hline	
	\cite{tatarchenko2017octree} & Adaptive &$-$ &$-$ & &OGN &$-$  &$-$ &$-$ & $-$& $-$  \\  


	\hline
	\cite{tulsiani2017multi} &Regular & Occupancy & \checkmark & $-$ & $-$ & $-$ & $-$ &$-$  & UpConv &   image $\to$ voxels \\ 
	
	\hline
	\cite{sun2018pix3d} & Regular &  Occupancy& $-$ & $128^3$ & $-$ & $-$ & $-$ & $-$ & UpConv&  (2) \\  
	
	\hline
	\cite{wang2018adaptive} & Adaptive& Occupancy & $-$ & O-CNN & patch-guided & $-$ & $-$ & $-$ &  UpConv & image $\to$ voxels     \\ 
	
	\hline
	\cite{Yang_2018_ECCVlearning}& Regular& Occupancy & \checkmark& $-$& $-$ & $-$& $-$& $-$ & UpConv & image $\to$ voxels  \\ 
	
	\hline
	\cite{Cao_2018_ECCV} & Regular & TSDF & $-$ &OctNet & $-$ & $-$ & &Global to local & UpConv & scans $\to$ voxels    \\ 
	
	\hline
	\cite{Tulsiani_2018_CVPR} &Regular & Occupancy & \checkmark & $-$ & $-$ &$-$ &$-$ & $-$ & UpConv &  image $\to$ voxels \\ 
	
	\hline
	\cite{hane2019hierarchical} & Adaptive & Occupancy & $-$ & $-$  & HSP & $-$  & $-$  & $-$ & UpConv nets   & image $\to$ voxels   \\ 


	\noalign{\hrule height 1.3pt}
	\end{tabular}
	}
}
\end{table*}

\subsection{Volumetric representations of 3D shapes}
\label{sec:volumetric_reps}

There are four main volumetric representations that have been used in the literature:   
\begin{itemize}
	\item \textit{Binary occupancy grid. } In this representation, a voxel  is set to one if it belongs to the objects of interest,  whereas background voxels are set to zero. 
	
	\item \textit{Probabilistic occupancy grid. } Each voxel in a probabilistic occupancy grid encodes its probability of belonging to the objects of interest.
	
	\item \textit{The Signed Distance Function (SDF). } Each voxel encodes its signed distance to the closest surface point. It is negative if the voxel is located inside the object and positive otherwise.  
	
	\item \textit{Truncated Signed Distance Function (TSDF). } Introduced by Curless and Levoy~\cite{curless1996volumetric},  TSDF is computed by first estimating distances along the lines of sight of a range sensor, forming a projective signed distance field, and then truncating the field at small negative and positive values.
	
\end{itemize}

\noi Probabilistic occupancy grids are particularly suitable for machine learning algorithms which output likelihoods. SDFs provide an unambiguous estimate of surface positions and normal directions. However, they are not trivial to construct from partial data such as depth maps. TSDFs sacrifice  the full signed distance field that extends indefinitely away from the surface geometry, but allow for local updates of the field based on partial observations. They are suitable for reconstructing  3D volumes from a set of depth maps~\cite{dai2017shape,Cherabier_2018_ECCV,kundu20183d,Cao_2018_ECCV}.

In general, volumetric representations are created by regular sampling of the volume around the objects. Knyaz \etal~\cite{knyaz2018image} introduced a representation method called  Frustum Voxel Model or Fruxel, which combines  the depth representation with voxel grids. It uses the slices of the camera's 3D frustum to build the voxel space, and thus  provides precise alignment of voxel slices with the contours in the input image.

Also, common SDF and TSDF representations are discretised into a regular grid. Recently, however,  Park \etal~\cite{park2019deepsdf} proposed Deep SDF (deepSDF),  a generative deep learning model that produces a continuous SDF field from an input point cloud. Unlike the traditional SDF representation, DeepSDF can handle  noisy and incomplete data. It can also represent an entire class of shape

\subsection{Low resolution 3D volume reconstruction} 
\label{sec:lowres_reconstruction_decoding}


Once a compact vector representation of the input is learned using an encoder,  the next step is to learn the decoding function $\decodingfunc$, known as the \emph{generator} or the \emph{generative model}, which maps the vector representation into a volumetric voxel grid.  The standard approach uses a \emph{convolutional decoder}, called also  \emph{up-convolutional network}, which mirrors the convolutional encoder.   Wu \etal~\cite{wu20153d} were among the first to propose this methodology to reconstruct 3D volumes from depth maps.  
Wu \etal~\cite{wu2017marrnet} proposed a two-stage reconstruction network called MarrNet.  The first stage uses an encoder-decoder  architecture to reconstruct, from an input image, the depth map, the normal map, and the silhouette map.  These three maps, referred to as $2.5$ sketches,  are then used as input to another encoder-decoder  architecture, which regresses a volumetric 3D shape. The network has been later extended by Sun \etal~\cite{sun2018pix3d} to also regress  the pose of the input. The main advantage of this two-stage approach is that, compared to full 3D models, depth maps, normal maps, and silhouette maps are much easier to recover from 2D images. Likewise, 3D models are much easier to recover from these three modalities than from 2D images alone. This method, however,  fails to reconstruct complex, thin structures. 

Wu \etal's work~\cite{wu20153d} has led to several extensions~\cite{choy20163d,wu2016learning,gadelha20173d,smith2017improved,tulsiani2017multi}.    In particular,  recent works tried to  directly regress the 3D voxel grid~\cite{tulsiani2017multi,liu2017learning,Yang_2018_ECCVlearning,Tulsiani_2018_CVPR} without intermediation. Tulsiani \etal~\cite{tulsiani2017multi}, and later in~\cite{Tulsiani_2018_CVPR}, used a decoder composed of 3D upconvolution layers to  predict the voxel occupancy probabilities. Liu \etal~\cite{liu2017learning}  used a 3D upconvolutional neural network, followed by an element-wise logistic sigmoid, to decode the learned latent features into a 3D occupancy probability grid. These methods  have been  successful in performing 3D reconstruction from a single or a collection of images captured with uncalibrated cameras. Their main advantage  is that the deep  learning architectures proposed for the analysis of 2D images can be easily adapted to 3D models by replacing the 2D up-convolutions in the decoder with 3D up-convolutions, which also can be efficiently implemented on the GPU.  However, given the computational complexity and memory requirements, these methods  produce low resolution  grids, usually of size  $32^3$ or $64^3$. As such, they fail to recover fine details.   

\subsection{High resolution 3D volume reconstruction} 
\label{sec:highres_volumes}

There have been attempts to upscale the deep learning architectures for high resolution volumetric reconstruction.  For instance, Wu \etal~\cite{wu2017marrnet} were able to reconstruct voxel grids of size $128^3$ by simply expanding the network.  Volumetric grids, however, are  very expensive in terms of memory  requirements, which  grow cubically with the grid resolution. This section reviews some of the techniques that have been used to infer high resolution volumetric grids,   while keeping the computational and memory requirements tractable.  We classify these methods  into four  categories based on whether they use space partitioning, shape partitioning,   subspace parameterization, or coarse-to-fine refinement strategies. 

\begin{figure*}[t]
\centering{

\resizebox{\linewidth}{!}{%
	\begin{tabular}{@{}c@{}c@{}c@{}}
		\includegraphics[width=0.24\textwidth]{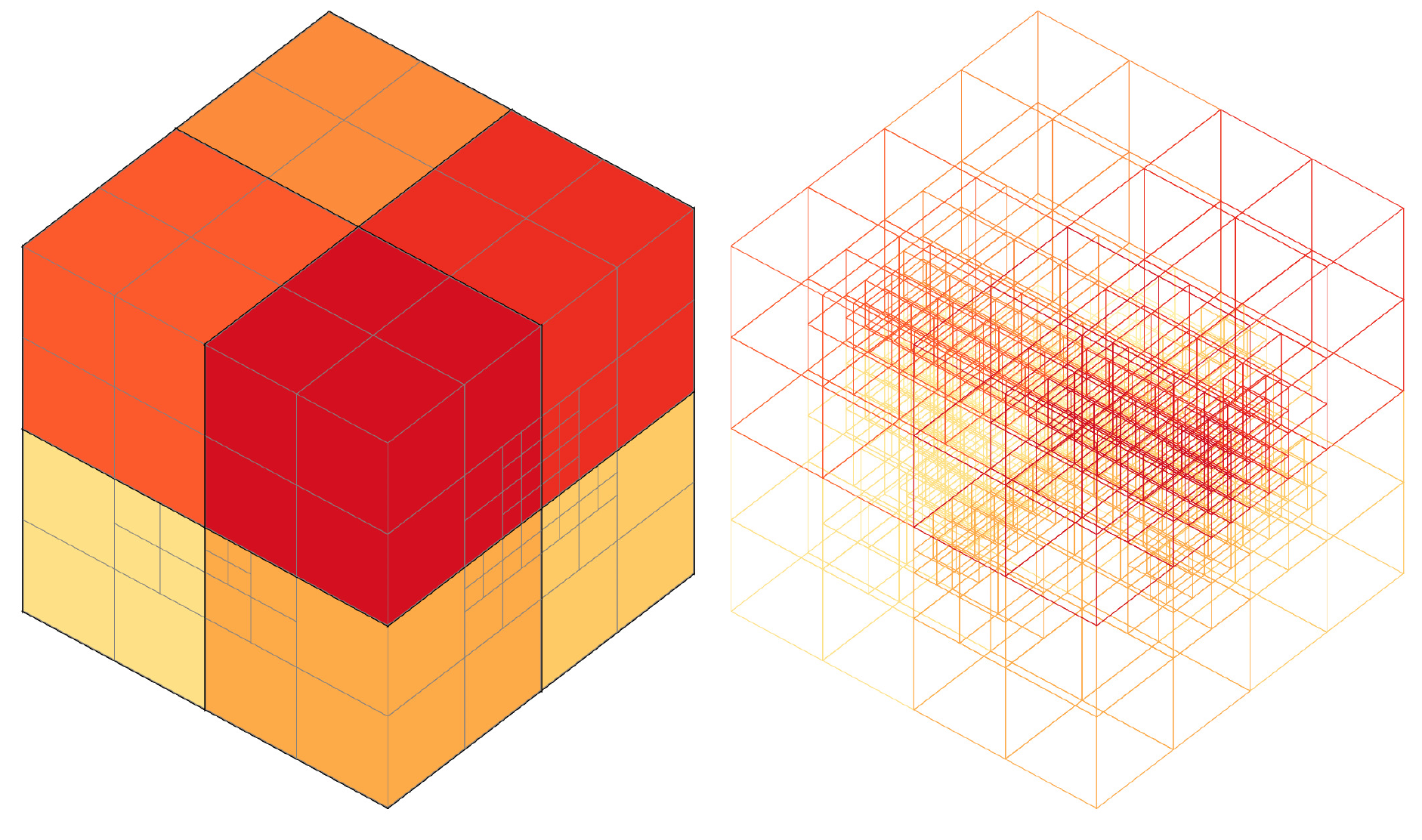}&
		\includegraphics[width=0.5\textwidth]{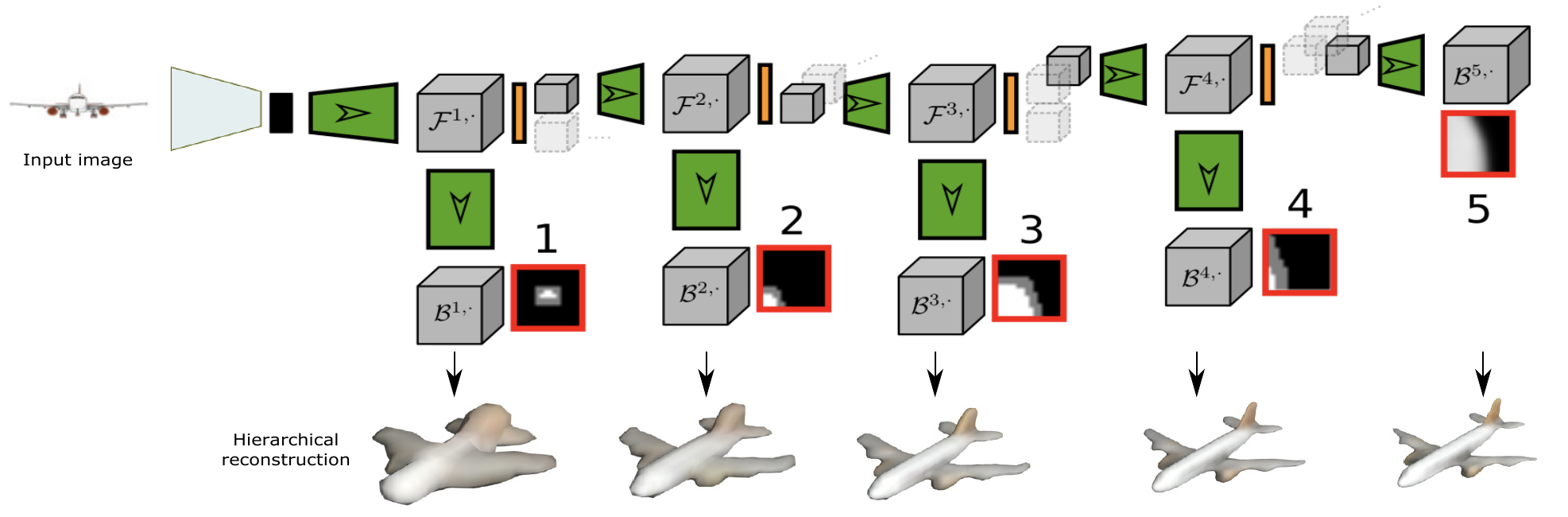}&
		\includegraphics[width=0.24\textwidth]{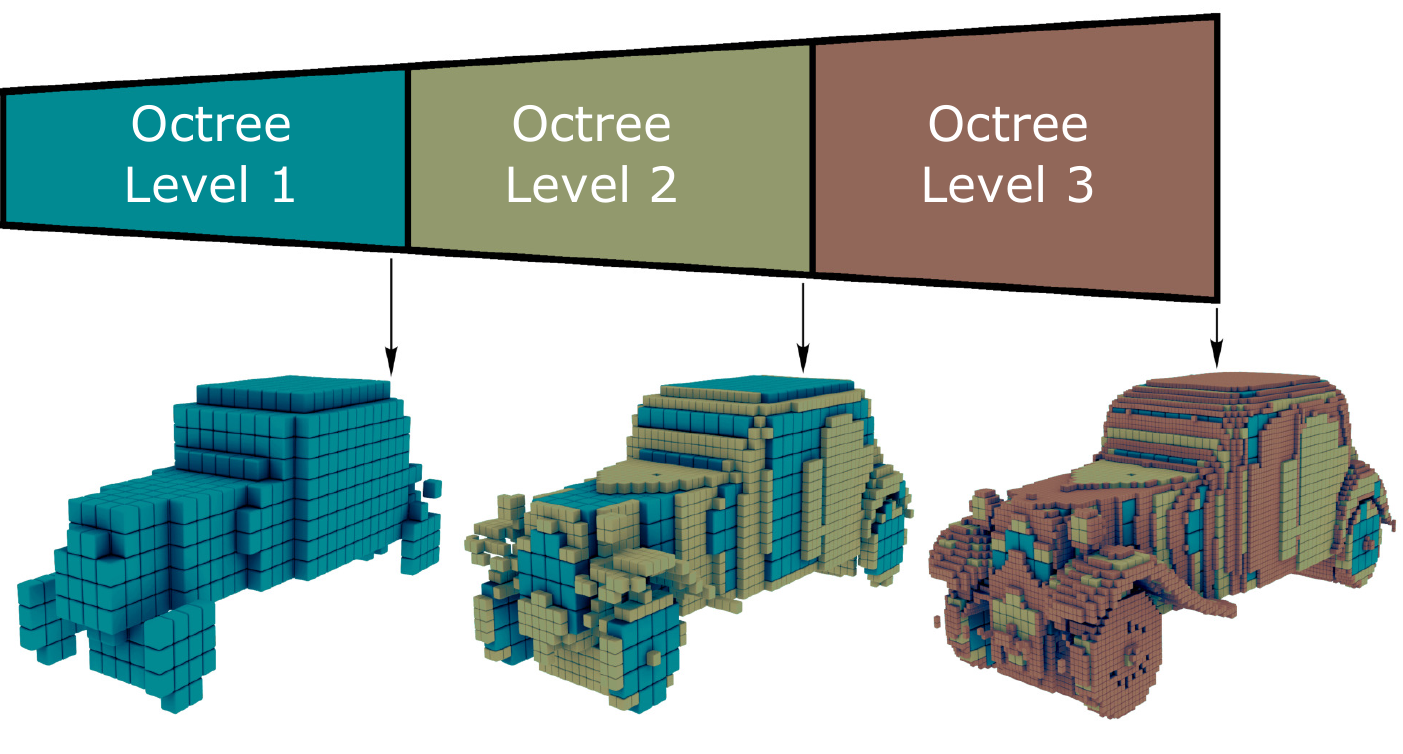} \\
		
		{\fontsize{7}{7} \selectfont (a) Octree Network (OctNet)~\cite{riegler2017octnet}.} & {\fontsize{7}{7} \selectfont  (b) Hierarchical Space Partionning (HSP)~\cite{hane2019hierarchical}.} & {\fontsize{7}{7} \selectfont  (c) Octree Generative Network (OGN)~\cite{tatarchenko2017octree}.} 
		
%
	\end{tabular}
	}
	\caption{\label{fig:volume_partitioning} Space partitioning. OctNet~\cite{riegler2017octnet}  is a  hybrid grid-octree, which enables deep and high-resolution 3D CNNs. High-resolution octrees can also be generated, progressively, in a depth-first~\cite{hane2019hierarchical} or  breadth-first~\cite{tatarchenko2017octree} manner. }
}
\end{figure*}

\subsubsection{Space partitioning} 
\label{sec:space_partitioning}
While regular volumetric grids facilitate convolutional operations, they are very sparse since  surface elements are  contained in few voxels.  Several papers have exploited this sparsity to address the resolution problem~\cite{riegler2017octnet,wang2017cnn,tatarchenko2017octree,li2017grass}. They were able to reconstruct 3D volumetric grids of size  $256^3$ to $512^3$ by using space partitioning techniques such as octrees.  There are, however,  two main challenging issues when using octree structures for deep-learning based reconstruction.  The first  one is computational since convolutional operations are easier to implement (especially on GPUs) when operating on regular grids.  For this purpose, Wang \etal~\cite{wang2017cnn}  designed O-CNN,  a novel octree data structure,  to efficiently store the octant information and CNN features into the graphics memory and execute the entire  training and evaluation on the GPU. O-CNN supports various CNN structures and works with 3D shapes of different representations. By restraining the computations on the octants occupied by 3D surfaces, the memory and computational costs of the O-CNN grow quadratically as the depth of the octree increases, which makes the 3D CNN feasible for high-resolution 3D models.

The second challenge  stems from the fact that the octree structure is object-dependent. Thus, ideally, the deep neural network needs to learn how to infer both the structure of the octree and its content.  In this section, we will discuss how these challenges have been addressed in the literature.


%
\vspace{6pt}		
\paragraph{Using pre-defined octree structures} The simplest approach is to assume that, at runtime, the structure of the octree is known. This is fine for applications such as semantic segmentation where the structure of the output octree can be set to be identical to that of the input.  However, in many important scenarios, \eg 3D reconstruction, shape modeling, and RGB-D fusion, the structure of the octree is not known in advance and must be predicted. To this end, Riegler \etal~\cite{riegler2017octnet} proposed a hybrid grid-octree structure called OctNet (Fig.~\ref{fig:volume_partitioning}-(a)). The key idea is to restrict the maximal depth of an octree to a small number, \eg three, and place several such shallow octrees on a regular grid. 
This representation enables 3D convolutional networks that are both deep and of high resolution. However, at test time,  Riegler \etal~\cite{riegler2017octnet} assume that the structure of the individual octrees is  known. Thus, although the method is able to reconstruct 3D volumes at a resolution of $256^3$, it lacks flexibility since different types of objects may require different training.

\vspace{6pt}
\paragraph{Learning the octree structure }  Ideally, the octree structure and its content should be simultaneously estimated. This can be done as follows; 
\begin{itemize}
	\item First, the input  is encoded into a compact feature vector using a  convolutional encoder (Section~\ref{sec:encoding}). 
	
	\item Next, the feature vector is decoded using a standard up-convolutional network. This results in a coarse volumetric reconstruction of the input, usually of resolution $32^3$ (Section~\ref{sec:lowres_reconstruction_decoding}). 
	
	\item The reconstructed volume, which forms the root of the octree,  is  subdivided into $8$ octants. Octants  with boundary voxels are upsampled and further processed, using an up-convolutional network, to refine the reconstruction of the regions  in that octant. 
	
	\item The octants are processed recursively until the desired resolution is reached.
\end{itemize}

\noi H{\"a}ne \etal~\cite{hane2019hierarchical} introduced  the Hierarchical Surface Prediction (HSP), see Fig.~\ref{fig:volume_partitioning}-(b), which used the approach described above to reconstruct volumetric grids of resolution up to $256^3$. In this approach, the octree is explored in depth-first manner.  Tatarchenko \etal~\cite{tatarchenko2017octree}, on the other hand,  proposed the Octree Generating Networks (OGN), which follows the same idea  but the octree is explored in breadth-first manner, see Fig.~\ref{fig:volume_partitioning}-(c). As such, OGN produces a hierarchical reconstruction of the 3D shape. The approach was  able to reconstruct volumetric grids of size $512^3$. 

Wang \etal~\cite{wang2018adaptive} introduced a patch-guided partitioning strategy. The core idea is to represent a 3D shape with an octree where each of its leaf nodes approximates a planar surface. To infer such structure from a latent representation, Wang \etal~\cite{wang2018adaptive} used a cascade of decoders, one per octree level. At each octree level, a decoder predicts the planar patch within each cell, and a predictor (composed of fully connected layers) predicts the patch approximation status for each octant, \ie whether the cell is "empty", "surface well approximated" with a plane, and "surface poorly approximated". Cells of poorly approximated surface patches are further subdivided and processed by the next level. This approach reduces the memory requirements  from $6.4$GB for volumetric grids of size $256^3$~\cite{wang2017cnn} to $1.7$GB, and the computation time from $1.39$s to $0.30$s, while maintaining the same level of accuracy. Its main limitation is that  adjacent patches are not seamlessly reconstructed. Also, since  a plane is fitted to each octree cell, it does not approximate well curved surfaces.

\subsubsection{Occupancy networks} 
\label{sec:occupancy_networks}
While it is possible to reduce the memory footprint by using various space partitionning techniques,  these approaches lead to complex implementations and existing data-adaptive algorithms are still limited to relatively small  voxel grids ($256^3$ to $512^2$).  Recently, several papers proposed to learn implicit representations of 3D shapes using deep neural networks. For instance, Chen and 
Zhang~\cite{chen2019learning} proposed a decoder  that takes the latent representation of a shape and a 3D point,  and returns a value indicating whether the point is outside or inside the shape. The network can be used to reconstruct high  resolution 3D volumetric representations. However, when retrieving generated shapes, volumetric CNNs only need one shot to obtain the voxel model, while this method needs to pass every point in the voxel grid to the network to obtain its value. Thus, the time required to generate a sample depends on the sampling resolution. 

Tatarchenko \etal~\cite{Tatarchenko_2019_CVPR} introduced occupancy networks that implicitly represent the 3D surface of an object as the continuous decision boundary of a deep neural network classifier.  Instead of predicting a voxelized representation at a fixed resolution, the approach  predicts the complete occupancy function with a neural network that  can be evaluated at any arbitrary resolution. This drastically reduces the memory footprint during training. At inference time, a mesh can be extracted from the learned model using a simple multi-resolution isosurface extraction algorithm.

\subsubsection{Shape partitioning}
\label{sec:shape_partitionning}

Instead of partitionning the volumetric space in which the 3D shapes are embedded,  an alternative approach is to consider the shape as an arrangement of geometric parts,  reconstruct the individual parts independently from each other, and then stitch the parts together to form the complete 3D shape. There has been a few works which attempted this approach. For instance, Li \etal~\cite{li2017grass}  only generate voxel representations at the part level. They proposed a Generative Recursive Autoencoder for Shape Structure (GRASS). The idea is to split the problem into two steps. The first step uses a Recursive Neural Nets (RvNN) encoder-decoder architecture  coupled with a Generative Adversarial Network to learn how to best organize a shape structure into a symmetry hierarchy and  how to synthesize the part arrangements. The second step learns, using another generative model, how to synthesize the geometry of each part, represented as a voxel grid of size $32^3$.  Thus, although the part generator network synthesizes the 3D geometry of parts at only $32^3$ resolution, the fact that individual parts are treated separately enables the reconstruction of 3D shapes at high resolution.

Zou \etal\cite{zou20173d} reconstruct a 3D object as a collection of primitives using a generative recurrent neural network called 3D-PRNN. The architecture transforms the input  into a feature vector of size $32$  via an encoder network. Then, a recurrent generator composed of  stacks of Long Short-Term Memory (LSTM) and a Mixture Density Network (MDN) sequentially predicts from the feature vector the different parts of the shape. At each time step, the network predicts a set of primitives conditioned on both the feature vector and the previously estimated single primitive.   The predicted parts are then combined together to form the reconstruction result.  This approach  predicts only an abstracted representation in the form of cuboids. Coupling it with volumetric-based reconstruction techniques, which would focus on individual cuboids, could lead to a refined 3D reconstruction at the part level.  

\subsubsection{Subspace parameterization}
\label{sec:subspace_parameteirzation}

The space of all possible shapes can be parameterized  using a set of orthogonal basis $\bases = \{\base_1, \dots, \base_{\nbases}\}$. Every shape $\shape$ can then be represented as a linear combination of the bases, \ie $\shape = \sum_{i=1}^{\nbases}\alpha_i \base_i, \text{ with } \alpha_i \in \real$.  This formulation simplifies the reconstruction problem; instead of trying to learn how to reconstruct the volumetric grid $\vgrid$, one can design a decoder composed of fully connected layers to estimate the coefficients $\alpha_i, i=1, \dots, \nbases$ from the latent representation,  and then recover the complete 3D volume.  Johnston \etal~\cite{johnston2017scaling} used the Discrete Cosine Transform-II (DCT-II) to define $\bases$. They then proposed a convolutional encoder to predict the low frequency DCT-II coefficients $\alpha_i$. These coefficients are then converted by a simple Inverse DCT (IDCT) linear transform, which replaces the decoding network, to a solid 3D volume. This  had a profound impact on the computational cost of training and inference: using $n= 20^3$  DCT coefficients, the network is able to reconstruct surfaces at volumetric grids of size $128^3$. 

The main issue when using generic bases such as the DCT bases is that, in general, one requires a large number of basis elements to accurately represent  complex 3D objects. In practice, we usually deal with objects of known categories, \eg human faces and 3D human bodies, and usually, training data is available, see Section~\ref{sec:applications}. As such, one can use Principal Component (PCA) bases, learned from the training data, to parameterize the space of shapes~\cite{kundu20183d}.  This would require a significantly smaller number of bases (in the order of $10$) compared to the number of generic basis, which is in the order of thousands. 

\subsubsection{Coarse-to-fine refinement}
\label{sec:coarse_to_fine_refinement}
Another way to improve the resolution  of volumetric techniques is by using multi-staged approaches~\cite{dai2017shape,han2017high,wang2017shape,yang2018dense,Cao_2018_ECCV}. The  first stage recovers a low resolution voxel grid, say $32^3$, using an encoder-decoder architecture. The subsequent stages, which function as upsampling networks,   refine the reconstruction by focusing on local regions.  Yang \etal~\cite{yang2018dense} used an up-sampling module which simply consists of two up-convolutional layers. This simple up-sampling module  upgrades the output 3D shape to a higher resolution of $256^3$. 

Wang \etal~\cite{wang2017shape}  treat the reconstructed coarse voxel grid as a sequence of images (or slices). The 3D object is then reconstructed  slice by slice at high resolution.  While this approach allows efficient refinement using 2D up-convolutions, the 3D shapes  used for training  should be consistently aligned so that the volumes can be sliced along the first principal direction. Also,  reconstructing individual slices independently from each other may result in discontinuities and incoherences in the final volume.  To capture the dependencies between the slices, Wang \etal~\cite{wang2017shape} use a Long term Recurrent Convolutional Network (LRCN)~\cite{donahue2015long} composed of a 3D encoder, an LSTM unit, and a 2D decoder.  At each time, the 3D encoder processes five  consecutive slices to produce a fixed-length vector representation as input to the LSTM. The output  of the LSTM is passed to the 2D convolutional decoder  to produce a high resolution image.  The concatenation of the high-resolution 2D images forms the high-resolution output 3D volume.

Instead of using volume slicing, other papers used additional CNN modules, which focus on  regions that require refinement. For example,  Dai \etal~\cite{dai2017shape} firstly predict a coarse but complete shape volume of size $32^3$ and then refine it into a $128^3$  grid via an iterative volumetric patch synthesis process, which copy-pastes voxels from the k-nearest-neighbors retrieved from a database of 3D models.        Han \etal~\cite{han2017high}  extend Dai \etal's approach by introducing a local 3D CNN to perform patch-level surface refinement.   Cao \etal~\cite{Cao_2018_ECCV}, which recover in the first stage a volumetric grid of size $128^3$,   take volumetric blocks of size $16^3$ and predict whether they require further refinement. Blocks that require refinement are resampled into $512^3$ and  fed into another encoder-decoder for refinement, along with the initial coarse prediction to guide the refinement. Both subnetworks adopt the U-net architecture~\cite{ronneberger2015u} while substituting convolution and pooling layers with the corresponding operations from OctNet~\cite{riegler2017octnet}. 

Note that these methods need separate and sometimes  time-consuming steps before local inference. For example, Dai \etal~\cite{dai2017shape} require nearest neighbor searches from a 3D database. Han \etal~\cite{han2017high} require  3D boundary detection while Cao \etal~\cite{Cao_2018_ECCV} require assessing whether a block requires further refinement or not.

\subsection{Deep marching cubes}

While volumetric   representations can handle 3D shapes of arbitrary topologies,  they require a post processing step, \eg   marching cubes~\cite{lorensen1987marching},   to retrieve the actual 3D surface mesh, which is the quantity of interest in 3D reconstruction.  As such, the whole pipeline cannot be trained end-to-end.  To overcome this limitation, Liao \etal~\cite{liao2018deep} introduced the Deep Marching Cubes, an end-to-end trainable network, which predicts explicit surface representations of arbitrary topology.  They  use a modified differentiable representation, which separates the mesh topology from the geometry. The network is composed of an encoder and  a two-branch decoder. Instead of predicting signed distance values, the first branch  predicts the probability of occupancy for each voxel. The mesh topology is then implicitly (and probabilistically) defined by the state of the occupancy variables at its corners. The second branch of the decoder predicts a vertex location for every edge of each cell. The combination of both implicitly-defined topology and vertex location defines a distribution over meshes that is differentiable and can be used for back propagation.  While the approach is trainable end-to-end, it is limited to low resolution grids of size $32^3$.

Instead of directly estimating high resolution volumetric grids, some methods produce multiview depth maps, which are  fused  into an output volume. The main advantage   is that, in the decoding stage,  one can use 2D convolutions, which are more efficient, in terms of computation and memory storage, than 3D convolutions. Their main limitation, however, is that depth maps only encode the external surface. To capture internal structures, Richter \etal~\cite{Richter_2018_CVPR} introduced Matryoshka Networks, which use $L$ nested depth layers; the shape is recursively reconstructed by first fusing   the depth maps in the first layer, then  subtracting shapes in even layers, and adding shapes in odd layers. The method is able to reconstruct volumetric grids of size $256^3$.

\section{3D surface decoding }   
\label{sec:3D_surface_decoding}

Volumetric representation-based  methods are computationally very wasteful since information is rich only on or near the surfaces of 3D shapes. The main challenge when working directly with surfaces is that  common  representations such as meshes or point clouds are not regularly structured and thus, they do not easily fit into deep learning architectures, especially those using CNNs. This section reviews the techniques used to address this problem. We classify the state-of-the-art into three main categories:  parameterization-based (Section~\ref{sec:parameterization_surface}),  template deformation-based (Section~\ref{sec:deformationbased}), and point-based methods (Section~\ref{sec:point_based}). 


\begin{table}
\label{tab:volumetric_decoder_taxonomy}
\centering{

\caption{\small{Taxonomy of mesh decoders.  GCNN: graph CNN. MLP:  Multilayer Perceptron. Param.: parameterization.}
}

\resizebox{\linewidth}{!}
{%
	\begin{tabular}{?@{ }c|c|c|c|c?}
	\noalign{\hrule height 1pt}
	
	\multirow{4}{*}{\textbf{} }& \multirow{2}{*}{\textbf{Param.-based}} & \multicolumn{2}{c|}{\textbf{Deformation-based}} & \multirow{2}{1.9cm}{\textbf{Decoder architecture} } \\
		\cline{3-4} 
							& 									& Defo. model & Template & \\
	\cline{2-5}
							& Geometry Images & Vertex defo. & Sphere / ellipse & FC layers \\
							& Spherical maps & Morphable  &(k-)NN& UpConv \\
							& Patch-based &FFD & Learned (PCA) & \\
							&			& 	& Learned (CNN) & \\
	\noalign{\hrule height 1pt}
	\cite{sinha2017surfnet}  & Geometry Image & $-$ & $-$& UpConv \\
	
	\hline
	\cite{Pumarola_2018_CVPR}  & Geometry Image & $-$& $-$&   \multirow{2}{2.2cm}{ResNet blocks + 2 Conv layers}\\
									       & 			& 	&		& \\
	\hline
	\cite{groueix2018atlasNet}  & Patch-based &$-$ & $-$& MLP \\
	
	\hline
	\cite{kato2018neural}  & Mesh & vertex defo.&  sphere& FC\\
	
	\hline
	\cite{Wang_2018_ECCV} & Mesh& vertex defo. & ellipse&  GCNN blocks\\
	
	\hline
	\cite{henderson2018learning}   & Mesh & vertex & cube &  UpConv\\

%
	
	\hline
	\cite{kanazawa2018learning}  &Mesh & vertex defo. & Learned (CNN)& FC layer \\
	
	\hline
	\cite{jack2018learning}  & Mesh&FFD  & $k$-NN& FC \\
	
	\hline
	  \cite{kurenkov2017deformnet} & Mesh & FFD & NN& UpConv\\
	
	\hline
	 \cite{pontes2017image2mesh} & Mesh & FFD & $k$-NN& Feed-forward \\
	 
%
	\noalign{\hrule height 1pt}
	\end{tabular}
}
}
\end{table}

\subsection{Parameterization-based 3D reconstruction}
\label{sec:parameterization_surface}

Instead of working directly with triangular meshes, we can represent the surface of a 3D shape $\shape$ as a mapping $\shapefunc: \domain \to \rthree$ where $\domain$ is a regular parameterization domain.  The goal of the 3D reconstruction process is then to recover the shape function $\shapefunc$ from an input $\images$.  When $\domain$ is a 3D domain then the methods in this class fall within the volumetric techniques described in Section~\ref{sec:volumetric_3D_reconstruction}. Here, we focus on the case where $\domain$ is a regular 2D domain, which can be a subset of the two dimensional plane, \eg $\domain = [0, 1]^2$, or the unit sphere, \ie $\domain = \stwo$.  In the first case,  one can implement encoder-decoder architectures using standard 2D convolution operations. In the latter case, one has to use spherical convolutions~\cite{monti2017geometric} since the domain is spherical.


Spherical parameterizations and geometry images~\cite{gotsman2003fundamentals,praun2003spherical,sheffer2007mesh}  are the most commonly used parameterizations. They are, however, suitable only for genus-0 and disk-like surfaces.  Surfaces of arbitrary topology need to be cut into disk-like patches, and then unfolded  into a regular 2D domain.   Finding the optimal cut for a given surface, and more importantly, findings cuts that are consistent across shapes within the same category is challenging. In fact,  naively creating independent geometry images for a shape category and feeding them into deep neural networks would fail to generate coherent 3D shape surfaces~\cite{sinha2017surfnet}.  

To create, for genus-0 surfaces,  robust geometry images that are consistent across a shape category, the 3D objects within the  category should be first put in correspondence~\cite{laga2017numerical,wang2018shape,wang2018statistical}. Sinha \etal~\cite{sinha2017surfnet}  proposed a cut-invariant procedure, which  solves a large-scale correspondence problem, and an extension of deep residual nets to automatically generate geometry images encoding the $x, y, z$ surface coordinates. The approach uses three separate encoder-decoder networks, which learn, respectively, the  $x, y$ and $z$ geometry images.  The three networks are  composed of  standard convolutions, up-residual, and down-residual blocks. They take as input  a depth image or a RGB image, and learn the 3D reconstruction  by minimizing a shape-aware $\ltwo$ loss function.  


Pumarola \etal~\cite{Pumarola_2018_CVPR}   reconstruct the shape of a deformable surface using a network which has two  branches: a detection branch and a depth estimation branch, which operate in parallel, and a third shape branch, which  merges the detection mask and the depth map into a  parameterized surface.  Groueix \etal~\cite{groueix2018atlasNet} decompose the surface of a 3D object into $m$ patches, each patch $i$  is defined as a mapping $\shapefunc_i: \domain = [0, 1]^2 \mapsto \rthree$.  They have then designed a decoder which is composed of $m$ branches. Each branch $i$ reconstructs the $i-$th patch  by estimating the function $\shapefunc_i$. At the end, the reconstructed patches are merged together to form the entire surface. Although this approach can handle surfaces of high genus,  it is still not general enough to handle surfaces of arbitrary genus. In fact,  the optimal number of patches depends on the genus of the surface ($n=1$ for genus-0, $n=2$ for genus-1, etc.). Also, the patches are not guaranteed to be connected, although in practice one can still post-process the result and fill in the gaps between disconnected patches. 

In summary, parameterization methods are limited to low-genus surfaces. As such, they are suitable for the reconstruction of objects that belong to a given shape category, \eg human faces and bodies. 

\subsection{Deformation-based 3D reconstruction}
\label{sec:deformationbased}

Methods in this class take an input $\images$ and estimate a deformation field $\deformationfield$, which, when applied to a template 3D shape, results in the reconstructed 3D model $\shape$.  Existing techniques differ in the type of deformation models they use (Section~\ref{sec:deformation_model}),  the way the template is defined (Section~\ref{sec:template}), and in the network architecture used to estimate the deformation field $\deformationfield$ (Section~\ref{sec:architectures_deformation}).  In what follows, we assume that a 3D shape $\shape = (\vertices, \faces)$ is represented with  $n$ vertices $\vertices = \{\vertex_1, \dots, \vertex_n \}$ and faces $\faces$. Let $\templateshape = (\tilde{\vertices}, \faces)$ denote a template shape.

\subsubsection{Deformation models}
\label{sec:deformation_model}
\begin{figure}[t]
\centering{
	\begin{tabular}{@{}c@{}c@{}}
		\includegraphics[width=0.5\textwidth]{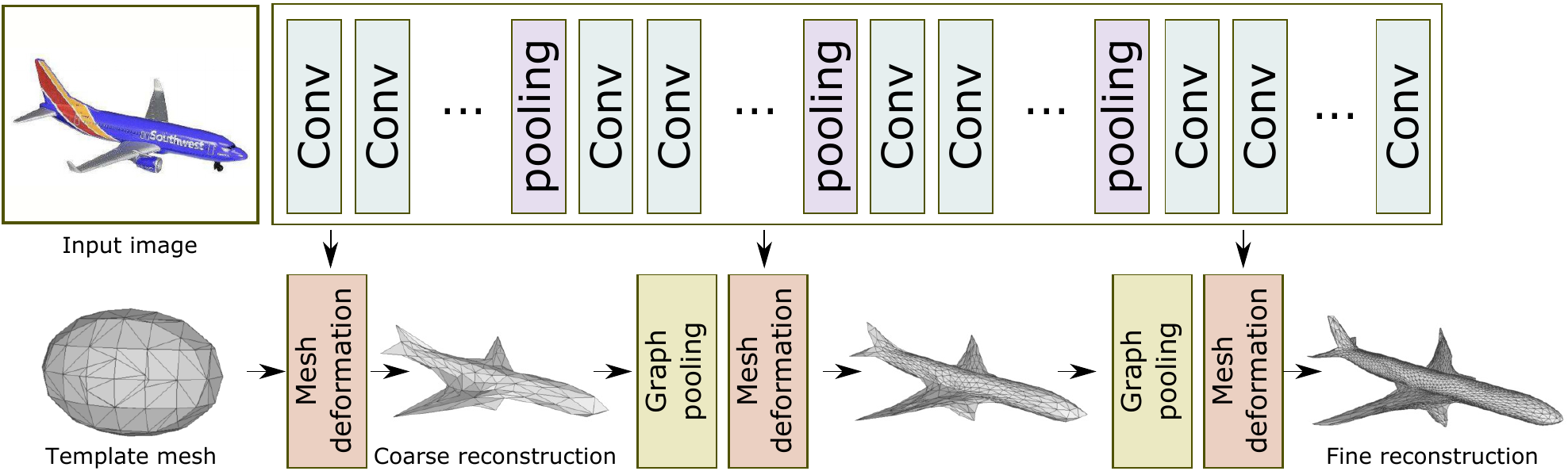} \\
		 \includegraphics[width=0.3\textwidth]{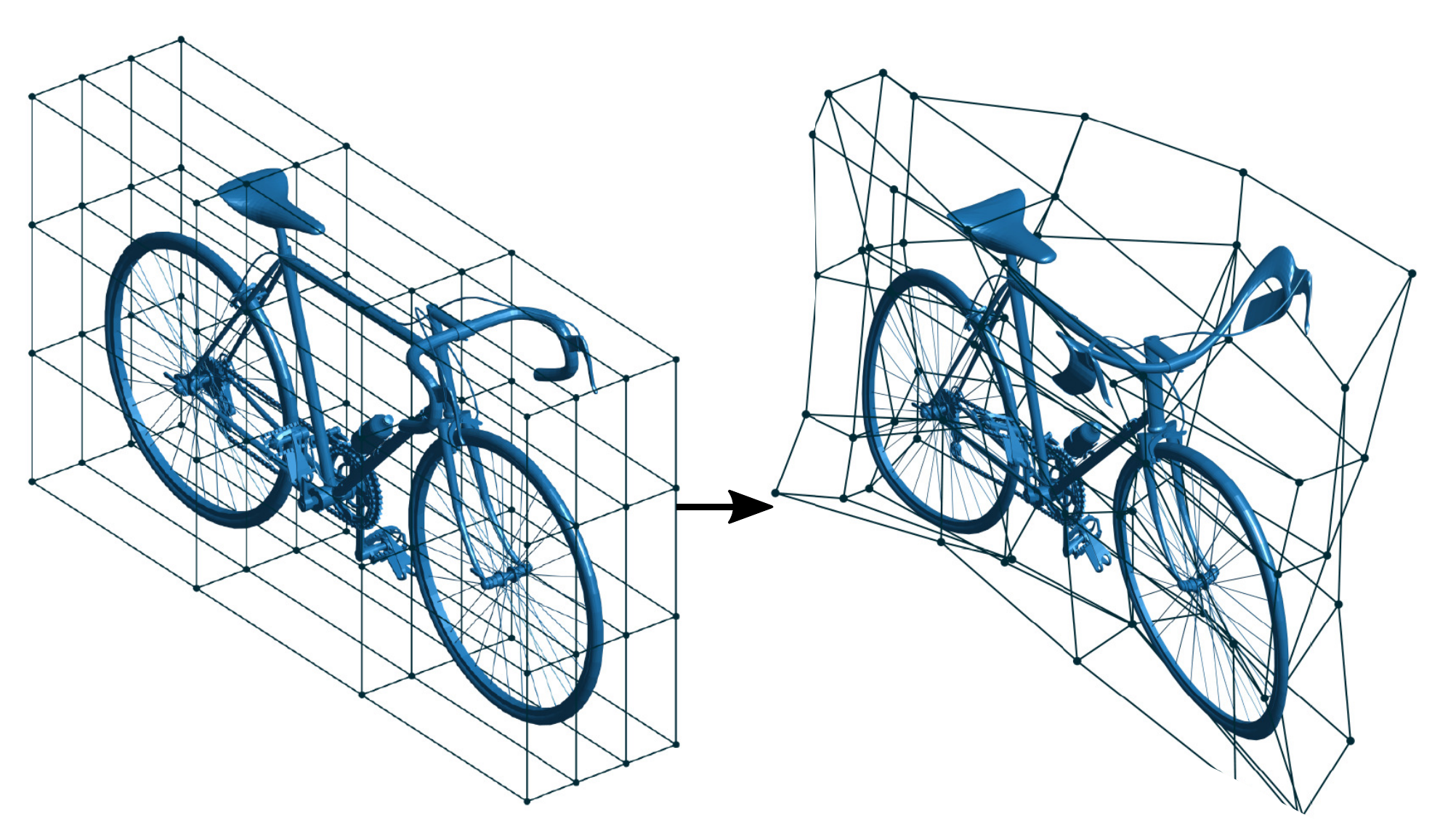}
	\end{tabular}
	\caption{\label{fig:defo_models}Template deformation (top)~\cite{Wang_2018_ECCV} vs. domain deformation (bottom)~\cite{pontes2017image2mesh}.}
}
\end{figure}

\vspace{6pt}
\noi\textit{(1) Vertex deformation. } This model  assumes that a 3D shape $\shape$ can be written in terms of linear displacements of the individual vertices of the template, \ie  $\forall \vertex_i \in \vertices, \vertex_i = \tilde{\vertex}_i + \deformation_i$, where $\deformation_i \in \rthree$. The deformation field is defined as $\deformationfield = (\deformation_1, \dots, \deformation_n)$. This deformation model, illustrated in Fig.~\ref{fig:defo_models}-(top),  has been used in)~\cite{Wang_2018_ECCV,kato2018neural,kanazawa2018learning}. It assumes that \textbf{(1)} there is a one-to-one correspondence between the vertices of the shape $\shape$ and those of the template $\templateshape$, and \textbf{(2)} the shape $\shape$ has the same topology as the template $\templateshape$.


\vspace{6pt}
\noi\textit{(2) Morphable models. } Instead of using a generic template, one can use learned morphable models~\cite{blanz1999morphable} to parameterize a 3D mesh. Let $\tilde{\vertices}$ be the mean shape and $\eigenvector_1, \dots, \eigenvector_K$ be a set of orthonormal basis. Any shape $\vertices$  can be written in the form:
\begin{equation}
\vertices = \tilde{\vertices} + \sum_{i=1}^{K} \alpha_i \eigenvector_i, \alpha_i \in \real. 
	\label{eq:pca}
\end{equation}

\noi The second term of Equation~\eqref{eq:pca} can be  seen as a deformation field,  $\deformationfield = \sum_{i=1}^{K} \alpha_i \eigenvector_i$, applied to the vertices $\tilde{\vertices}$ of the mean shape. By setting $ \eigenvector_0 = \tilde{\vertices} \text{ and } \alpha_0 = 1$, Equation~\eqref{eq:pca} can be written as $\vertices = \sum_{i=0}^{K} \alpha_i \eigenvector_i$. In this case, the mean $\tilde{\vertices}$ is treated as a bias term.

One approach to learning a morphable model  is by using Principal Component Analysis (PCA) on a collection of  clean 3D mesh exemplars~\cite{blanz1999morphable}. Recent techniques showed that, with only 2D annotations,  it is possible to build category-specific 3D morphable models from 2D silhouettes or 2D images~\cite{vicente2014reconstructing,tulsiani2017learning}. These methods require efficient detection and segmentation of the objects, and camera pose estimation, which can also be done using CNN-based techniques. 

\vspace{6pt}
\noi\textit{(3) Free-Form Deformation (FFD). } Instead of directly deforming  the  vertices of the template $\templateshape$, one can deform the space around it, see Fig.~\ref{fig:defo_models}-(bottom). This can be done by defining around $\templateshape$  a set  $P \in \real^{m\times 3}$ of $m$ control points, called deformation handles. When the deformation field  $\deformationfield = (\deformation_1, \dots, \deformation_m), m \ll n$, is applied to these control points, they deform the entire space around the shape and thus, they also deform the vertices $\vertices$ of the shape  according to the following equation:
\begin{equation}
\small{
	\vertices^\top =  B \Phi  (P + \deformationfield)^\top,
	}
	\label{eq:ffd}
\end{equation}
where the deformation matrix $B \in \real^{n\times m}$ is a set of polynomial basis, \eg the Bernstein polynomials~\cite{pontes2017image2mesh}.  $\Phi$ is a $m\times m$ matrix used to impose symmetry in the FFD field, see~\cite{pontes2017compact}, and  $\deformationfield$ is the displacements.

This approach has been used by  Kuryenkov \etal~\cite{kurenkov2017deformnet},  Pontes \etal~\cite{pontes2017image2mesh}, and Jack \etal~\cite{jack2018learning}. The main advantage of free-form deformation is that it does not require one-to-one correspondence between the shapes  and the template. However, the shapes that can be approximated by the FFD of the template are only those that have the same topology as the template.

\subsubsection{Defining the template}
\label{sec:template}
Kato \etal~\cite{kato2018neural} used a sphere as a template. Wang \etal~\cite{Wang_2018_ECCV}  used an ellipse. Henderson \etal~\cite{henderson2018learning} defined two types of templates: a complex shape abstracted into cuboidal primitives,  and a cube subdivided into multiple vertices. While the former is suitable for  man-made shapes that have multiple components, the latter is suitable for representing genus-0 shapes and does not offer advantage compared to using a sphere or an ellipsoid. 

To speed up the convergence, Kuryenkov \etal~\cite{kurenkov2017deformnet}  introduced DeformNet,  which takes an image as input, searches the nearest shape  from a database, and    then deforms,  using the FFD model of Equation~\eqref{eq:ffd}, the retrieved model  to match the query image.  This method allows detail-preserving 3D reconstruction.

Pontes \etal~\cite{pontes2017image2mesh} used an approach that is similar to DeformNet~\cite{kurenkov2017deformnet}. However, once the FFD field  is estimated and applied to the template, the result is further refined  by adding a residual defined as a weighted sum of some 3D models retrieved from a dictionary.  The role of the deep neural network is to learn how to estimate the deformation field $\deformationfield$ and the weights used in computing the refinement residual.   Jack \etal~\cite{jack2018learning}, on the other hand, deform, using FFD,  multiple templates and select the one that provides the best fitting accuracy. 

Another approach is to learn the template, either separately using statistical shape analysis techniques, \eg PCA, on a set of training data, or jointly with the deformation field using deep learning techniques. For instance, Tulsiani \etal~\cite{tulsiani2017learning} use the mean shape of each category of 3D models as a class-specific template. The deep neural network  estimates both the class of the input shape, which is used to select the class-specific mean shape, and the deformation field that needs to be applied to the class-specific mean shape. Kanazawa \etal~\cite{kanazawa2018learning}  learn, at the same time, the mean shape and the deformation field. Thus, the approach does not require a separate 3D training set to learn the morphable model.  In both cases,  the reconstruction results lack details and are limited to  popular categories such as cars and birds. 

\subsubsection{Network architectures}
\label{sec:architectures_deformation}
Deformation-based methods also use encoder-decoder architectures. The encoder maps  the input into a latent variable $\featurevector$ using successive convolutional operations. The latent space can be discrete or continuous as in~\cite{henderson2018learning}, which used a variational auto-encoder (see Section~\ref{sec:encoding}). The decoder is, in general, composed of fully-connected layers.  Kato \etal~\cite{kato2018neural}, for example, used two fully connected layers to estimate the deformation field to apply to a sphere to match the input's silhouette. 

Instead of deforming a sphere or an ellipse,  Kuryenkov \etal~\cite{kurenkov2017deformnet} retrieve from a database the 3D model that is most similar to the input $\images$ and then estimate the FFD needed to deform it to match the input.  The retrieved template is first voxelized and encoded, using a 3D CNN, into another latent variable $\featurevector_t$. The latent representation of the input image  and the latent representation of the retrieved template are then concatenated and  decoded, using  an up-convolutional network, into an FFD field defined on the vertices of a voxel grid.

Pontes \etal~\cite{pontes2017image2mesh} used  a similar approach, but the latent variable $\featurevector$ is used as input into a classifier which finds, from a database, the closest model to the input. At the same time, the latent variable is decoded, using a feed-forward network, into a deformation field $\deformationfield$ and weights $\alpha_i, i=1, \dots, K$. The retrieved template is then deformed using   $\deformationfield$ and a weighted combination of a dictionary of CAD models, using the weights $\alpha_i$. 

Note that, one can design several variants to these approaches. For instance, instead of using a 3D model retrieved from a database as a template, one can use a class-specific mean shape. In this case, the latent variable $\featurevector$ can be used to classify the input into one of the shape categories, and then pick the learned mean shape of this category as a template~\cite{tulsiani2017learning}. Also, instead of learning separately the mean shape, \eg using  morphable models,   Kanazawa \etal~\cite{kanazawa2018learning} treated the mean shape as a bias term, which can then be predicted by the network, along with the deformation field $\deformationfield$.  Finally, Wang \etal~\cite{Wang_2018_ECCV} adopted a coarse to fine strategy, which makes the procedure more stable.  They proposed a deformation network composed of three deformation blocks, each block is a graph-based CNN (GCNN),  intersected by two graph unpooling layers. The deformation blocks update the location of the vertices while the graph unpoolling layers increase the number of vertices.

Parameterization and deformation-based techniques can only reconstruct surfaces of fixed topology. The former is limited to surfaces of low genus while the latter is limited to the topology of the template.

\subsection{Point-based techniques}
\label{sec:point_based}

A 3D shape can be represented using an unordered  set $\pointset = \{(x_i, y_i, z_i) \}_{i=1}^{N}$  of $N$ points. Such point-based representation  is simple but efficient in terms of memory requirements. It is well suited for  objects with intriguing parts and fine details.  As such, an increasing number of papers, at least one in $2017$~\cite{fan2017point}, more than $12$ in $2018$~\cite{lin2018learning,Jiang_2018_ECCV,mandikal20183d,gadelha2018multiresolution,li2018point,sun2018pointgrow,mandikal20183d,insafutdinov2018unsupervised,li2018efficient,li2018optimizable,zeng2018inferring,mandikal2019dense,wang2018mvpnet}, and a few others in 2019~\cite{mandikal2019dense},   explored their usage for deep learning-based reconstruction. This section  discusses the state-of-the-art point-based representations and their corresponding network architectures. 

\subsubsection{Representations}

The main challenge with point clouds is that they are not regular structures and do not easily fit into the convolutional  architectures that exploit the spatial regularity.  Three  representations have been proposed to overcome this limitation:
\begin{itemize}
	\item Point set representation  treats a point cloud as a matrix of size $N \times 3$~\cite{fan2017point,mandikal20183d,gadelha2018multiresolution,li2018point,insafutdinov2018unsupervised,mandikal2019dense}.  
		
	\item One  or multiple  3-channel grids of size $\height \times \width \times 3$~\cite{fan2017point,wang2018mvpnet,lin2018learning}. Each pixel in a grid encodes the $(x, y, z)$ coordinates of a 3D point.  
	
	\item Depth maps  from multiple viewpoints~\cite{tatarchenko2016multi,li2018efficient}. 
\end{itemize}

\noi The last two representations, hereinafter referred to as grid representations, are well suited for convolutional networks. They are also computationally efficient as they can be inferred using only 2D convolutions. Note that depth map-based methods require an additional fusion step to infer the entire 3D shape of an object. This can be done in a straightforward manner if the camera parameters are known. Otherwise, the fusion can be done using point cloud registration techniques~\cite{besl1992method,chen1992object}  or  fusion networks~\cite{xie2019pix2vox}.  Also, point set representations  require fixing in advance the number of points $N$ while  in methods that use grid representations, the number of points can vary based on the nature of the object but it is always bounded by the grid resolution. 

\subsubsection{Network architectures}
\begin{figure}
	\begin{tabular}[h]{@{}l@{}}
	
	\includegraphics[width=0.48\textwidth]{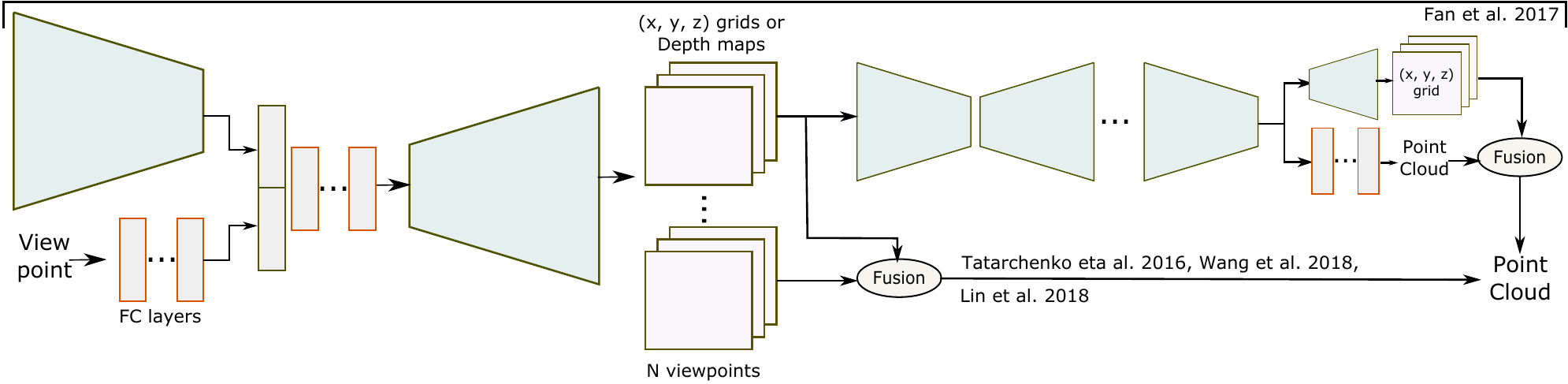} \\
	{\fontsize{7}{7} \selectfont (a) Fan \etal~\cite{fan2017point},  Tatarchenko \etal~\cite{tatarchenko2016multi}, Wang \etal~\cite{wang2018mvpnet}, and Lin \etal~\cite{lin2018learning}.} \\
	\\
	\includegraphics[width=0.48\textwidth]{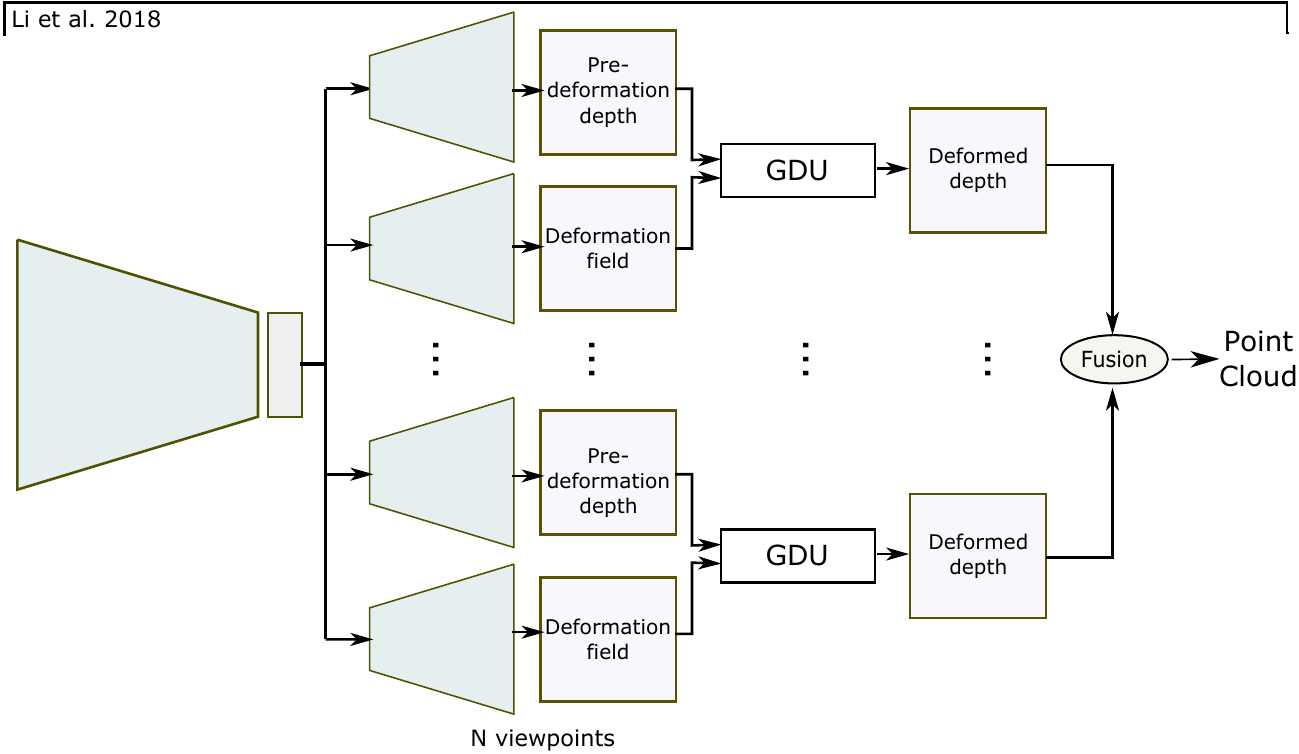} \\
	{\fontsize{7}{7} \selectfont (b) Li \etal~\cite{li2018efficient}.}\\
	\\
	\includegraphics[width=0.48\textwidth]{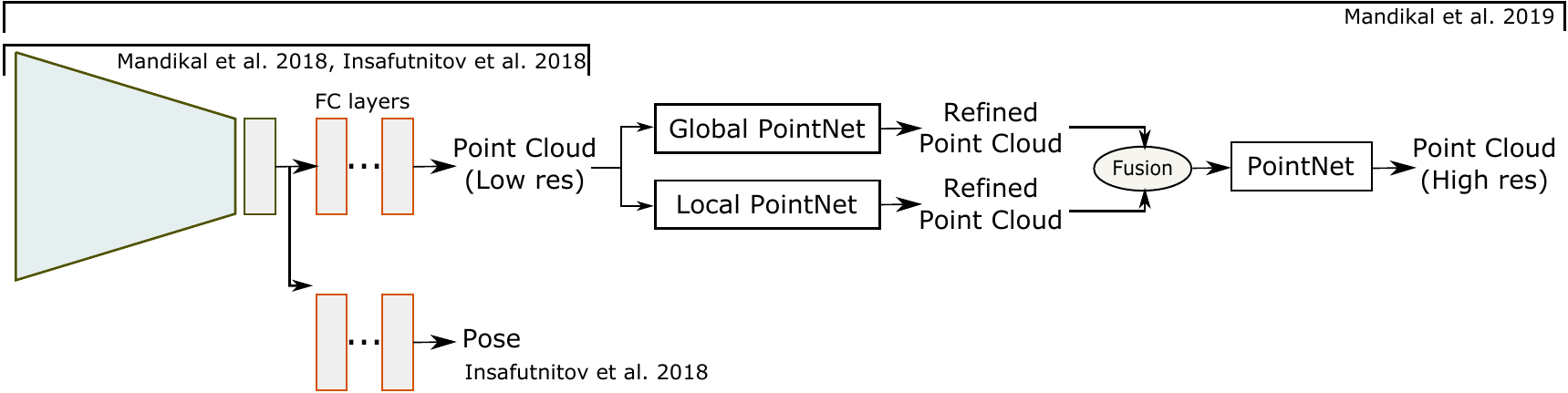} \\
	{\fontsize{7}{7} \selectfont (c)   Mandikal \etal~\cite{mandikal20183d}, Insafutdinov~\cite{insafutdinov2018unsupervised}, and Mandikal \etal~\cite{mandikal2019dense}.}\\
	\\
	\includegraphics[width=0.48\textwidth]{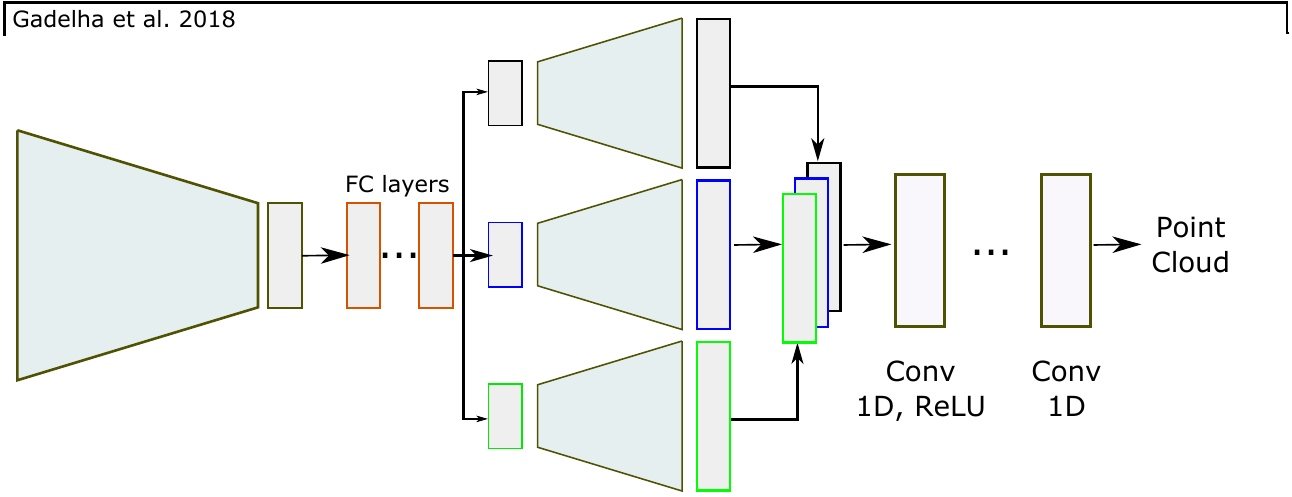}\\
	 {\fontsize{7}{7} \selectfont (d) Gadelha \etal~\cite{gadelha2018multiresolution}. }
	\end{tabular}
	\caption{\label{fig:architectures_pointbased}The different network architectures used in point-based 3D reconstruction. }
\end{figure}

Similar to volumetric and surface-based representations, techniques that use point-based representations follow the encoder-decoder model. While they all use the same architecture for the encoder, they differ in the type and architecture of their decoder, see Fig.~\ref{fig:architectures_pointbased}.  In general, grid representations use up-convolutional networks to decode the latent variable~\cite{fan2017point,wang2018mvpnet,lin2018learning,li2018efficient}, see Fig.~\ref{fig:architectures_pointbased}-(a) and (b).  Point set representations (Fig.~\ref{fig:architectures_pointbased}-(c)) use fully connected layers~\cite{fan2017point,Jiang_2018_ECCV,mandikal20183d,insafutdinov2018unsupervised,mandikal2019dense} since point clouds are unordered. The main advantage of fully-connected layers is that they capture the global information. However, compared to convolutional operations, they are computationally expensive. To benefit from the efficiency of convolutional operations,  Gadelha \etal~\cite{gadelha2018multiresolution} order, spatially, the point cloud using a space-partitionning tree such as KD-tree and then process them using 1D convolutional operations, see Fig.~\ref{fig:architectures_pointbased}-(d). With a conventional CNN, each convolutional operation has a restricted receptive field and is not able to leverage both global and local information effectively. Gadelha \etal~\cite{gadelha2018multiresolution} resolve this issue by maintaining three different resolutions. That is, the latent variable is decoded into three different resolutions, which are then concatenated and further processed with 1D convolutional layers to generate a point cloud of size $4$K.

Fan \etal~\cite{fan2017point}  proposed  a generative deep network that combines both the point set representation and the grid representation (Fig.~\ref{fig:architectures_pointbased}-(a)). The network is composed of a cascade of encoder-decoder blocks:
\begin{itemize}
	\item  The first block takes the input image and maps it into a latent representation, which is then decoded into a 3-channel image of size $\height \times \width$. The three values at each pixel are the coordinates of a point. 
	
	\item Each of the subsequent blocks takes the output of its previous block and further encodes and decodes it into a 3-channel image of size $\height \times \width$.
	
	\item The last block is an encoder, of the same type as the previous ones, followed by a predictor composed of two branches. The first branch is a decoder  which predicts a 3-channel image of size $\height \times \width$ ($32\times 24$ in this case), of which the three values at each pixel are the coordinates of a point. The second branch is a fully-connected network, which predicts a matrix of size $N\times 3$, each row is a 3D point ($N=256$).
	
	\item The predictions of the two branches are merged using set union to produce a 3D point set of size $1024$.
\end{itemize}

\noi This approach has been also used by Jiang \etal~\cite{Jiang_2018_ECCV}. The main difference between the two is in the training procedure, which we will discuss in Section~\ref{sec:training_deeplearning}.


Tatarchenki \etal~\cite{tatarchenko2016multi},  Wang \etal~\cite{wang2018mvpnet}, and Lin \etal~\cite{lin2018learning} followed the same idea but their   decoder regresses $N$ grids, see Fig.~\ref{fig:architectures_pointbased}-(a). Each grid encodes the depth map~\cite{tatarchenko2016multi} or the $(x, y, z)$ coordinates~\cite{wang2018mvpnet,lin2018learning} of the visible surface from that view point. The viewpoint, encoded with a sequence of fully connected layers,  is provided as input to the decoder along with the latent representation of the input image. Li \etal~\cite{li2018efficient}, on the other hand, used a multi-branch decoder, one for each viewpoint, see Fig.~\ref{fig:architectures_pointbased}-(b). Unlike~\cite{tatarchenko2016multi}, each branch  regresses a canonical depth map from a given view point and a deformation field, which deforms the estimated canonical depth map to match the input, using Grid Deformation Units (GDUs). The reconstructed grids are then lifted to 3D and merged together. 



Similar to volumetric techniques, the vanilla architecture for point-based 3D reconstruction only recovers low resolution geometry.  For high-resolution reconstruction, Mandikal \etal~\cite{mandikal2019dense}, see Fig.~\ref{fig:architectures_pointbased}-(c), use a cascade of multiple networks. The first  network  predicts a low resolution point cloud.  Each subsequent block  takes the previously predicted point cloud, computes global features, using a multi-layer perceptron architecture (MLP) similar to PointNet~\cite{qi2017pointnet} or Pointnet++~\cite{qi2017pointnetpp}, and local features by applying MLPs in balls around each point. Local and global features are then aggregated and fed to another MLP, which predicts a dense point cloud. The process can be repeated recursively until the desired resolution is reached.


Mandikal \etal~\cite{mandikal20183d}  combine TL-embedding with a variational auto-encoder (Fig.~\ref{fig:architectures_pointbased}-(c)). The former allows mapping a 3D  point cloud and its corresponding views onto the same location in the latent space. The latter enables the reconstruction of multiple plausible point clouds from the input image(s).

Finally,  point-based representations can handle 3D shapes of arbitrary topologies. However, they require a post processing step, \eg Poisson surface reconstruction~\cite{kazhdan2013screened} or SSD~\cite{calakli2011ssd},  to retrieve the  3D surface mesh, which is the quantity of interest.   The pipeline, from the input until the final mesh is obtained, cannot be trained end-to-end. Thus, these methods only optimise  an auxiliary loss defined on an intermediate representation.

\section{Leveraging other cues}   
\label{sec:other_cues}

The previous sections discussed methods that directly reconstruct 3D objects from their 2D observations. This section shows how additional cues such as intermediate representations (Section~\ref{sec:intermediating}) and temporal correlations (Section~\ref{sec:temporal_correlations}) can be used to boost  3D reconstruction.

\subsection{Intermediating}
\label{sec:intermediating}

\begin{figure}[t]
\centering{
	\begin{tabular}{@{}c@{}}
		\includegraphics[width=.95\linewidth]{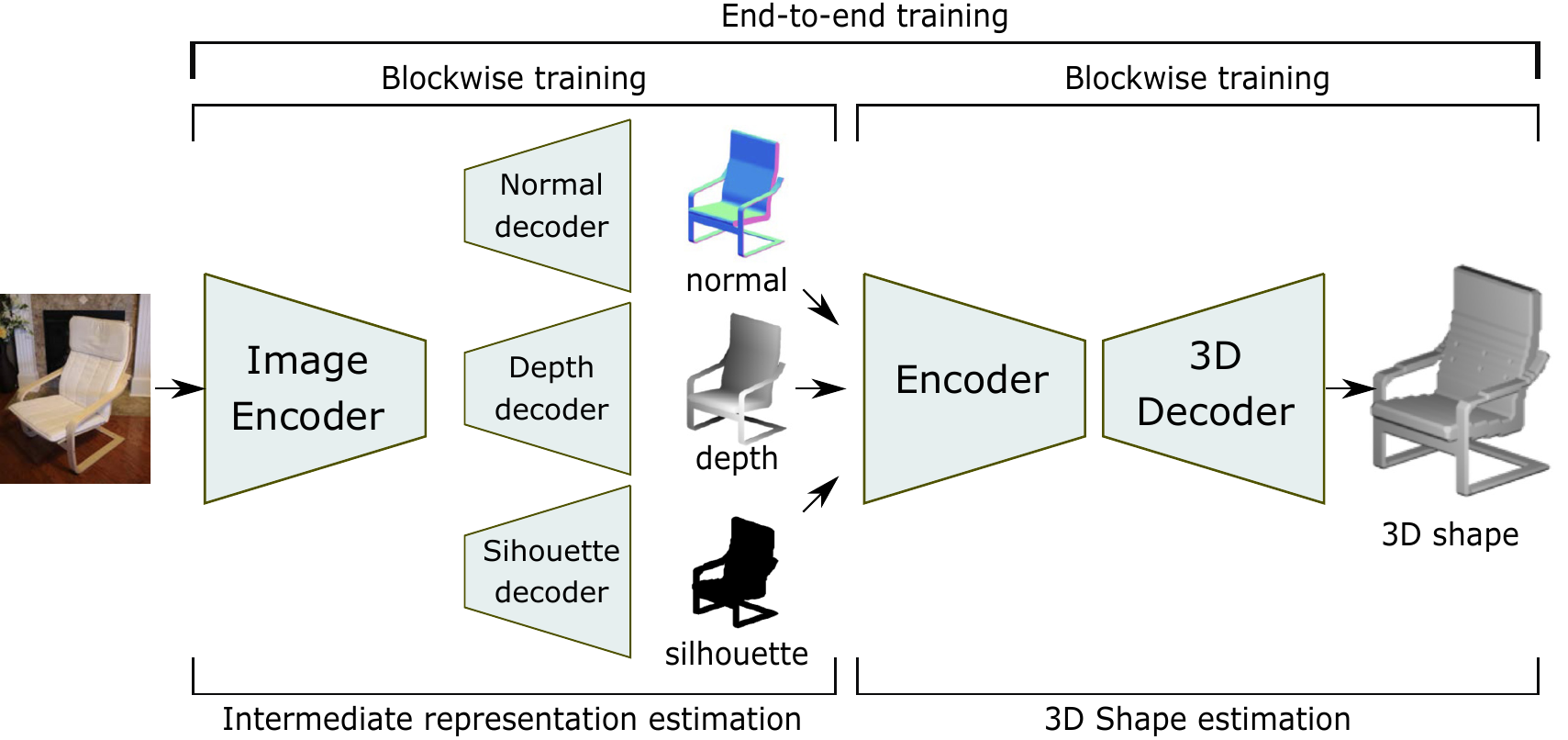} 
	\end{tabular}
	\caption{\label{fig:intermediating}Intermediating via 2.5D sketches (depth, normals, and silhouettes).} 
}
\end{figure}

Many of the deep learning-based 3D reconstruction algorithms directly predict the 3D geometry of an object  from  RGB images.  Some techniques, however,  decompose the problem into sequential steps, which estimate $2.5$D information such as depth maps, normal maps, and/or segmentation masks, see Fig.~\ref{fig:intermediating}.  The last step, which  can be implemented  using traditional techniques such as space carving or 3D back-projection followed by filtering and registration,  recovers the full 3D geometry and  the pose of the input. 

While early methods train separately the different modules, recent works proposed end-to-end solutions~\cite{wu2017marrnet,Pumarola_2018_CVPR,sun2018pix3d,Wu_2018_ECCV,Cherabier_2018_ECCV,zeng2018inferring,zhang2018learning}. For instance, Wu \etal~\cite{wu2017marrnet} and later Sun \etal~\cite{sun2018pix3d} used two blocks. The first block is an encoder followed by a three-branch decoder, which estimates the depth map, the normal map, and the segmentation mask (called $2.5$D sketches). These are then concatenated and  fed into another encoder-decoder, which regresses a full 3D volumetric grid~\cite{wu2017marrnet,sun2018pix3d,Wu_2018_ECCV}, and a set of fully-connected layers, which regress the camera pose~\cite{sun2018pix3d}. The entire network is trained end-to-end.

 Other techniques convert the intermediate depth map into \textbf{(1)} a 3D occupancy grid~\cite{yang2018dense}  or a truncated signed distance function volume~\cite{Cherabier_2018_ECCV}, which is then processed using a 3D encoder-decoder network for completion and refinement, or \textbf{(2)} a partial point cloud, which is further processed using a point-cloud completion module~\cite{zeng2018inferring}.  Zhang \etal~\cite{zhang2018learning} convert the inferred depth map into a spherical map and unpaint it, to fill in holes, using another encoder-decoder. The unpainted spherical depth map is then  back-projected to 3D and refined using a voxel refinement network, which estimates a voxel occupancy grid of size $128^3$.

%

Other techniques estimate multiple depth maps from pre-defined or arbitrary viewpoints. Tatarchenko \etal~\cite{tatarchenko2016multi} proposed a network, which takes as input an RGB image and a target viewpoint $\viewpoint$, and infers the depth map of the object as seen from the viewpoint $\viewpoint$. By varying the viewpoint, the network is able to estimate multiple depths, which can then be merged into a complete 3D model. The approach uses a standard encoder-decoder and an additional  network composed of three fully-connected layers to encode the viewpoint.  Soltani \etal~\cite{soltani2017synthesizing} and Lin \etal~\cite{lin2018learning}   followed the same approach but predict the depth maps, along with their binary masks, from pre-defined view points. In both methods, the merging is performed in a post-processing step.  Smith \etal~\cite{smith2018multi} first estimate a low resolution voxel grid. They then take the depth maps, of the low resolution voxel grid, computed from the six axis-aligned views and refine them using  a silhouette and depth refinement network. The refined depth maps are finally combined into a volumetric grid of size $256^3$ using space carving techniques.

Tatarchenko \etal~\cite{tatarchenko2016multi},  Lin \etal~\cite{lin2018learning}, and Sun \etal~\cite{sun2018pix3d}  also estimate the binary/silhouette masks, along with the depth maps. The binary masks  have been used to filter out points that are not back-projected to the surface in 3D space.    The side effect of these depth mask-based approaches is that  it is a huge computation waste as a large number of points are discarded, especially for objects with thin structures.  Li \etal~\cite{li2018efficient} overcome this problem by deforming a regular depth map using a learned deformation field. Instead of directly inferring depth maps that best fit the input, Li \etal~\cite{li2018efficient}  infer  a set of 2D pre-deformation depth maps and their corresponding deformation fields at pre-defined canonical viewpoints. These are each passed to a Grid Deformation Unit (GDU) that transforms the regular grid of the depth map to a \emph{deformed depth map}. Finally, the deformed depth maps are transformed into a common coordinate frame for fusion into a  dense point cloud. 

The main advantage of  multi-staged approaches is that  depth, normal, and silhouette maps are much easier to recover from 2D images. Likewise, 3D models are much easier to recover from these three modalities than from 2D images alone.

\subsection{Exploiting spatio-temporal correlations}
\label{sec:temporal_correlations}



There are many situations where multiple spatially distributed images of the same object(s) are acquired  over an extended period of time.   Single image-based reconstruction techniques  can be used to reconstruct the 3D shapes by processing individual frames independently from each other, and then merging the reconstruction using registration techniques.  Ideally, we would like to leverage  on the spatio-temporal correlations that exist between the frames to resolve ambiguities especially in the presence of occlusions and highly cluttered scenes.  In particular, the network at time $t$ should remember what has been reconstructed up to time $t-1$, and use it, in addition to the new input, to reconstruct the scene or objects at time $t$. This  problem of processing sequential data  has been addressed by using Recurrent Neural Networks (RNN) and Long-Short Term Memory (LSTM) networks, which  enable  networks to remember their inputs over a  period of time.

Choy \etal~\cite{choy20163d} proposed an architecture called 3D Recurrent Reconstruction Network (3D-R2N2), which allows the network to  adaptively and consistently learn a suitable 3D representation of an object as (potentially conflicting) information from different viewpoints becomes available. The network can perform incremental refinement every time a new view  becomes available.  It  is composed of two parts; a standard convolution encoder-decoder  and a set of 3D Convolutional Long-Short Term Memory (3D-LSTM) units  placed at the start of the convolutional decoder.  These take the output of the encoder, and then either selectively update their cell states or retain the states by closing the input gate.  The decoder then decodes the hidden states of the LSTM units and generates a  probabilistic  reconstruction in the form of a voxel occupancy map. 

The 3D-LSTM allows the network to retain what it has seen and update its memory when it sees a new image. It is able to effectively handle object self-occlusions when multiple views are fed to the network. At each time step, it selectively updates the memory cells that correspond to parts that became visible while retaining the states of the other parts.

LSTM and RNNs are time consuming since the input images are processed sequentially without parallelization.  Also, when given the same set of images with different orders, RNNs are unable to estimate the 3D shape of an object consistently  due to permutation variance. To overcome these limitations, Xie \etal~\cite{xie2019pix2vox} introduced Pix2Vox, which is composed of multiple encoder-decoder blocks, running in parallel, each one predicts a coarse volumetric grid from its input frame.  This eliminates the effect of the order of input images and accelerates the computation. Then, a context-aware fusion module selects high-quality reconstructions from the coarse 3D volumes and generates a fused
3D volume, which fully exploits information of all input images without long-term memory loss.

\section{Training}
\label{sec:training_deeplearning}
In addition to their architectures, the performance of deep learning networks depends on the way they are trained. This section discusses  the various supervisory modes (Section~\ref{sec:degree_of_supervision}) and training procedures that have been used in the literature (Section~\ref{sec:training_procedures}).

\subsection{Degree of supervision} 
\label{sec:degree_of_supervision}

Early methods rely  on 3D supervision (Section~\ref{sec:3dsupervision_loss}). However, obtaining ground-truth 3D data, either manually or using traditional 3D reconstruction techniques, is extremely difficult and expensive. As such, recent techniques try to minimize the amount of 3D supervision by exploiting other supervisory signals such consistency across views (Section~\ref{sec:2D_supervision}).  

\subsubsection{Training with 3D supervision} 
\label{sec:3dsupervision_loss}
Supervised methods  require training using images  paired with their corresponding ground-truth 3D shapes. The training process then minimizes a loss function that measures the discrepancy between the reconstructed 3D shape and the corresponding  ground-truth 3D model. The discrepancy is measured using loss functions, which  are required to be differentiable so that gradients can be computed. Examples of such functions include:

\vspace{6pt}
\noi\textbf{(1) Volumetric loss. } It is defined as the  distance between the reconstructed and the ground-truth volumes; 
\begin{equation}
	\objectivefunc_{vol} (\images) = d\left(\recofunc(\images), \shape  \right). 
\label{eq:ltwo_loss}
\end{equation}

\noi Here, $d(\cdot, \cdot)$ can be the  $\ltwo$ distance between the two volumes or  the  negative Intersection over Union (IoU) $\objectivefunc_{IoU}$ (see Equation~\eqref{eq:IoU}). Both metrics are suitable for binary occupancy grids and TSDF representations. For probabilistic occupancy grids, the cross-entropy loss is the most commonly used~\cite{girdhar2016learning}: 
\begin{equation}
	\loss_{CE} = -\frac{1}{\npoints} \sum_{i=1}^{\npoints} \left\{ p_i \log\hat{p}_i  + (1 - p_i)\log(1 - \hat{p}_i      \right\}.
	\label{eq:cross_entropy_loss}
\end{equation}

\noi Here,  $p_i$ is the ground-truth probability of voxel $i$ being occupied, $\hat{p}_i $  is the estimated probability, and $\npoints$ is the number of voxels.

\vspace{6pt}
\noi \textbf{(2) Point set  loss. } When using point-based representations, the reconstruction loss can be measured using the Earth Mover's Distance (EMD)~\cite{kurenkov2017deformnet,fan2017point} or the Chamfer Distance (CD)~\cite{kurenkov2017deformnet,fan2017point}.  The EMD is defined as the minimum of the sum of distances between a point in one set and a point in another set over all possible permutations of the correspondences. More formally, given two sets of points $\pointset_{gt}$ and $\pointset_{rec}$, the EMD is defined as:
\begin{equation}
	\loss_{EMD} = \min_{\pointset_{gt} \to \pointset_{rec}}  \sum_{\point \in \pointset_{gt}}  \| \point - \phi(\point) \|.
	\label{eq:emd}
\end{equation}

\noi Here, $\phi(\point) \in \pointset_{rec}$ is the closest point on $\pointset_{rec}$ to $\point \in \pointset_{gt}$. The CD loss,   on the other hand,  is defined as:
\begin{equation}
	\loss_{CD} = \frac{1}{\npoints_{gt}} \min_{\point \in \pointset_{gt}  }     \|\point -q\|^2 +
			    \frac{1}{\npoints_{rec}}   \min_{q \in \pointset_{rec}  }  \|\point -q\|^2.
	\label{eq:CD}
\end{equation}

\noi  $\npoints_{gt}$ and  $\npoints_{rec}$ are, respectively, the size of  $\pointset_{gt} $ and $\pointset_{rec}$.  The CD  is computationally easier than EMD since it uses sub-optimal matching to determine the pairwise relations.

\vspace{6pt}
\noi \textbf{(3) Learning to generate multiple plausible reconstructions. }\label{sec:multiple_plausible_shapes}
3D reconstruction from a single image is an ill-posed problem, thus for a given input there might be multiple plausible reconstructions.  Fan \etal~\cite{fan2017point} proposed the Min-of-N (MoN)  loss to train neural networks to  generate distributional output.  The idea is to use a random vector $r$ drawn from a certain distribution to perturb the input. The network learns to generate a plausible 3D shape from each perturbation of the input. It is trained using a loss defined as follows;
\begin{equation}
	\loss_{MoN} = \sum_{i} \min_{r \sim \mathbb{N}(0, \text{I} )}  \left\{ d \left(  \recofunc (\images, r),   \pointset_{gt}   \right)   \right\}.
\end{equation}

\noi Here, $\recofunc (\images, r)$ is the reconstructed 3D point cloud after perturbing the input with the random vector $r$ sampled from the multivariate normal distribution $\mathbb{N}(0, \text{I} )$,  $\pointset_{gt}$ is the ground-truth point cloud, and $d(\cdot, \cdot)$ is a reconstruction loss, which can be any of the loss functions defined above. At runtime, various plausible reconstructions can be generated from a given input by sampling different random vectors $r$ from  $\mathbb{N}(0, \text{I} )$.

\subsubsection{Training with 2D supervision}
\label{sec:2D_supervision}

Obtaining 3D ground-truth data for supervision is an expensive and  tedious process even for a small scale training.  However, obtaining multiview $2$D or $2.5$D images  for training is relatively easy. Methods in the category use the fact that if the estimated 3D shape  is as close as possible to the ground truth then the discrepancy between views of the 3D model and the projection of the reconstructed 3D model onto any of these views is also minimized.  Implementing this idea requires defining a projection operator, which renders the reconstructed 3D model from a given viewpoint (Section~\ref{sec:projection_operator}), and a loss function that measures the reprojection error (Section~\ref{sec:reprojection_loss}).

\vspace{6pt}
\paragraph{Projection operators}
\label{sec:projection_operator}

Techniques from projective geometry can be used to render views of a 3D object. However, to enable end-to-end training  without gradient approximation~\cite{kato2018neural}, the projection operator should be differentiable.  Gadelha~\etal~\cite{gadelha20173d} introduced a differentiable projection operator $P$   defined as $P((i, j), \vgrid) = 1 - e^{-\sum_k\vgrid(i,j, k)}$, where $\vgrid$ is the 3D voxel grid.  This operator sums up the voxel occupancy values along each line of sight. However, it assumes  an orthographic projection.  Loper and Black~\cite{loper2014opendr}  introduced OpenDR, an approximate differentiable renderer, which is suitable for orthographic and perspective projections. 

Petersen \etal~\cite{petersen2019pix2vex} introduced a novel $C^\infty$  smooth differentiable renderer for image-to-geometry reconstruction. The idea is that instead of taking a discrete decision of which triangle is the visible from a pixel, the approach softly blends their visibility. Taking the weighted SoftMin of the $z$-positions in the camera space constitutes a smooth z-buffer, which leads to a $C^\infty$ smooth renderer, where the $z$-positions of triangles are differentiable with respect to occlusions. In previous  renderers, only the $xy$-coordinates were locally differentiable with respect to occlusions.

Finally, instead of using fixed renderers, Rezende \etal~\cite{rezende2016unsupervised}   proposed a learned projection operator, or a learnable camera, which is built by first applying an affine transformation to the reconstructed volume, followed by a combination of 3D and 2D convolutional layers, which map the 3D volume onto a 2D image.

\vspace{6pt}
\paragraph{Re-projection loss functions}
\label{sec:reprojection_loss}
There are several loss functions that have been proposed for 3D reconstruction using 2D supervision. We classify them into two main categories; (1) silhouette-based and (2) normal and depth-based loss functions.

\vspace{6pt}
\noi\textbf{(1) Silhouette-based loss functions. } The idea is that a 2D silhouette projected from the reconstructed volume, under certain camera intrinsic and extrinsic parameters, should match the ground truth 2D silhouette of the input image. The discrepancy, which is inspired by space carving,  is then:
\begin{equation}
	\objectivefunc_{proj} (\images) = \frac{1}{n}\sum_{j=1}^n d\left(P \left( \recofunc(\images); \cameraparams^{(j)} \right) ,  S^{(j)}\right),
	\label{eq:2Dloss_reprojection}
\end{equation}

\noi where $S^{(j)}$ is the $j-$th ground truth 2D silhouette of the original 3D object $\shape$, $n$ is the number of silhouettes or views used for each 3D model,   $P(\cdot)$ is a 3D to 2D projection function, and $\cameraparams^{(j)}$ are the camera parameters of the $j$-th silhouette. The distance metric $d(\cdot, \cdot)$ can be the standard $\ltwo$ metric~\cite{insafutdinov2018unsupervised},  the negative Intersection over Union (IoU)   between the true and reconstructed silhouettes~\cite{kato2018neural}, or the binary cross-entropy loss~\cite{zhu2017rethinking,yan2016perspective}. 

Kundu \etal~\cite{kundu20183d} introduced the render-and-compare loss, which is  defined in terms of the IoU between the ground-truth silhouette $G_s$ and the rendered silhouette $R_s$, and the $\ltwo$ distance between the ground-truth depth  $G_d$ and the rendered depth $R_d$, \ie
\begin{equation}
	\loss_{r} = 1 - IoU(R_s, G_s; I_s)   + d_{\ltwo}(R_d, G_d; I_d).
\end{equation}

\noi Here,  $I_s$ and $I_d$ are binary ignore masks that have value of one at pixels which do not contribute to the loss. Since this loss is not differentiable,  Kundu \etal~\cite{kundu20183d}  used finite difference to approximate its gradients. 

Silhouette-based loss functions cannot distinguish between some views, \eg front and back. To alleviate this issue, Insafutdinov and Dosovitskiy~\cite{insafutdinov2018unsupervised} use multiple pose regressors during training, each one using silhouette loss. The overall network is trained with  the min of the individual losses.  The predictor with minimum loss is used at test time. 

Gwak \etal~\cite{gwak2017weakly} minimize the reprojection error subject to the reconstructed shape being a valid member of a certain class, \eg  chairs.  To constrain the reconstruction to remain in the manifold of the shape class, the approach defines a barrier function $\phi$, which is set to be $1$ if the shape is in the manifold and $0$ otherwise. The loss function is then:
			\begin{equation}
				\loss = \loss_{reprojection } - \frac{1}{t}\log \phi(\estimatedshape).
			\end{equation}
			
\noi The barrier function is learned as the discriminative function of a GAN, see Section~\ref{sec:adversarial_training}.

Finally, Tulsiani \etal~\cite{tulsiani2017multi} define the re-projection loss using a differentiable ray consistency loss for volumetric reconstruction.  First, it assumes that the estimated shape $\estimatedshape$ is defined in terms of the probability occupancy grid. Let   $(O, C)$ be an observation-camera pair. Let also $\mathcal{R}$ be a set of rays where each ray $r\in \mathcal{R}$ has the camera center as origin and is casted through the image plane of the camera $C$. The ray consistency loss is then defined as:
\begin{equation}
	\loss_{ray\_cons}(\estimatedshape; (O, C)) = \sum_{r\in \mathcal{R}} \loss_{r} (\estimatedshape), 
\end{equation}

\noi where $\loss_{r} (\estimatedshape)$ captures if the inferred 3D model $\estimatedshape$ correctly explains the observations associated with the specific ray $r$. If the observation $O$ is a ground-truth foreground mask taking values $0$ at foreground pixels and $1$ elsewhere, then  $\loss_{r}$ is   the probability that the ray $r$ hits a surface voxel weighted by the mask value at the pixel associated with the ray $r$. This loss is differentiable with respect to the network predictions. Note that when using foreground masks as observations, this loss, which requires known camera parameters,  is similar to the approaches designed to specifically use mask supervision  where a learned~\cite{girdhar2016learning} or a fixed~\cite{yan2016perspective} reprojection function is used. Also, the binary cross-entropy loss used in~\cite{yan2016perspective,zhu2017rethinking}  can be thought of as an approximation of the one derived using ray consistency.


\vspace{6 pt}
\noi\textbf{(2) Surface normal and depth-based loss. }  Additional cues such as surface normals and depth values can  be  used to guide the training process.  Let $n_{x,y} = (n_a, n_b, n_c)$ be a normal vector to a surface at a point $(x, y, z)$.  The vectors $n_x = (0, -n_c, n_b)$ and $(-n_c, 0, n_a)$ are orthogonal to $n_{x, y}$. By normalizing them, we obtain two vectors $n'_x = (0, -1, n_b/n_c)$ and $n'_y = (-1, 0, n_a/n_c)$. The normal loss tries to guarantee that the voxels  at $(x, y, z) \pm n'_x$ and $(x, y, z) \pm n'_{y}$ should be $1$ to match the estimated surface normals. This constraint only applies when the target voxels are inside the estimated silhouette. The projected surface normal loss is then:
	{
	\small{
	\begin{eqnarray}
		\loss_{normal} &=& \left(  1 - v_{x, y-1, z + \frac{n_b}{n_c}} \right)^2  +   \left(  1 - v_{x, y+1, z - \frac{n_b}{n_c}} \right)^2 + \nonumber \\ 
				       &  & \left(  1 - v_{x-1, y, z + \frac{n_a}{n_c}} \right)^2 +  \left(  1 - v_{x+1, y, z - \frac{n_a}{n_c}} \right)^2.
	\end{eqnarray}
	}
	}

\noi This loss has been used by Wu \etal~\cite{wu2017marrnet}, which includes, in addition to the normal loss,  the projected depth loss.  The idea is  that the voxel with depth $v_{x,y,d_{x,y}}$  should be $1$, and all voxels in front of it should be $0$. The depth loss is then defined as:

			\begin{equation}
				\loss_{depth}(x, y, z) = \left\{ \begin{tabular}{ll}
										$v^2_{x, y, z}$  & \text{ if } $z < d_{x, y}$,\\
										$(1 - v_{x, y, z})^2$ 	& \text{ if } $z = d_{x, y}$, \\
										$0$ 				& \text{ otherwise.}
									   \end{tabular}
								  \right.
			\end{equation}

\noi This ensures the estimated 3D shape matches the estimated depth values.

\vspace{6 pt}
\noi\textbf{(3) Combining multiple losses.} One can also combine 2D and 3D losses. This is particularly useful when some ground-truth 3D data is available. One can for example train first the network using 3D supervision, and then fine-tune it using 2D supervision. Yan \etal~\cite{yan2016perspective}, on the other hand, take the weighted sum of a 2D and a 3D loss.

In addition to the reconstruction loss, one can impose additional constraints to the solution. For instance, Kato \etal~\cite{kato2018neural} used a weighted sum of silhouette loss, defined as the negative intersection over union (IoU) between the true and reconstructed silhouettes,  and a smoothness loss. For surfaces, the smoothness loss ensures that the angles between adjacent faces is close to $180^o$, encouraging flatness.

\vspace{6pt}
\paragraph{Camera parameters and viewpoint estimation} 
Reprojection-based loss functions  use the camera parameters to render the estimated 3D shape onto image planes. Some methods assume the availability of one or multiple observation-camera pairs~\cite{tulsiani2017multi,zisserman2017silnet,yan2016perspective}. Here, the observation can be an RGB image, a silhouette/foreground mask or a depth map of the target 3D shape. Other methods optimize at the same time for the camera parameters and the 3D reconstruction that best describe the input~\cite{gadelha20173d,insafutdinov2018unsupervised}. 
 
Gadelha \etal~\cite{gadelha20173d} encode an input image into a latent representation and a pose code using fully-connected layers. The pose code is then used as input to the 2D projection module, which renders the estimated 3D volume onto the view of the input. 
Insafutdinov and Dosovitskiy~\cite{insafutdinov2018unsupervised}, on the other hand, take two views of the same object, and predict the corresponding shape (represented as a point cloud) from the first view, and the camera pose (represented as a quaternion) from the second one. The approach then uses  a differentiable projection module to generate the view of the predicted shape from the predicted camera pose.  The shape and pose predictor is implemented as a convolutional network with two branches. The network starts with a convolutional encoder with a total of $7$ layers followed by $2$ shared fully connected layers, after which the network splits into two branches for shape and pose prediction.  The pose branch is implemented as a multi-layer perceptron. 

There has been a few papers that only estimate the camera pose~\cite{kendall2015posenet,su2015render,tulsiani2017learning}. Unlike techniques that do simultaneously reconstruction, these approaches are trained with only pose annotations. For instance, Kendall \etal~\cite{kendall2015posenet} introduced PoseNet, a convolutional neural network which estimates the camera pose from a single image. The network, which  represents the camera pose using its location vector and orientation quaternion, is trained to minimize the $\ltwo$ loss between the ground-truth and the estimate pose. Su \etal~\cite{su2015render} found that CNNs trained for viewpoint estimation of one class do not perform well on another class, possibly due to the huge geometric variation between the classes. As such, they proposed a network architecture where the lower layers (both convolutional layers and fully connected layers) are shared by all classes, while class-dependent fully-connected layers are stacked over them.


%

\subsection{Training with video supervision}

Another approach to significantly lower the level of supervision required to learn the 3D geometry of objects is by replacing 3D supervision with motion. To this end, Novotni \etal~\cite{novotny2018capturing} used Structure-from Motion (SfM) to generate a supervisory signal from videos. That is, at training, the approach takes a video sequences, generates a partial point cloud and the relative camera parameters using SfM~\cite{schonberger2016structure}. Each RGB frame is then processed with a network that estimates a depth map, an uncertainty map, and the camera parameters. The different depth estimates are fused, using the estimated camera parameters, into a partial point cloud, which is further processed for completion using the point cloud completion network PointNet~\cite{qi2017pointnet}. The network is trained using the estimates of the SfM as supervisory signals. That is, the loss functions measure the discrepancy between the  depth maps estimated by the network and the depth maps estimated by SfM, and between the camera parameters estimated by the network and those estimated by SfM. At test time, the network is able to recover a full 3D geometry from a single RGB image.

\subsection{Training procedure}
\label{sec:training_procedures}

In addition to the datasets, loss functions, and degree of supervision, there are a few practical aspects that one needs to consider when training deep learning architectures for 3D reconstruction. 

\begin{figure}[t]
\centering{
	\begin{tabular}{@{}c@{}c@{}}
		\includegraphics[width=0.24\textwidth]{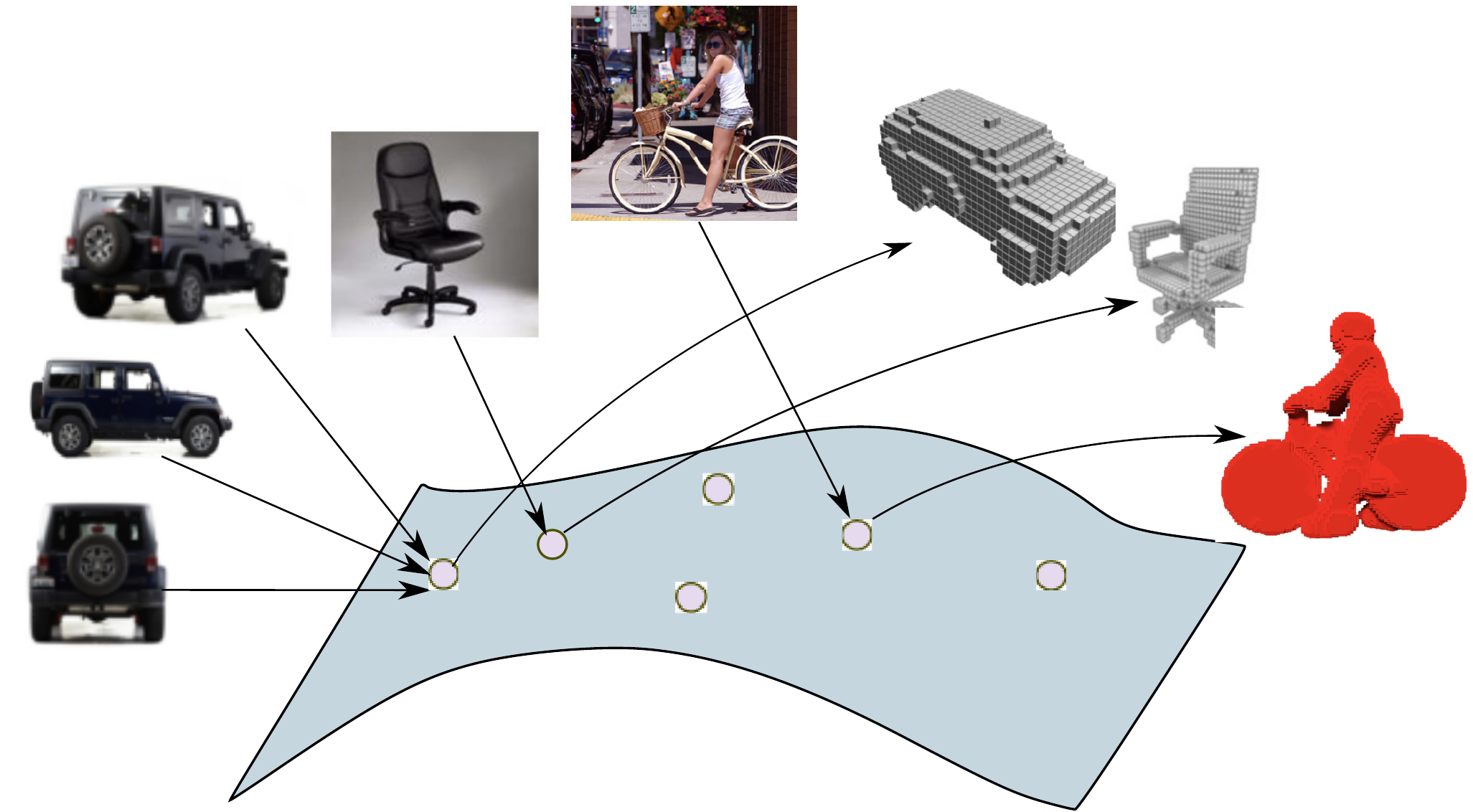}&\includegraphics[width=0.24\textwidth]{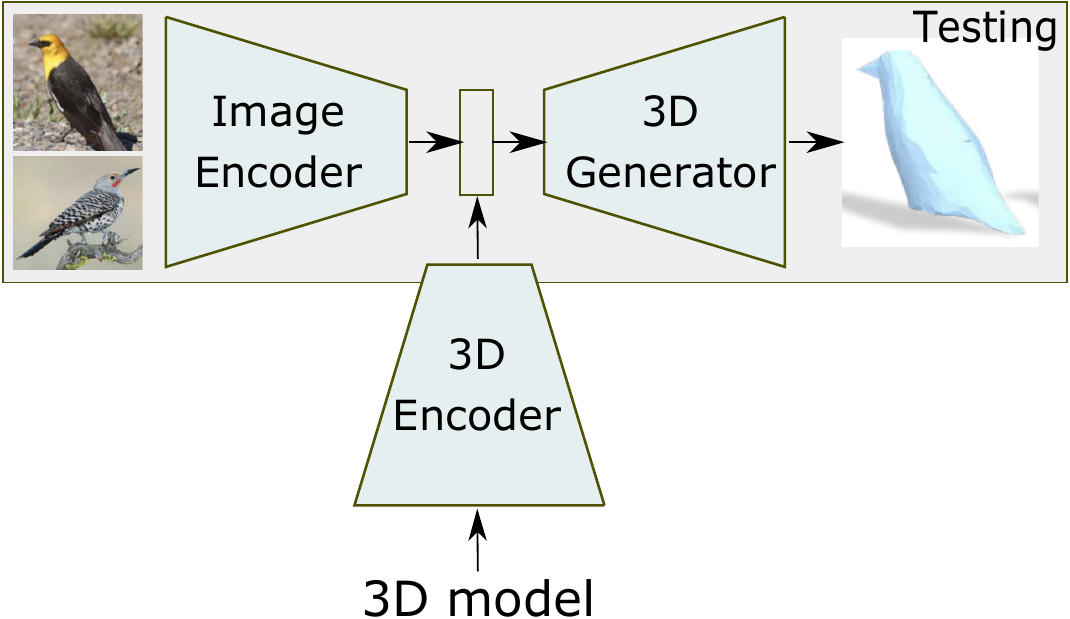} \\
		\small{(a) Joint 2D-3D embedding.} & \small{(b)  TL-network.}\\
		\\
		\multicolumn{2}{c}{\includegraphics[width=0.4\textwidth]{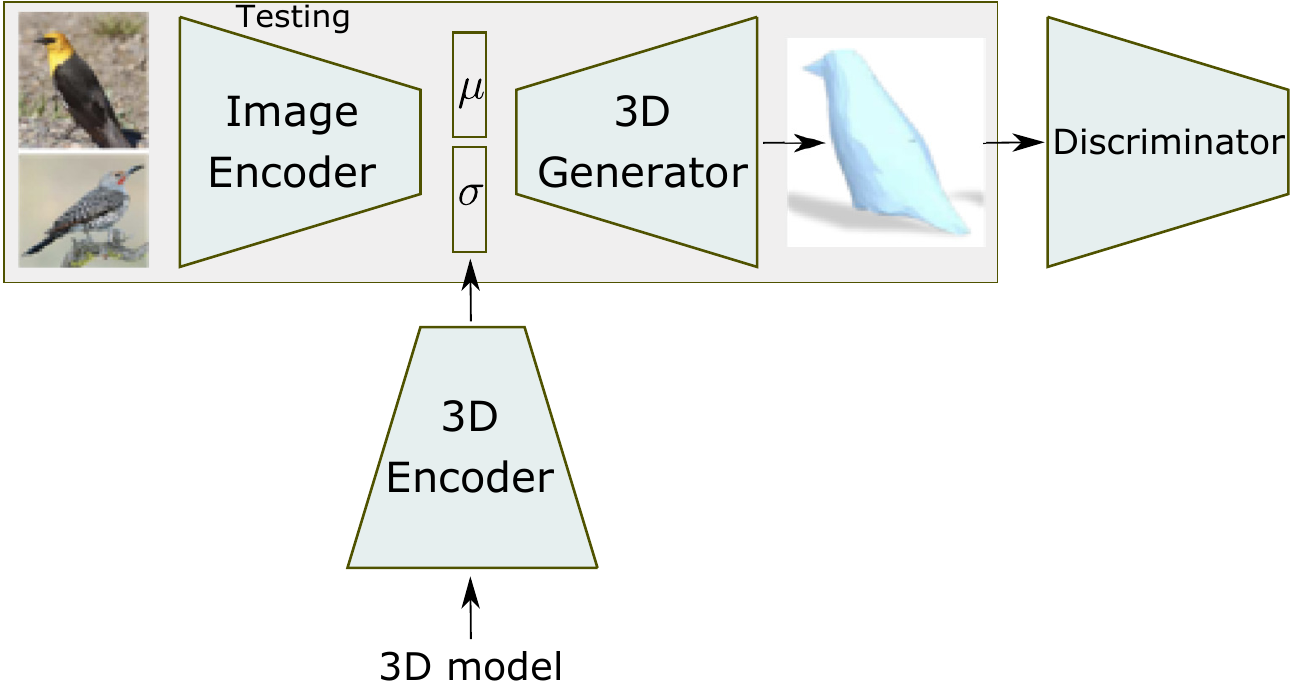}} \\
		\multicolumn{2}{c}{\small{(c) 3D-VAE-GAN architecture.}}
		
	\end{tabular}
	\caption{\label{fig:TL_GAN_architectures} At test time, the 3D encoder and the discriminator are removed and only the highlighted modules are kept. }
}
\end{figure}
\subsubsection{Joint 2D-3D embedding}
\label{sec:joint_image_3D_embedding}

Most of the state-of-the-art works map the input (\eg RGB images) into a latent representation, and then decode the latent representation into a 3D model. A good latent representation should be  \textbf{(1)} generative in 3D, \ie we should be able to reconstruct objects in 3D from it, and \textbf{(2)}  it must be predictable from 2D, \ie we should be able to easily infer this representation from images~\cite{girdhar2016learning}.  Achieving these two goals has been addressed by using TL-embedding networks  during the training phase, see Fig.~\ref{fig:TL_GAN_architectures}-(a) and (b). It is composed of two jointly trained encoding branches: the 2D encoder and the 3D encoder. They map, respectively, a 2D image and its corresponding 3D annotation into the same point in the latent space~\cite{girdhar2016learning,zhu2017rethinking}. 


Gidhar \etal~\cite{girdhar2016learning}, which use the TL-embedding network to reconstruct volumetric shapes from RGB images, train the network using batches of (image, voxel) pairs. The images are generated by rendering the 3D model and the network is then trained in a three stage procedure. 
\begin{itemize}
	\item In the first stage, the 3D encoder part of the network and its decoder are initialized at random. They are then trained,  end-to-end with the sigmoid cross-entropy loss,  independently of the 2D encoder. 
	
	\item In the second stage, the 2D encoder is trained to regress  the latent representation. The encoder generates the embedding for the voxel, and the image network is trained to regress the embedding.
	
	\item The final stage jointly  fine-tunes the entire network. 
\end{itemize}

\noi This approach has been extended by Li \etal~\cite{li2018optimizable} and Mandikal \etal~\cite{mandikal20183d} for point cloud-based 3D reconstruction by replacing the volume encoder by a point cloud auto-encoder. 

	
%

\subsubsection{Adversarial training} 
\label{sec:adversarial_training}


In general, a good reconstruction  model should be able to go beyond what has been seen during training. Networks trained with standard procedures may not generalize well to unseen data. Also, Yang \etal~\cite{yang2018dense}  noted that  the results of standard techniques  tend to be grainy and lack fine details. To overcome these issues, several recent papers train their networks with adversarial loss by using Generative Adversarial Networks (GAN). GANs generate a signal from a given random vector~\cite{goodfellow2014generative}. Conditional GANs, on the other hand, conditions the generated signal on the input image(s), see Fig.~\ref{fig:TL_GAN_architectures}-(c).   It  consists of a generator $\decodingfunc$, which mirrors the encoder $\encodingfunc$,  and a discriminator $ \discriminatingfunc$, which mirrors the generator.  

In the case of 3D reconstruction, the encoder can be a ConvNet/ResNet ~\cite{yang20173d,yang2018dense} or a variational auto-encoder (VAE)~\cite{wu2016learning}. The generator decodes the latent vector $\featurevector$ into a 3D  shape $\shape = \decodingfunc(\featurevector)$.  The discriminator, which is only used during training,  evaluates the authenticity of the decoded data. It outputs a confidence   $\confidencefunc(\shape)$ between $0$ and $1$ of whether  the 3D object $\shape$  is real or synthetic, \ie coming from the generator. The goal is to jointly train the generator and the discriminator to make the reconstructed shape as close as possible to the ground truth.   

Central  to GAN is the adversarial loss function used to jointly train the discriminator and the generator. Following Goodfellow \etal~\cite{goodfellow2014generative}, Wu \etal~\cite{wu2016learning} use the binary cross entropy as the classification loss. The overall adversarial loss function is defined as:
\begin{equation}
	\lossgan = \log\left( \discriminatingfunc(\shape ) \right)  + \log\left(  1 -   \discriminatingfunc\left( \decodingfunc(\featurevector) \right) \right). 
\end{equation}

\noi Here  $\featurevector = \encodingfunc(\images)$ where   $\images$ is the 2D images(s) of the training shape $\shape$.  Yang \etal~\cite{yang20173d,yang2018dense} observed that the original GAN loss function presents an overall loss for both real and fake input. They then proposed to use the WGAN-GP loss~\cite{arjovsky2017wasserstein,gulrajani2017improved},  which separately represents the loss  for generating fake reconstruction pairs and the loss for discriminating fake and real reconstruction pairs, see~\cite{arjovsky2017wasserstein,gulrajani2017improved} for the details.

To  jointly train the three components of the network, \ie the encoder, the generator, and the discriminator, the overall loss is defined as the sum of the reconstruction loss, see Section~\ref{sec:degree_of_supervision}, and the GAN loss. When the network uses a variational auto-encoder, \eg the 3D VAE-GAN~\cite{wu2016learning}, then an additional term is added to the overall loss in order to push the variational distribution towards the prior distribution. For example, Wu \etal~\cite{wu2016learning} used a KL-divergence metric, and a multivariate Gaussian distribution with zero-mean and unit variance as a prior distribution.

The potential of GANs is huge, because they can learn to mimic any distribution of data.  They are also very suitable  for single-view 3D shape reconstruction. They have been used for volumetric~\cite{wu2016learning,smith2017improved,yang20173d,yang2018dense,Yang_2018_ECCVlearning,knyaz2018image} and point cloud~\cite{Jiang_2018_ECCV,li2018point} reconstruction. They have been used with 3D supervision~\cite{wu2016learning,smith2017improved,yang20173d,yang2018dense,knyaz2018image} and with 2D supervision as in~\cite{gadelha20173d,gwak2017weakly,Yang_2018_ECCVlearning}, see Section~\ref{sec:2D_supervision}.  The latter methods train a single discriminator with 2D silhouette images. However, among plausible shapes, there are still multiple shapes that fit the 2D image equally well.  To address this ambiguity, Wu \etal~\cite{Wu_2018_ECCV} used the discriminator of the GAN to penalize  the 3D estimator if the predicted 3D shape is unnatural.   Li \etal~\cite{Li_2019_CVPR}, on the other hand,  use multiple discriminators, one for each view, resulting in a better generation quality.

GANs are hard to train, especially for the complex joint data distribution over 3D objects of many categories and orientations. They also become unstable for high-resolution shapes. In fact, one must carefully balance the learning of the generator and the discriminator, otherwise the gradients can vanish,  which will prevent improvement~\cite{smith2017improved}. To address this issue,  Smith and Meger~\cite{smith2017improved}  and later  Wu \etal~\cite{Wu_2018_ECCV}  used as a training objective  the Wasserstein distance normalized with the gradient penalization.


\subsubsection{Joint training with other tasks}

Jointly training for  reconstruction and segmentation leads to improved performance in both tasks, when compared to training for each task individually. Mandikal \etal~\cite{mandikal20183dECCV}  proposed 	an approach, which generates a part-segmented 3D point cloud from one RGB image. The idea is to  enable  propagating information between the two tasks so as to generate more faithful part reconstructions while also improving segmentation accuracy. This is done using a weighted sum of a reconstruction loss, defined using the Chamfer distance, and a segmentation loss, defined using the symmetric softmax cross-entropy loss. 




\section{Applications and special cases}
\label{sec:applications}

Image-based 3D reconstruction is an important problem and a building block to many applications ranging from robotics and autonomous navigation to graphics and entertainment. While some of these applications deal with generic objects, many of them deal with objects that belong to specific classes such as human bodies or body parts (\eg faces and hands), animals in the wild, and cars. The techniques described above can be applied to these specific classes of shapes. However, the quality of the reconstruction can be significantly improved by designing customised methods that leverage the prior knowledge of the shape class. In this section, we will briefly summarize recent developments in the image-based 3D reconstruction of human body shapes  (Section~\ref{sec:3Dhumanbody}), and body parts such as faces (Section~\ref{sec:3DFaces}). We will also discuss  in Section~\ref{sec:3D_scene_parsing} methods that deal with the parsing entire 3D scenes.

\subsection{3D human body reconstruction}
\label{sec:3Dhumanbody}

3D static and dynamic digital humans are essential for a variety of applications ranging from gaming, visual effects to free-viewpoint videos. However, high-end  3D capture solutions use a large number of cameras and active sensors, and are restricted to professionals as they  operate under controlled lighting conditions and studio settings. With the avenue of deep learning techniques, several papers have explored more lightweight solutions that are able to recover the 3D human shape and pose from a few RGB images. We can distinguish two classes of methods; \textbf{(1)} volumetric methods (Section~\ref{sec:volumetric_3D_reconstruction}), and \textbf{(2)} template or parameteric-based methods (Section~\ref{sec:deformationbased}). Some methods in both categories  reconstruct naked 3D human body shapes~\cite{dibra2016hs,dibra2017human}, while others recover also cloths and garments~\cite{alldieck19cvpr,bhatnagar2019mgn}.

\subsubsection{Parametric methods}

Parametric methods regularize the problem using statistical  models such as morphable models~\cite{allen2003space}, SCAPE~\cite{anguelov2005scape}, and SMPL~\cite{loper2015smpl}. The problem of 3D human body shape reconstruction then boils down to estimating the parameters of the model. 

Dibra \etal~\cite{dibra2016hs} used  an encoder followed by three fully connected layers to regress the SCAPE parameters from one or multiple silhouette images. Later, Dibra \etal~\cite{dibra2017human} first learn a common embedding of 2D silhouettes and 3D human body shapes (see Section~\ref{sec:joint_image_3D_embedding}). The latter are represented using their  Heat Kernel Signatures~\cite{sun2009concise}. Both methods can only predict naked body shapes in nearly neutral poses. 

SMPL has the advantage of encoding in a disentangled manner both the  shape, the pose, and the pose specific details,  and thus it has been extensively used in deep learning-based human body shape reconstruction~\cite{bogo2016keep,omran2018NBF,alldieck19cvpr,alldieck2019tex2shape}.  Bogo \etal \cite{bogo2016keep} proposed SMPLify, the first  3D human pose and shape reconstruction from one image. They first used a CNN-based architecture, DeepCut~\cite{pishchulin2016deepcut}, to estimate the 2D joint locations. They then  fit an  SMPL model to the predicted 2D joints giving the estimation of 3D human body pose and shape. The training procedure minimizes an objective function of five terms: a joint-based data term, three pose priors, and a shape prior. Experimental results show that this method is effective in  3D human body reconstruction from arbitrary poses.


Kanazawa \etal~\cite{kanazawa2018end}, on the other hand,  argue that such a stepwise approach is not optimal and propose an end-to-end solution to learn a direct mapping from image pixels  to model parameters.  This approach addresses two important challenges: \textbf{(1)} the lack of large scale ground truth 3D annotations for in-the-wild images, and \textbf{(2)} the inherent ambiguities in single 2D-view-to-3D mapping of human body shapes. An example is depth ambiguity where multiple 3D body configurations can explain the same 2D projections~\cite{bogo2016keep}.  To address the first challenge, Kanazawa \etal \ observe that  there are large-scale 2D keypoint annotations of in-the-wild images and a separate large-scale dataset of 3D meshes of people with various poses and shapes. They then  take advantage of these unpaired 2D keypoint annotations and 3D scans in a conditional generative adversarial manner. They propose a network that infers the SMPL~\cite{loper2015smpl} parameters of a 3D mesh and the camera parameters such that the 3D keypoints match the annotated 2D keypoints after projection. To deal with ambiguities, these parameters are sent to a discriminator whose task is to determine if the 3D parameters correspond to bodies of real humans or not. Hence, the network is encouraged to output parameters on the human manifold. The discriminator acts as a weak supervisor. 

These approaches can handle complex poses from images with complex backgrounds, but are limited to a single person per image and does not handle clothes.  Also, these approaches do not capture details such as hair and clothing with garment wrinkles, as well has details on the body parts. To capture these details, Alldieck \etal~\cite{alldieck2019tex2shape} train an encoder-decoder to predict  normals and displacements, which can then be applied to  the reconstructed  SMPL model.

\subsubsection{Volumetric methods}

Volumetric techniques for 3D human body reconstruction do not use statistical models. Instead, they directly infer 3D occupancy grids.  As such, all the methods reviewed in Section~\ref{sec:volumetric_3D_reconstruction} can be used for 3D human body shape reconstruction. An example  is the  approach of Huang \etal~\cite{huang2018deep}, which takes multiple RGB views and their corresponding camera calibration parameters as input, and  predicts a dense 3D field that encodes for each voxel its probability of being inside or outside the human body shape. The surface geometry can then be faithfully reconstructed from the 3D probability field using marching cubes. The approach uses a multi-branch encoder, one for each image, followed by a multi-layer perceptron which aggregates the features that correspond to the same 3D point into a probability value. The approach is able to recover detailed geometry even  on human bodies with cloth but it is limited to simple backgrounds.

To exploit domain-specific knowledge, Varol \etal~\cite{Varol_2018_ECCV} introduce BodyNet, a volumetric approach for inferring, from a single RGB image, the 3D human body shape, along with its 2D and 3D pose, and its partwise segmentation. The approach uses a cascade of four networks;  \textbf{(1)} a 2D pose  and a 2D segmentation network, which operate in parallel, \textbf{(2)} a 3D pose inference network, which estimates the 3D pose of the human body from the input RGB image and the estimated 2D pose and 2D partwise segmentation, and \textbf{(4)} finally, a 3D  shape estimation network, which infers a volumetric representation of the human body shape and its partwise segmentation from the input RGB image and the estimates of the previous networks. By dividing the problem into four tasks, the network can benefit from intermediate supervision, which results in an improved performance.


\subsection{3D face reconstruction}
\label{sec:3DFaces}

Detailed and dense image-based 3D reconstruction of the human face,  which aims to recover shape, pose, expression, skin reflectance,  and finer scale surface details,   is a longstanding problem in computer vision and computer graphics.  Recently, this problem has been formulated as a regression problem and solved using convolutional neural networks.


In this section, we review some of the representative papers.  Most of the recent techniques use parametric representations, which parametrize the manifold of 3D faces. The most commonly used representation is the 3D morphable model (3DMM) of Blanz and Vetter~\cite{blanz1999morphable}, which is an extension of the 2D active appearance model~\cite{cootes2001active} (see also Section~\ref{sec:deformation_model}).  The model captures facial variabily in terms of geometry and texture. Gerig \etal~\cite{gerig2018morphable} extended the model by including expressions as a separate space.  Below, we discuss the various network architectures (Section~\ref{sec:networks_faces}) and their training procedures (Section~\ref{sec:training_faces}).  We will also discuss some of the model-free techniques (Section~\ref{sec:model_free_faces}).
 

\subsubsection{Network architectures}
\label{sec:networks_faces}
The backbone architecture is an encoder, which maps the input image into the parametric model parameters. It is composed of convolutional layers followed by fully connected layers.  In general, existing techniques use generic networks such as AlexNet, or networks specifically trained on facial images such as VGG-Face~\cite{parkhi2015deep} or FaceNet~\cite{schroff2015facenet}.  Tran \etal~\cite{tran2017regressing} use this architecture to regress the $198$ parameters of a 3DMM  that encodes facial identity (geometry) and texture.  It has been trained with 3D supervision using  $\ltwo$ asymetric loss, \ie a loss function that favours 3D reconstructions that are far from the mean.  

Richardson \etal~\cite{richardson20163d} used a similar architecture but perform the reconstruction iteratively. At each iteration, the network  takes  the previously reconstructed face, but projected onto an image using a frontal camera, with the input image, and regresses the parameters of a 3DMM. The reconstruction is initialized with the average face.  Results show that, with three iterations,  the approach can successfully handle face reconstruction  from  images with various expressions and illumination conditions.

One of the main issues with 3DMM-based approaches is that they tend to reconstruct smooth facial surfaces, which lack fine details such as wrinkles and dimples. As such, methods in this category  use a refinement module to recover the fine details.  For instance, Richardson \etal~\cite{richardson20163d} refine the reconstructed face using Shape from Shading (SfS) techniques. Richardson \etal~\cite{Richardson2016Learning}, on the other hand, add a second refinement block, FineNet, which takes as input the depth map of the coarse estimation and recovers using an encoder-decoder network a high resolution facial depth map. To enable end-to-end training, the two blocks are connected with a differentiable rendering layer.  Unlike traditional SfS, the introduction of FineNet treats the calculation of albedo and lighting coefficients as part of the loss function without explicitly estimating these information. However,  lighting is modeled by first-order spherical harmonics, which lead to an inaccurate reconstruction of the facial details.

\subsubsection{Training and supervision}
\label{sec:training_faces}

One of the main challenges  is in how to collect enough training images labelled with their corresponding 3D faces, to feed the network. Richardson \etal~\cite{richardson20163d,Richardson2016Learning}  generate synthetic training data by drawing random samples from the morphable model and rendering the resulting faces. However, a network trained on purely synthetic data may perform poorly when faced with occlusions, unusual lighting, or ethnicities that are not well represented. Genova \etal~\cite{Genova_2018_CVPR} address the lack of training data by including randomly generated synthetic faces in each training batch to provide ground truth 3D coordinates, but train the network on real photographs at the same time. Tran \etal~\cite{tran2017regressing} use an iterative optimization to fit an expressionless model to a large number of photographs, and treat the results where the optimization converged as ground truth. To generalize to faces with expressions, identity labels and at least one neutral image are required. Thus, the potential size of the training dataset is restricted.  


Tewari \etal~\cite{tewari2017mofa}   train, without 3D supervision,  an encoder-decoder  network to simultaneously predict the facial  shape, expression, texture, pose, and lighting. The encoder is a regression network from images to morphable model coordinates, and the decoder is a fixed, differentiable rendering layer that attempts to reproduce the input photograph.  The loss measures the discrepancy between the reproduced photograph and the input one. Since the training loss is based on individual image pixels, the network is vulnerable to confounding variation between related variables. For example, it cannot readily distinguish between dark skin tone and a dim lighting environment.

To remove the need for supervised training  with 3D data and the reliance on inverse rendering, Genova \etal~\cite{Genova_2018_CVPR} propose a framework that learns to minimize a loss based on the facial identity features produced by a face recognition network such as VGG-Face~\cite{parkhi2015deep} or Google's FaceNet~\cite{schroff2015facenet}. In other words, the face recognition network encodes the input photograph as well as the image rendered from the reconstructed face  into feature vectors that are robust to pose, expression, lighting, and even non-photorealistic inputs.  The method then applies a loss that measures the discrepancy between these two feature vectors instead of using pixel-wise distance between the rendered image and the input photograph. The 3D facial shape and texture regressor network is trained   using only a face recognition network, a morphable face model, and a dataset of unlabelled facial images. The approach does not only improve on the accuracy of previous works but also produces 3D reconstructions that are often recognizable as the original
subjects.

\subsubsection{Model-free approaches}
\label{sec:model_free_faces}

Morphable model-based techniques are restricted to the modelled subspace.  As such, implausible reconstructions are possible outside the span of the training data. Other representations such as volumetric grids, which do not suffer from this problem,  have been also explored in the context of 3D face reconstruction. Jackson \etal~\cite{jackson2017large}, for example, propose  a Volumetric  Regression Network (VRN), which takes as input  2D images and predicts their corresponding 3D binary volume instead of a 3DMM. Unlike \cite{tran2017regressing}, the approach can deal with a wide range of expressions, poses and occlusions  without alignment and  correspondences. It, however, fails to recover fine details due to the resolution restriction of volumetric techniques.

Other techniques use intermediate representations. For example, Sela \etal \cite{sela2017unrestricted}  use an Image-to-Image Translation Network based on U-Net~\cite{ronneberger2015u} to estimate a depth image and a facial correspondence map. Then, an iterative deformation-based registration is performed followed by a geometric refinement procedure to reconstruct subtle facial details. Unlike 3DMM, this method can handle large geometric variations.

Feng \etal~\cite{feng20183d} also investigated a model-free method.  First, a densely connected CNN framework is designed to regress 3D facial curves from horizontal and vertical epipolar plane images. Then, these curves are transformed into a 3D point cloud and the grid-fit algorithm \cite{d2005surface} is used to fit a facial surface. Experimental results suggest that this approach is robust to varying poses, expressions and illumination.

\subsection{3D scene parsing}
\label{sec:3D_scene_parsing}

Methods discussed so far are primarily dedicated to the 3D reconstruction of objects in isolation. Scenes with multiple objects pose the additional challenges of delineating objects,   properly handling occlusions, clutter, and uncertainty in shape and pose, and estimating the scene layout. Solutions to this problem involve 3D object detection and recognition, pose estimation, and 3D reconstruction. Traditionally, many of these tasks have been addressed using hand-crafted features.  In the deep learning-based era,  several of the blocks of the pipeline have been replaced with CNNs.

For instance, Izadinia \etal~\cite{izadinia2017im2cad} proposed an approach that is based on recognizing objects in indoor scenes, inferring room geometry, and optimizing 3D object poses and sizes in the room to best match synthetic renderings to the input photo. The approach detects object regions, finds from a CAD database the most similar shapes, and then deforms them to fit the input. The room geometry is estimated using a fully convolutional network. Both the detection and retrieval of objects are performed using Faster R-CNN~\cite{ren2015faster}.  The deformation and fitting, however, are performed via render and match. 
 Tulsiani \etal~\cite{tulsiani2018factoring}, on the other hand, proposed an approach that is entirely based on deep learning. The input, which consists of an RGB image and the bounding boxes of the objects, is processed with a four-branch network. The first branch is an encoder-decoder with skip connections, which estimates the disparity of the scene layout. The second branch takes a low resolution image of the entire scene and maps it into a latent space using a CNN followed by three fully-connected layers. The third branch, which  has the same architecture as the second one,   maps the image at its original resolution to convolutional feature maps, followed by ROI pooling to obtain features for the ROI. The last layer  maps the bounding box location through fully connected layers. The three features are then concatenated and  further processed with fully-connected layers followed by a decoder, which produces a $32^3$ voxel grid of the object in the ROI and its pose in the form of position, orientation, and scale. The method has been trained using synthetically-rendered images with their associated ground-truth 3D scene.  
%
%
%
%
%


\section{Datasets}
\label{sec:datasets_evaluation}

\begin{table*}[t]
	\caption{\label{tab:3ddatasets}Some of the datasets that are used to  train and evaluate the performance of deep learning-based 3D reconstruction algorithms. "img": image. "obj": object. "Bkg": background. "cats": categories. "GT": ground truth. }
	
	\resizebox{\linewidth}{!}{%
	\begin{tabular}{@{} lcccccccccccccc@{} }
	\noalign{\hrule height 1pt}
		&   \multirow{2}{*}{\textbf{Year}} &  \multicolumn{5}{c}{\textbf{Images}} &&   \multicolumn{2}{c}{\textbf{Objects}} & & \multicolumn{3}{c}{\textbf{3D ground truth}} &  \textbf{Camera}  \\ 
			\cline{3-7} \cline{9-10} \cline{12-14}
		& 	     & \textbf{No. imgs} & \textbf{Size} & \textbf{objs per img} & \textbf{Type}& \textbf{Bkg} & &\textbf{Type} & \textbf{No. cats} & & \textbf{No.} & \textbf{Type} & \textbf{Img with 3D GT} & \textbf{params}  \\ 
		
	\noalign{\hrule height .8pt}

	ShapeNet~\cite{chang2015shapenet} & $2015$ & 	$-$ &$-$	& Single &	rendered	 & Uniform & & Generic  & $55$ &  & $51,300$ & 3D model & $51,300$ & Intrinsic \\
	\hline
%
	ModelNet~\cite{wu20153d} & 2015 & $-$ & $-$&  Single & Rendered & Uniform & & Generic & $662$ & & $127,915$ & 3D model & $127,915$ & Intrinsic  \\ 
	\hline
	
	IKEA~\cite{lim2013parsing} &$2013$ & $759$& Variable & Single & Real, indoor & 	Cluttered & & Generic & $7$&  & $219$ & 3D model  & $759$	& Intrinsic+extrinsic \\
	\hline
	
	Pix3D~\cite{sun2018pix3d} & 2018 & 9,531	& $110\times110$ to & Single & Real, indoor& Cluttered & & Generic &	$9$ & & $1015$ & 3D model	& $9,531$ & 	Focal length,  \\ 
						  & 		&		& $3264\times2448$ &  & & & & & & &  & & & extrinsic\\
	\hline
	
	PASCAL 3D+~\cite{xiang2014beyond} & 2014& $30,899$	 & Variable & Multiple & Real, indoor, outdoor & Cluttered & & Generic & $12$ && $36,000$ & 3D model	&  $30,809$ & 	Intrinsic+extrinsic \\ 
	\hline
	
	ObjectNet3D~\cite{xiang2016objectnet3d}	& 2016& $90,127$ & 	Variable & Multiple & Real, indoor, outdoor & Cluttered &  & Generic	& $100$ & & $44,147$ &  3D model & 	$90,127$ & 	Intrinsic+extrinsic \\ 
	\hline
	
	KITTI12~\cite{geiger2012we} & 2012 & $41,778$ & $1240\times376$& Multiple &	Real, outdoor & Cluttered & & Generic & $2$ & &  $40,000$ & Point cloud & $12,000$ & 	Intrinsic+extrinsic \\ 
	\hline
	
	ScanNet~\cite{dai2017scannet}	 & 2017, &   $2,492,518$ & $640 \times 480$, 	& Multiple & Real, indoor & Cluttered &  & Generic & $296$	& & $36,123$ & 	Dense depth & $2,492,518$ & Intrinsic+extrinsic \\ 
							 & 	2018	&  & RGB $1296 \times 968$ \\
	\hline
	Stanford Car~\cite{krause20133d} & 2013	& $16,185$& Variable & Single & Real, outdoor &Cluttered & & Cars & 196&  & $-$ & $-$ & $-$  & $-$ \\ 
	\hline
	
	Caltech-UCSD &	2010, & $6,033$ & Variable & Single & Real & Cluttered & & Birds & $200$ & & $-$ &	$-$ & $-$ & Extrinsic	 \\ 
	Birds 200~\cite{WelinderEtal2010,WahCUB_200_2011}  & 2011 & \\
	\hline
	
	SUNCG~\cite{song2017semantic} &2017& $130,269$ & $-$ & Multiple &Synthetic & Cluttered	& & Generic & $84$ & & $5,697,217$ & Depth, voxel grid & $130,269$ & 	Intrinsic \\ %
	\hline
	Stanford 2D-3D-S & 2016,  & $70,496$ & $1080 \times 1080 $ & Multiple&  Real, indoor & Cluttered & & Generic & $13$ & & 	6,005 & Point cloud,  mesh & $70,496$ &Intrinsic + extrinsic	 \\ 
	~\cite{armeni_cvpr16,armeni2017joint}& 2017 \\
	\hline
	ETHZ CVL  & 2014 & $428$ & $-$ &  Multiple & Real, outdoor& Cluttered & & Generic & $8$ & & $-$ & Point cloud, mesh & $428$ & 	Intrinsic \\ 
	RueMonge~\cite{riemenschneider2014learning} & \\
	\hline
	
	NYU V2~\cite{silberman2012indoor}	& 2012 & $1,449$ & 	$640 \times 480 $ & Multiple & Real, indoor & Cluttered & &Generic & $894$ & & $35,064$ & Dense depth, & $1,449$ & Intrinsic \\ 
		& & & & & & & & & & & & 3D points  \\
	
	\noalign{\hrule height .8pt}
	\end{tabular}
	}
\end{table*}

%
%
%
%
%
%
%
%
%
%
%

Table~\ref{tab:3ddatasets} lists and summarizes the properties of the most commonly used datasets.  Unlike traditional  techniques, the success of deep learning-based 3D reconstruction algorithms depends on the availability of large training datasets. Supervised techniques require images and their corresponding 3D annotations in the form of \textbf{(1)} full 3D models represented as volumetric grids, triangular meshes, or point clouds, or \textbf{(2)} depth maps, which can be dense or sparse. Weakly supervised and unsupervised techniques, on the other hand, rely on additional supervisory signals such as the extrinsic and intrinsic camera parameters and segmentation masks. 

The main challenge in collecting training datasets for deep learning-based 3D reconstruction is two-fold. \textbf{First}, while one can easily collect 2D images,  obtaining their corresponding 3D ground truth is challenging. As such, in many datasets, IKEA, PASCAL 3D+, and ObjectNet3D,  only a relatively small subset of the images are annotated with their corresponding 3D models.  \textbf{Second}, datasets such as ShapeNet and ModelNet, which are the largest 3D datasets currently available, contain 3D CAD models without their corresponding \emph{natural} images since they have been originally intended to benchmark 3D shape retrieval algorithms. 


This issue has been addressed in the literature by data augmentation, which is the process of  augmenting the original sets with synthetically-generated data.  For instance, one  can generate new images and new 3D models by applying  geometric transformations, \eg translation, rotation, and scaling,  to the existing ones. Note that, although some transformations are similarity-preserving, they still enrich the datasets. One can also synthetically render, from existing 3D models, new  2D and 2.5D (\ie depth) views from various (random) viewpoints, poses,  lighting conditions, and backgrounds. They can also be overlaid with natural images or random textures. This, however, results in the domain shift problem, \ie the space of synthetic images is different from the space of real images, which often results in a decline in performance when methods are tested on images of a completely different type.    

Domain shift problem in machine learning has been traditionally addressed using domain adaptation or translation techniques, which are becoming popular in depth estimation~\cite{laga2019survey}. They  are, however, not commonly used for 3D reconstruction. An exception is the work of Petersen \etal~\cite{petersen2019pix2vex}, which observed that a differentiable renderer used in unsupervised techniques may produce images that are different in appearance compared to the input images.  This has been alleviated  through the use of image domain translation.

Finally, weakly supervised and unsupervised techniques (Section~\ref{sec:2D_supervision}) minimize the reliance on 3D annotations. They, however,  require \textbf{(1)} segmentation masks, which can be obtained using the recent state-of-the-art object detection and segmentation algorithms~\cite{ward2019rgb},   and/or \textbf{(2)} camera parameters. Jointly training for 3D reconstruction, segmentation, and camera parameters estimation can be a promising direction for feature research.


\section{Performance comparison}
\label{sec:performance_comparison}

This section discusses the performance of some key methods. We will  present the various performance criteria and metrics (Section~\ref{sec:evaluation_criteria}), and  discuss and compare the performance  of some key methods (Section~\ref{sec:comparison}).

\subsection{Accuracy metrics and performance criteria}
\label{sec:evaluation_criteria}

Let $\shape$ be the ground truth 3D shape and $\hat{\shape}$ the reconstructed one.  Below, we discuss some of the accuracy metrics (Section~\ref{sec:accurcay_metrics}) and performance criteria (Section~\ref{sec:performance_criteria}) used to compare 3D reconstruction algorithms.

\subsubsection{Accuracy metrics}
\label{sec:accurcay_metrics}
The most commonly used quantitative metrics for evaluating the accuracy of 3D reconstruction algorithms include:

\vspace{6pt}
\noi\textbf{(1) The Mean Squared Error (MSE)~\cite{pontes2017image2mesh}}. It is defined as the symmetric surface distance between the reconstructed shape $\hat{\shape}$  and the ground-truth shape  $\shape$, \ie 
	\begin{equation}
		d(\hat{\shape}, \shape) = \frac{1}{n_{\shape}}\sum_{\point \in \shape} d(\point, \hat{\shape})  + \frac{1}{n_{\hat{\shape}}}\sum_{\hat{\point} \in \hat{\shape}} d(\hat{\point}, {\shape}).
	\end{equation}
	Here,  $n_{{\shape}}$  and $n_{\hat{\shape}}$  are, respectively, the number of densely sampled points on $\shape$  and $\hat{\shape}$, and $d(\point, \shape)$ is the distance, \eg the $\lone$ or $\ltwo$ distance, of $\point$ to $\shape$ along the normal direction to $\shape$.  The smaller this measure is, the better is the reconstruction.

\vspace{6pt}
\noi\textbf{(2) Intersection over Union (IoU). } The IoU measures the ratio of the intersection between the volume of the predicted shape  and the volume of the ground-truth,  to the union of the two volumes, \ie 
	\begin{equation}
	IoU_{\epsilon} = \frac{\hat{\vgrid} \cap {\vgrid}  }{ \hat{\vgrid} \cup {\vgrid}  } = \frac{ \sum_{i} \{I(\hat{\vgrid}_{i} > \epsilon  ) * I({\vgrid}_{i}  ) \}  }{  \sum_{i} \{I(I(\hat{\vgrid}_{i}  > \epsilon ) + I({\vgrid}_{i}  ) )\}  },
	\label{eq:IoU}
	\end{equation}
	where $I(\cdot)$ is the indicator function,  $\hat{\vgrid}_{i} $ is the predicted value at the $i-$th voxel,  $\vgrid_{i} $ is the ground truth, and $\epsilon$ is a threshold.   The higher the IoU value, the better is the reconstruction.  This metric is suitable for volumetric reconstructions. Thus, when dealing with surface-based representations, the reconstructed and ground-truth 3D models need to be voxelized. 
	
	
\vspace{6pt}
\noi \textbf{(3) Mean of Cross Entropy (CE) loss~\cite{yang20173d}. } It is defined as follows;
	\begin{equation}
	\small{
		CE =  -\frac{1}{\npoints} \sum_{i=1}^{\npoints} \left\{ p_i \log\hat{p}_i  + (1 - p_i)\log(1 - \hat{p}_i      \right\}.
	}
	\end{equation}
	where $N$ is the total number of voxels or points, depending whether using a volumetric or a point-based representation.  $p$ and $\hat{p}$ are, respectively, the ground-truth and the predicted value at the $i$-voxel or point. The lower the CE value is, the better is the reconstruction.

\vspace{6pt}
\noi \textbf{(4) Earth Mover Distance (EMD) and Chamfer Distance (CD). } These distances are, respectively, defined in Equations~\eqref{eq:emd} and \eqref{eq:CD}.

\subsubsection{Performance criteria} 
\label{sec:performance_criteria}

In addition to these quantitative  metrics, there are several qualitative aspects that are used to evaluate the efficiency of these methods. This includes:

\vspace{6pt}
\noi \textbf{(1) Degree of 3D supervision. } An  important aspect of deep learning-based 3D reconstruction methods is the degree of 3D supervision they require at training. In fact, while obtaining RGB  images is easy, obtaining their corresponding ground-truth 3D data is quite challenging. As such, techniques that require minimal or no 3D supervision are usually preferred  over those that require ground-truth 3D information  during training. 
	

\vspace{6pt}
\noi \textbf{(2) Computation time. }  While training can be slow,  in general, its is desirable to achieve real-time performance at runtime. 
	
\vspace{6pt}
\noi \textbf{(3) Memory footprint. }  Deep neural networks have a large number of parameters. Some of them operate on volumes  using 3D convolutions.  As such, they usually require a large memory storage, which can affect their performance at runtime and limit their usage.


\subsection{Comparison and discussion}
\label{sec:comparison}

\begin{figure}[t]
\centering{
	\begin{tabular}{@{}c@{}}
		\includegraphics[trim={2.6cm 1.1cm 3cm .5cm},clip, width=0.48\textwidth]{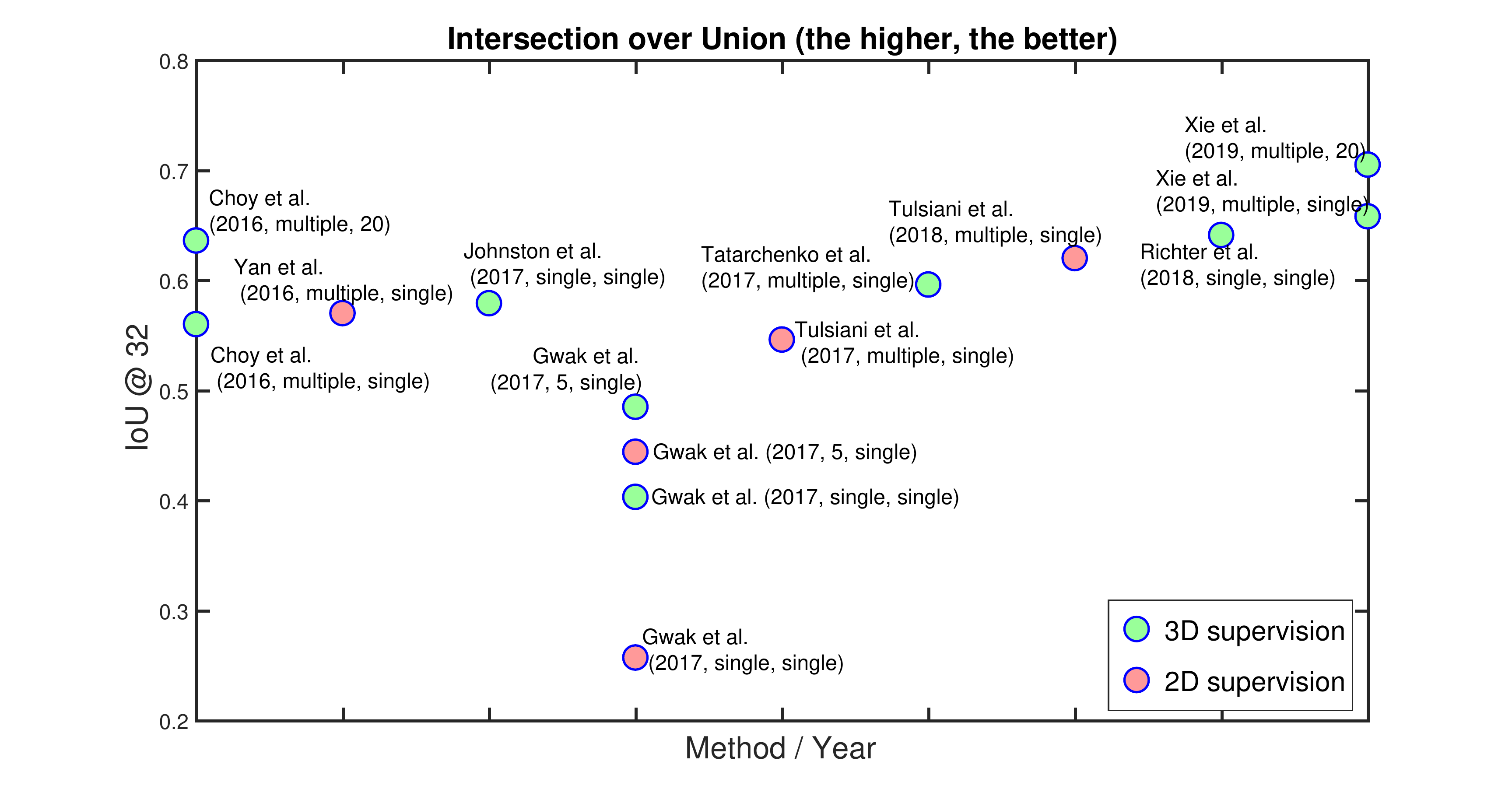}\\
		\small{(a) IoU of volumetric methods. }\\
		\includegraphics[trim={2.6cm 1.1cm 3cm .5cm},clip, width=.48\textwidth]{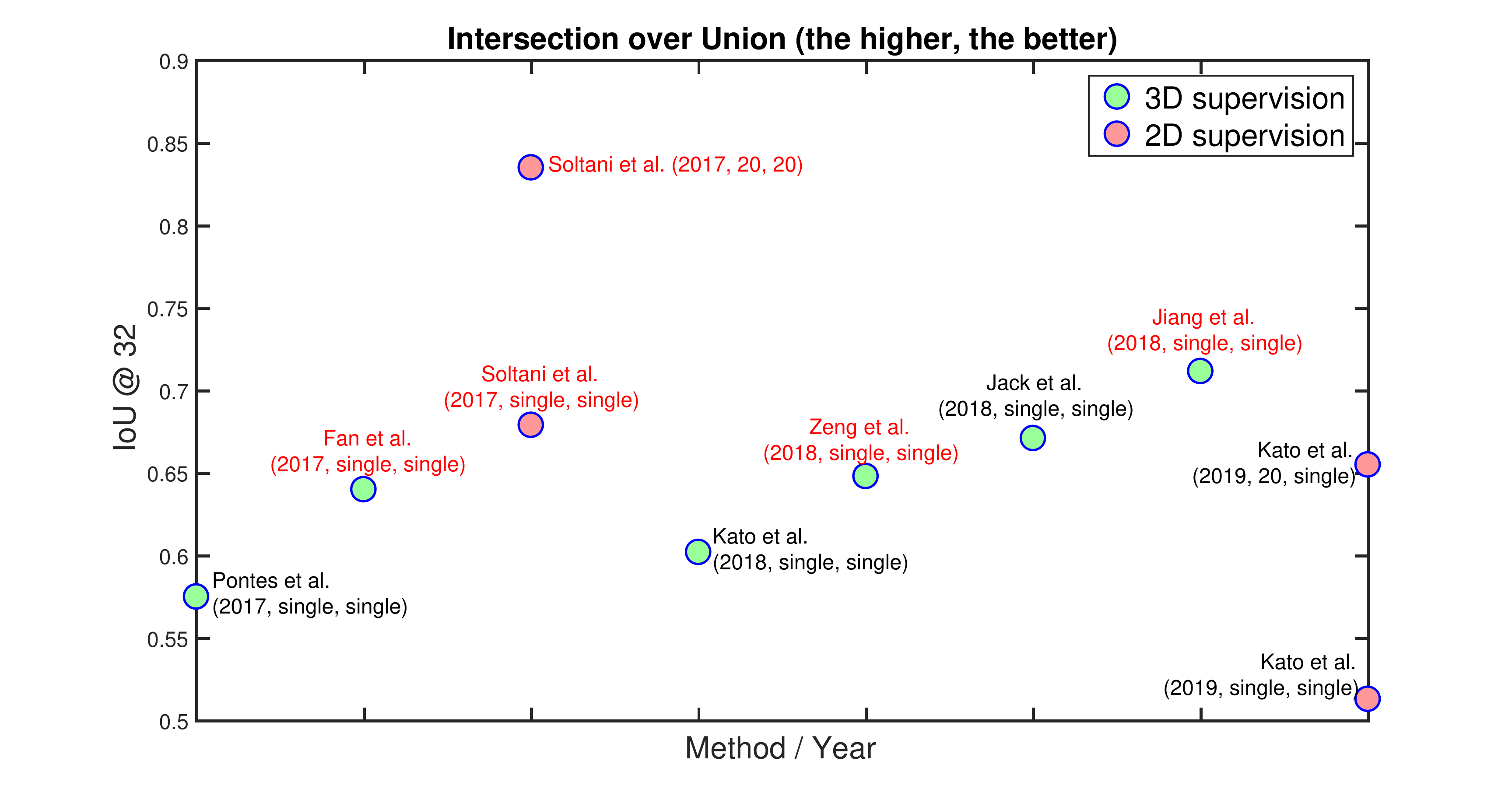} \\
		 \small{(b)  IoU of surface-based methods.}\\
		\includegraphics[trim={2.6cm 1.1cm 3cm .5cm},clip, width=.48\textwidth]{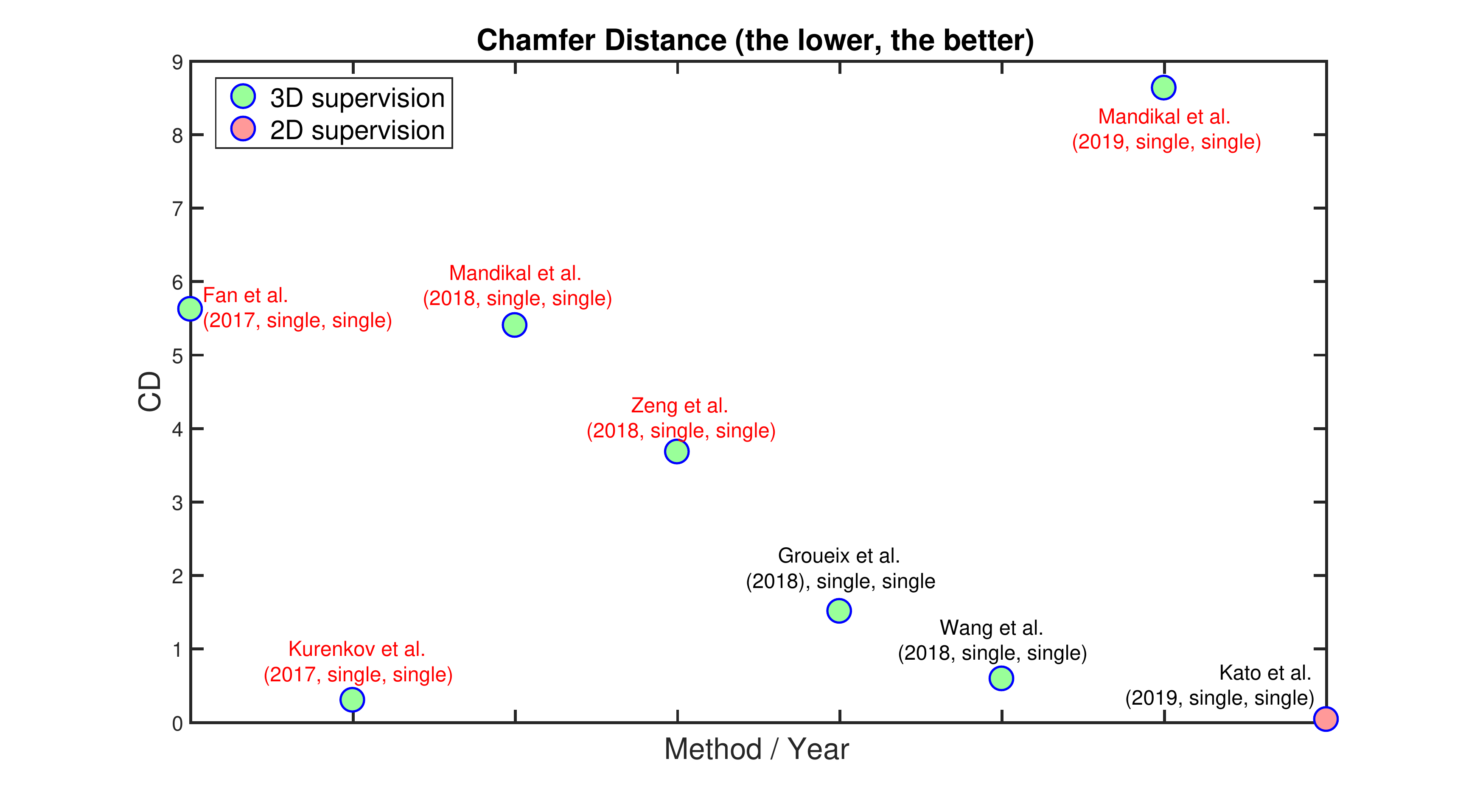}\\
		 \small{(c)  CD  of surface-based methods.}\\				
	\end{tabular}
\caption{\label{fig:performance_shapenet} Performance of some key methods on the ShapeNet dataset. References highlighted in red are point-based. The IoU is computed on grids of size $32^3$. The label next to each circle is encoded as follow: First author \etal~(year, $\nimages$ at training, $\nimages$ at test), where $\nimages$ is the number of  input images. Table~\ref{tab:performance_summary} provides a  detailed comparison. }
}
\end{figure}

We present the improvement in reconstruction accuracy over the past $4$ years in Fig.~\ref{fig:performance_shapenet}, and the performance of  some representative methods in Table~\ref{tab:performance_summary}.

\begin{sidewaystable*}[ph!]
	\caption{\label{tab:performance_summary}Performance summary of some representative methods. "Obj.": objects.  "Time" refers to timing in milliseconds.  "U3D": unlabelled 3D. \# params: number of the parameters of the network.  "mem.": memory requirements. "Resol.": resolution. "Bkg": background.}
	\resizebox{\linewidth}{!}{%
	\begin{tabular}{@{}lccccccccccccccc@{}}
		\toprule
		\multirow{3}{*}{\textbf{Method}} & \multicolumn{4}{c}{\textbf{Input}} & & \multicolumn{2}{c}{\textbf{Output}} & & \multicolumn{2}{c}{\textbf{Training}}  & & \multicolumn{4}{c}{\textbf{Performance@(ShapeNet, Pix3D,Pascal3D+) }} \\
		\cline{2-5} \cline{7-8} \cline{10-11} \cline{13-16}
		& \multirow{2}{*}{Train} & \multirow{2}{*}{Test} & \multirow{2}{*}{Bkg} & \#  & { }  & \multirow{2}{*}{Type} & \multirow{2}{*}{Resol.} & { } & \multirow{2}{*}{Supervision} &  \multirow{2}{*}{Network} & { } & \multirow{2}{*}{IoU} &  \multirow{2}{*}{CD} & \multirow{2}{*}{Time} & \multirow{1}{*}{Memory} \\
		&  &  &  & objects & { }  &  &  & { } &  &   & { } & &   &  &  (\# params., mem.)\\
		\toprule

		\multirow{2}{*}{Xie \etal~\cite{xie2019pix2vox}} 	 & \multirow{2}{*}{$\nimages\ge 1$ RGB, 3D GT} &  1 	RGB  &  Clutter &  $1$  & &  	
								    Volumetric  & $32^2$ &   &		
								     3D   &  Encoder (VGG16),  &  &	
								    $(0.658, -, 0.669)@ 32^3$ & $(-, -, -)$ & $9.9$ & ($114.4$M, $-$) \\
								    
								    &   &  20 RGB  &   &    & &  	
								      &   &   &		
								     3D   & Decoder, Refiner   &  &	
								    $(0.705, -, -)@ 32^3$ & $(-, -,-)$ & $-$ & $-$ \\
								    
		\midrule
		Richter \etal~\cite{Richter_2018_CVPR} 	 & $\nimages = 1$ RGB, 3D GT &  1 RGB &  Clean  &  $1$  &  &  	
								      Volumetric & $512^3$ &   &		
								      3D  &  encoder, 2D decoder &  &	
								    $(0.641, -,-)@32^3$ & $(-, -,-)$ & $-$ & $-$ \\

		\midrule
		 Tulsiani \etal~\cite{Tulsiani_2018_CVPR}	 &  $\nimages> 1$ RGB, segmentation & 1 RGB  &  Clutter  &  $1$  & &  	
								     Volumetric + pose& $-$ &   &		
								     2D  (multiview) &  Encoder, decoder,  &  &	
								    $(0.62, -,-)@32^3$ & $(-, -,-)$ & $-$ & $-$ \\	
								    
								     &  &   &   &    & &  	
								        & &   &		
								        &pose CNN   &  &	
								        &   &  &  \\					    
		\midrule
		 Taterchenko \etal~\cite{tatarchenko2017octree}  & $\nimages \ge 1$ RGB + 3D GT  &   1 RGB & Clutter   &  $1$  & &  	
								      Volumetric & $512^3$ &   &		
								      3D    & Octree Generating  &  &	
								      $(-, -,-)$ & $(-, -,-)$ & $2.06$s & ($-$, $0.88$GB) \\
								    
 									&  &   &   &    & &  	
								      & $32^3$ &   &		
								        &  Network&  &	
								    $(0.596, -,0.504)@ 32^3$ & $(-, -,-)$ & $16$ & ($12.46$M, $0.29$GB)\\
		
		\midrule
		 Tulsiani \etal~\cite{tulsiani2017multi}  & $\nimages > 1$ RGB, silh., (depth) &   1 RGB & Clutter  &  $1$  & &  	
								      Volumetric & $32^3$ &   &		
								       2D (multiview) &  encoder-decoder&  &	
								    $(0.546, -,0.536)@ 32^3$ & $(-, -,-)$ & $-$ & $-$ \\

		\midrule
		 Wu \etal~\cite{wu2017marrnet}	 &  $\nimages=1$ RGB, 3D GT&   1 RGB &  Clutter &  $1$  & &  	
								     Volumetric  & $128^3$ &    &		
								     $2$D and $2.5$D   &  TL, 3D-VAE +  &  &	
								    $(0.57, -,0.39)@128^3$ & $(-, -,-)$ & $-$ & $-$ \\
								    
								    &  &   &   &    & &  	
								        & &   &		
								        &encoder-decoder   &  &	
								        &   &  &  \\

		\midrule
		 Gwak \etal~\cite{gwak2017weakly}	 &  1 RGB, silh., pose & 1 RGB  &   &   & &  	
								      & $-$ &   &		
								       2D  & GAN &  &	
								    $(0.257, -,-)@32^3$ & $(-, -,-)$ & $-$ & $-$ \\
								    
								    &  1 RGB, silh., pose, U3D & 1 RGB  & Clutter   &  $1$  & &  	
								    Volumetric  & $-$ &   &		
								       2D, U3D  & GAN &  &	
								    $(0.403, -,-)@32^3$ & $(-, -,-)$ & $-$ & $-$ \\
								    
								    &  5 RGB, silh., pose & 1 RGB  &   &    & &  	
								      & $-$ &   &		
								       2D  & GAN &  &	
								    $(0.444, -,-)@32^3$ & $(-, -,-)$ & $-$ & $-$ \\
								    
								     &  5 RGB, silh., pose, U3D & 1 RGB  &   &    & &  	
								      & $-$ &   &		
								       2D, U3D & GAN &  &	
								    $(0.4849, -,-)@32^3$ & $(-, -,-)$ & $-$ & $-$ \\
								    
		\midrule
		 	Johnston \etal~\cite{johnston2017scaling} &  $\nimages = 1$ RGB, 3D GT& 1 RGB  &  Clutter  &  $1$  & &  	
								      Volumetric & $128^3$ &   &		
								      3D  &  Encoder + IDCT &  &	
								    $(0.417, -,0.4496)@128^3$ & $(-, -,-)$ & $32$ & ($-$, $2.2$GB) \\
								    
								    &   &   &     &  $1$  & &  	
								      Volumetric & $32^3$ &   &		
								      3D  &  Encoder + IDCT &  &	
								    $(0.579, -,0.5474)@32^3$ & $(-, -,-)$ & $15$ & ($-$, $1.7$GB) \\
								    
		\midrule
		 	Yan \etal~\cite{yan2016perspective} & $\nimages > 1$ RGB, silh., pose  & 1 RGB  &  Clean &  $1$  & &  	
								      Volumetric & $32^3$ &   &		
								      2D  &  encoder-decoder &  &	
								    $(0.57, -,-)@32^3$ & $(-, -,-)$ & $-$ & $-$ \\								    
								    
		\midrule
		 Choy \etal~\cite{choy20163d}	 &  $\nimages\ge 1$ RGB, 3D GT&  1 RGB & Clutter   &  $1$  & &  	
								     Volumetric  & $32^3$ &   &		
								      3D   &  encoder-LSTM- &  &	
								    $(0.56, -,0.537)@ 32^3$ & $(-, -,-)$ & $73.35$ & ($35.97$M, $>4.5$GB) \\
						&  &  20 RGB &   &    & &  	
								       & $32^3$ &   &		
								        & decoder &  &	
								    $(0.636, -,-)@ 32^3$ & $(-, -,-)$ & $-$ & $-$ \\

		\bottomrule	
		
		Kato \etal~\cite{kato2019learning} & $\nimages = 1$ RGB, silh., pose & 1 RGB & Clutter &  $1$  & &  	
								    Mesh + texture & $-$ &   &		
								    2D   & Encoder-Decoder &  &	
								    $(0.513, -, -) @ 32^3$ & $(0.0378, -, -)$ &  $-$ & $-$ \\
				 	 			   & $20$ RGB, silh., poses &   &   &    & &  	
								      & $-$ &   &		
								        &  &  &	
								    $(0.655, -,-)@ 32^3$ & $(-, -,-)$ & $-$ & $-$ \\	
		
		\midrule
		 Mandikal \etal~\cite{mandikal2019dense}	 & $\nimages = 1$ RGB, 3D GT  &   1 RGB & Clean  &  $1$  & &  	
								      Point cloud & $16384$ &   &		
								     3D   &  Conv + FC layers,  &  &	
								    $(-, -,-)$ & $(8.63, -,-)$ & $-$ & ($13.3$M, $-$) \\	
								    
								      &  &  &  &  & &  	
								      &  &   &		
								   & Global to local &  &	
								    & & &  \\
								    								    
		\midrule
		 Jiang \etal~\cite{Jiang_2018_ECCV}  & $\nimages = 1$ RGB + 3D GT	 & 1 RGB  &  Clutter &  $1$  & &  	
								      Point cloud& $1024$ &   &		
								      3D  & $2\times$ (encoder-decoder),  &  &	
								    $(0.7116, -,-)@32^3$ & $(-, -,-)$ & $-$ & $-$ \\	
								    
								     &  &   &   &  & &  	
								        & &   &		
								        &GAN   &  &	
								        &   &  &  \\	
								    
		\midrule
		 Zeng \etal~\cite{zeng2018inferring}	 & 1 RGB, silh., pose, 3D &  1 RGB & Clutter  &  $1$  & &  	
								      Point cloud& $1024$ &   &		
								       3D + self & encoder-decoder $+$  &  &	
								        $(0.648, -,-)@32^3$ & $(3.678, -,-)$ & $-$ & $-$ \\	
								    
								    	&  &   &   &    & &  	
								        & &   &		
								        & point auto-encoder  &  &	
								        &   &  &  \\	
								    
		\midrule
		 Kato \etal~\cite{kato2018neural}    & $\nimages = 1$ RGB, silh., 	 & 1 RGB    &  Clutter &  $1$  & &  	
								      Mesh & $642$ &   &		
								        2D & encoder + FC layers &  &	
								    $(0.602, -,-)@32^3$ & $(-, -,-)$ & $-$ & $-$ \\	
								    
		\midrule
		 Jack \etal~\cite{jack2018learning}	 & 1 RGB, FFD params & 1 RGB  & Clean  &  $1$  & &  	
								      Mesh & $16384$ &   &		
								       3D  & Encoder + FC layers &  &	
								    $(0.671, -,-)@32^3$ & $(-, -,-)$ & $-$ & $-$ \\	

		\midrule
		Groueix  \etal~\cite{groueix2018atlasNet}		 & $\nimages = 1$ RGB, 3D GT  &   1 RGB&   Clean &  $1$  & &  	
								      Mesh& $1024$ &   &		
								      3D  & Multibranch MLP  &  &	
								    $(-, -,-)$ & $(1.51, -,-)$ & $-$ & $-$ \\

		\midrule
		 Wang  \etal~\cite{Wang_2018_ECCV} & 1 RGB, 3D GT &  1 RGB  &  Clean &  $1$  & &  	
								     Mesh  & $2466$ &   &		
								     3D   & see Fig.~\ref{fig:defo_models} (left) &  &	
								    $(-, -,-)$ & $(0.591, -,-)$ & $-$ & $-$ \\

		\midrule
		 Mandikal \etal~\cite{mandikal20183d}	 &  $\nimages = 1$ RGB, 3D GT & 1 RGB &   Clutter&  $1$  & &  	
								     Point cloud & $2048$ &   &		
								     3D   &  3D-VAE, &  &	
								    $(-, -,-)$ & $(5.4, -,-)$ & $-$ & $-$ \\	
								    
								     &  &   &   &    & &  	
								        & &   &		
								        &TL-embedding   &  &	
								        &   &  &  \\			
								        						    
		\midrule
		Soltani  \etal~\cite{soltani2017synthesizing} & 20 silh., poses & 1 silh.  &  Clear &  $1$  & &  	
								      		   20 depth maps	   & $-$ &   &		
								        2D &  3D-VAE &  &	
								    $(0.835, -,-)@32^3$ & $(-, -,-)$ & $-$ & $-$ \\	
								    
								    & 1 silh. &   1 silh. &   &    & &  	
								      	20 depth maps	   & $-$ &   &		
								        &    &  &	
								    $(0.679, -,-)@32^3$ & $(-, -,-)$ & $-$ & $-$ \\	
								    
		\midrule
		Fan  \etal~\cite{fan2017point}	 &  $\nimages = 1$ RGB, 3D GT&   1 RGB &  Clutter  &  $1$  & &  	
								      Point set & $1024$ &   &		
								     3D    & see Fig.~\ref{fig:architectures_pointbased}-(a) &  &	
								    $(0.64, -,-)@32^3$ & $(5.62, -,-)$ & $-$ & $-$ \\	
								    
		\midrule
		Pontes  \etal~\cite{pontes2017image2mesh}	 & $\nimages = 1$ RGB, GT FFD  & 1 RGB  & Clean  &  $1$  & &  	
								      Mesh & $-$ &   &		
								       3D & Encoder + FC layers &  &	
								    $(0.575, -,0.299)@32^3$ & $(-, -,-)$ & $-$ & $-$ \\

		\midrule
		Kurenkov  \etal~\cite{kurenkov2017deformnet}	 &  $\nimages = 1$ RGB, 3D GT & 1 RGB  &   Clean &  $1$  & &  	
								      Point cloud & $1024$ &   &		
								      3D  & 2D encoder, 3D encoder, &  &	
								    $(-, -,-)$ & $(0.3, -,-)$ & $-$ & $-$ \\	
								    &  &  &  &  & &  	
								      &  &   &		
								   & 3D decoder &  &	
								    & & &  \\

%
%

		\bottomrule
	\end{tabular}
	}
\end{sidewaystable*}

The majority of early works resort to voxelized representations~\cite{choy20163d,girdhar2016learning,varley2016shape,wu2016learning,yan2016perspective}, which can represent both the surface and the internal details of complex objects of arbitrary topology. With the introduction of space partitioning techniques such as O-CNN~\cite{wang2017cnn}, OGN~\cite{tatarchenko2017octree}, and OctNet~\cite{riegler2017octnet}, volumetric techniques can attain relatively high resolutions, \eg $512^3$. This is due to the significant gain in memory efficiency. For instance, the OGN of~\cite{tatarchenko2017octree} reduces the memory requirement for the reconstruction of volumetric grids of size $32^3$ from $4.5$GB in~\cite{choy20163d}  and $1.7$GB in~\cite{johnston2017scaling} to just $0.29$GB (see Table~\ref{tab:performance_summary}). However, only a few papers, \eg~\cite{Cao_2018_ECCV},  adopted these techniques due to the complexity of their implementation. To achieve high resolution 3D volumetric reconstruction, many recent  papers use intermediation, through multiple depth maps, followed by volumetric~\cite{yang2018dense,Cherabier_2018_ECCV,Richter_2018_CVPR,zhang2018learning} or point-based~\cite{zeng2018inferring} fusion. More recently, several papers start to focus on mechanisms for learning continuous Signed Distance Functions~\cite{park2019deepsdf,chen2019learning}  or continuous occupancy grids~\cite{mescheder2019occupancy}, which are less demanding in terms of memory requirement. Their advantage is that since they learn a continuous field, the reconstructed 3D object can be extracted at the desired resolution.

Fig.~\ref{fig:performance_shapenet} shows the evolution of the performance over the years, since $2016$, using the ShapeNet dataset~\cite{wu20153d}  as a benchmark. On the IoU metric, computed on volumetric grids of size $32^3$, we can see that methods that use multiple views at training and/or at testing outperform those that are based solely on single views. Also, surface-based techniques,  which started to emerge in $2017$ (both mesh-based~\cite{pontes2017image2mesh}  and point-based~\cite{fan2017point,kurenkov2017deformnet}), slightly outperform volumetric methods. Mesh-based techniques,  however,  are limited to genus-0 surfaces or surfaces with the same topology as the template. 

Fig.~\ref{fig:performance_shapenet} shows that, since their introduction in 2017 by Yan \etal~\cite{yan2016perspective}, 2D supervision-based methods significantly improved in performance. The IoU curves of Figures~\ref{fig:performance_shapenet}-(a) and (b), however,  show that methods that use 3D supervision achieve slightly better performance. This can be attributed to the fact that 2D-based supervision methods use loss functions that are based on 2D binary masks and silhouettes.  However, multiple 3D objects can explain the same 2D projections. This 2D to 3D ambiguity has been addressed either by using  multiple  binary masks captured from multiple viewpoints~\cite{soltani2017synthesizing},  which  can only reconstruct the visual hull and as such, they are limited in accuracy, or by using adversarial training~\cite{gwak2017weakly,Wu_2018_ECCV}, which constrains the reconstructed 3D shapes to be within the manifold of valid classes.



\section{Future research directions}
\label{sec:future_work}
In light of the extensive research undertaken in the past five years, image-based 3D reconstruction using deep learning techniques has achieved promising results. The topic, however, is
still in its infancy and further developments are yet to be expected. In this section, we present some of the current issues and highlight directions for future research.

\vspace{6pt}
\noi\textit{(1) Training data issue. } The success of deep learning techniques depends heavily on the availability of training data. Unfortunately, the size of the publicly available datasets that include both images and their 3D annotations is small compared to the training datasets used in tasks such as classification and recognition.  2D supervision techniques have been used to address the lack of 3D training data. Many of them, however, rely on silhouette-based supervision and thus they can only reconstruct the visual hull.   As such, we expect to see in the future more papers proposing new large-scale datasets, new weakly-supervised and unsupervised methods that leverage various visual cues, and new domain adaptation techniques where networks trained with data from a certain domain, \eg synthetically rendered images, are adapted to a new domain, \eg in-the-wild images, with minimum retraining and supervision.  Research on realistic rendering techniques that are able to close the gap between real images and synthetically rendered images can potentially contribute towards addressing the training data issue. 

\vspace{6pt}
\noi\textit{(2) Generalization to unseen objects. } Most of the state-of-the-art papers split a dataset into three subsets for training, validation, and testing, \eg ShapeNet or Pix3D, then report the performance on the test subsets.  However, it is not clear how these methods would perform on a completely unseen object/image categories. In fact, the ultimate goal of 3D reconstruction method is to be able to reconstruct any arbitrary 3D shape from arbitrary images. Learning-based techniques, however, perform well only on images and objects spanned by the training set. Some recent papers, \eg Cherabier \etal~\cite{Cherabier_2018_ECCV}, started to address this issue.  An interesting direction for future research, however,  would be to combine traditional and learning based  techniques to improve the generalization of the latter methods.

\vspace{6pt}
\noi\textit{(2) Fine-scale 3D reconstruction. } Current state-of-the-art techniques are able to recover the coarse 3D structure of shapes. Although recent works have significantly improved the resolution of the reconstruction by using refinement modules, they still fail to recover thin and small parts such as plants, hair, and fur. 
	
\vspace{6pt}	
\noi\textit{(3) Reconstruction vs. recognition. } 3D reconstruction  from images is an ill-posed problem. As such, efficient solutions  need to combine low-level image cues, structural knowledge, and high-level object understanding. As outlined in the recent paper of Tatarchenko \etal~\cite{Tatarchenko_2019_CVPR}, deep learning-based reconstruction methods  are biased towards recognition and retrieval. As such, many of them do not generalize well and fail to recover fine-scale details. Thus, we expect in the future to see more research on how to combine top-down approaches (\ie recognition, classification, and retrieval) with bottom-up approaches (\ie pixel-level reconstruction based on geometric and photometric cues).  This also has the potential to improve the generalization ability of the methods, see item (2) above.

\vspace{6pt}
\noi\textit{(4) Specialized instance reconstruction. } We expect in the future to see more synergy between class-specific knowledge modelling and deep learning-based 3D reconstruction in order to leverage domain-specific knowledge. In fact,  there is an increasing interest in reconstruction methods that are specialized in specific classes of objects such as human bodies and body parts (which we have briefly covered in this survey),  vehicles, animals~\cite{kanazawa2018learning}, trees, and buildings. Specialized methods exploit prior and domain-specific knowledge to optimise the network architecture and its training process. As such, they usually perform better than the general framework. However, similar to deep learning-based 3D reconstruction, modelling prior knowledge, \eg by using advanced statistical shape models~\cite{kurtek2013landmark,laga2017numerical,wang2018shape,wang2018statistical,laga2018survey}, requires 3D annotations, which are not easy to obtain for many classes of shapes, \eg animals in the wild.  

	
\vspace{6pt}
\noi\textit{(5) Handling multiple objects in the presence of occlusions and cluttered backgrounds. } Most of the state-of-the-art techniques deal with images that contain a single object. In-the-wild images, however, contain multiple objects of different categories. Previous works employ detection followed by reconstruction within regions of interests, \eg\cite{tulsiani2018factoring}. The detection and then reconstruction modules operate independently from each other. However, these tasks are inter-related and can benefit from each other if solved jointly. Towards this goal, two important issues should be addressed. The first one is the lack of training data for multiple-object reconstruction. Second,  designing appropriate CNN architectures, loss functions, and learning methods are important especially for  methods that  are trained without 3D supervision. These, in general, use  silhouette-based loss functions, which require accurate object-level segmentation.

	
\vspace{6pt}
\noi\textit{(6) 3D video. } This paper focused on 3D reconstruction from one or multiple images, but with no temporal correlation. There is, however, a growing interest in 3D video, \ie 3D reconstruction of entire video streams where successive frames are temporally correlated. On one hand, the availability of a sequence of frames can improve the reconstruction, since one can exploit the additional information available in subsequent frames to disambiguate and refine the reconstruction at the current frame. On the other hand, the reconstruction should be smooth and consistent across frames. 
	
\vspace{6pt}
\noi\textit{(7) Towards full 3D scene parsing. }  Finally, the ultimate goal is to be able to semantically parse a full 3D scene from one or multiple of its images. This requires joint  detection, recognition, and reconstruction. It would also require capturing and modeling spatial relationships and interactions between objects and between object parts.  While there have been a few attempts in the past to address this problem, they are mostly limited to indoor scenes with strong assumptions about the geometry and locations of the objects that compose the scene. 

\section{Summary and concluding remarks}
\label{sec:summary}

This paper provides a comprehensive survey of the past five years  developments  in the field of image-based 3D object reconstruction using deep learning techniques. We classified the state-of-the-art into volumetric, surface-based, and point-based techniques. We then discussed methods in each category based on their input, the network architectures, and the training mechanisms they use.  We have also discussed and compared the performance of some key methods. 


This survey focused on methods that define 3D reconstruction as the problem of recovering the 3D geometry of objects from one or multiple  RGB images. There are, however,  many other related problems that share similar solutions. The closest topics include depth reconstruction from RGB images, which has been recently addressed using deep learning techniques, see the recent survey of Laga~\cite{laga2019survey}, 3D shape completion~\cite{varley2016shape,yang20173d,dai2017shape,han2017high,wang2017shape,zelek2017point,litany2017deformable}, 3D reconstruction from depth images~\cite{yang20173d}, which can  be seen as a 3D fusion and  completion problem,  3D reconstruction and modelling  from hand-drawn 2D sketches~\cite{lun20173d,delanoy2017you}, novel view synthesis~\cite{li2017deep,niu2018im2struct}, and 3D shape structure recovery~\cite{tatarchenko2016multi,zisserman2017silnet,zou20173d,rezende2016unsupervised}.   These topics have been extensively investigated in the past five years and  require separate survey papers.

%

\section*{Acknowledgements}
Xian-Feng Han is supported by a China Scholarship Council (CSC) scholarship. This work was supported in part by ARC DP150100294 and DP150104251.

\bibliographystyle{IEEEtran}
\bibliography{reconstruction}

\end{document}